\documentclass[10pt,journal,compsoc]{IEEEtran}

\newcounter{docpart}
\def\savestatus{%
\newwrite\tempfile%
\immediate\openout\tempfile=docstatus\arabic{docpart}.dat%
\immediate\write\tempfile{\theassumption}%
\immediate\write\tempfile{\thealgorithm}%
\immediate\write\tempfile{\thetheorem}%
\immediate\write\tempfile{\theproposition}%
\immediate\write\tempfile{\thelemma}%
\immediate\write\tempfile{\theclaim}%
\immediate\write\tempfile{\thedefinition}%
\immediate\write\tempfile{\theassumption}%
\immediate\write\tempfile{\thecorollary}%
\immediate\write\tempfile{\theproperty}%
\immediate\write\tempfile{\theremark}%
\immediate\write\tempfile{\theequation}%
\immediate\closeout\tempfile%
}

\newcounter{olddocpart}

\ifCLASSINFOpdf
\else
\fi

\usepackage{graphicx,subfigure}
\usepackage[colorlinks,linkcolor=black,anchorcolor=black,citecolor=black]{hyperref}

\usepackage{amsfonts,amsmath,amssymb,amsthm,bm}
\usepackage{algorithm,algorithmic}
\usepackage{mathrsfs}

\usepackage[numbers]{natbib}
\usepackage{multibib}
\newcites{app}{References}

\usepackage[english]{babel}
\usepackage{mdwlist}

\usepackage{comment}
\usepackage{cases}
\usepackage{multirow}

\newtheorem{theorem}{Theorem}
\newtheorem{proposition}{Proposition}
\newtheorem{lemma}{Lemma}
\newtheorem{claim}{Claim}
\newtheorem{definition}{Definition}
\newtheorem{assumption}{Assumption}
\newtheorem{corollary}{Corollary}
\newtheorem{property}{Property}
\newtheorem{remark}{Remark}

\def\trans{^{\rm T}}
\def\calA{\mathcal{A}}
\def\calB{\mathcal{B}}
\def\calC{\mathcal{C}}
\def\calD{\mathcal{D}}

\def\calL{\mathcal{L}}
\def\calS{\mathcal{S}}

\def\ttheta{\mbox{\boldmath$\theta$}}
\def\bbP{\mathbb{P}}

\def\C{\mathbf{C}}
\def\X{\mathbf{X}}

\def\W{\mathbf{W}}
\def\U{\mathbf{U}}
\def\V{\mathbf{V}}
\def\P{\mathbf{P}}
\def\Q{\mathbf{Q}}
\def\I{\mathbf{I}}

\def\M{\mathbf{M}}
\def\S{\mathbf{S}}
\def\I{\mathbf{I}}
\def\E{\mathbf{E}}

\def\x{\mathbf{x}}
\def\y{\mathbf{y}}
\def\b{\mathbf{b}}
\def\z{\mbox{\boldmath$z$}}

\def\w{\mathbf{w}}

\def\p{\mathbf{p}}
\def\q{\mathbf{q}}

\def\ttheta{\mbox{\boldmath$\theta$}}

\def\vvartheta{\mbox{\boldmath$\vartheta$}}
\def\LLambda{\mbox{\boldmath$\Lambda$}}
\def\SSigma{\mbox{\boldmath$\Sigma$}}

\def\trans{^{\rm T}}
\def\calA{\mathcal{A}}
\def\calB{\mathcal{B}}
\def\calC{\mathcal{C}}
\def\calD{\mathcal{D}}

\def\calL{\mathcal{L}}
\def\calS{\mathcal{S}}

\def\ttheta{\mbox{\boldmath$\theta$}}
\def\bbP{\mathbb{P}}

\def\C{\mathbf{C}}
\def\X{\mathbf{X}}

\def\W{\mathbf{W}}
\def\U{\mathbf{U}}
\def\V{\mathbf{V}}
\def\P{\mathbf{P}}
\def\Q{\mathbf{Q}}
\def\I{\mathbf{I}}

\def\M{\mathbf{M}}
\def\S{\mathbf{S}}
\def\I{\mathbf{I}}
\def\E{\mathbf{E}}
\def\T{\mathbf{T}}

\def\x{\mathbf{x}}
\def\y{\mathbf{y}}
\def\b{\mathbf{b}}
\def\z{\mbox{\boldmath$z$}}

\def\w{\mathbf{w}}
\def\u{\mathbf{u}}

\def\p{\mathbf{p}}
\def\q{\mathbf{q}}

\def\ttheta{\mbox{\boldmath$\theta$}}

\def\vvartheta{\mbox{\boldmath$\vartheta$}}
\def\LLambda{\mbox{\boldmath$\Lambda$}}
\def\SSigma{\mbox{\boldmath$\Sigma$}}

\begin{document}
\title{Model-Protected Multi-Task Learning}

\author{
\IEEEauthorblockN{Jian Liang, Ziqi Liu, Jiayu Zhou, Xiaoqian Jiang, Changshui Zhang, \IEEEmembership{Fellow, IEEE}, Fei Wang}

\IEEEcompsocitemizethanks{\IEEEcompsocthanksitem J. Liang and C. Zhang are with the Department of Automation, Tsinghua University, State Key Laboratory of Intelligent Technologies and Systems, Tsinghua National Laboratory for Information Science and Technology, Beijing, 100084, China.\protect\\
E-mail: \{liangjian12,zcs\}@\{mails,mail\}.tsinghua.edu.cn.
\IEEEcompsocthanksitem Z. Liu was with the Department of Computer Science, Xi'an Jiaotong University, Xi'an, 710049, China.
\IEEEcompsocthanksitem J. Zhou is with the Department of Computer Science and Engineering, Michigan State University, East Lansing, MI, 48824, USA.
\IEEEcompsocthanksitem X. Jiang is with the Department of Biomedical Informatics, University of California, San Diego, La Jolla, CA, 92093, USA.
\IEEEcompsocthanksitem F. Wang is with the Department of Healthcare Policy and Research, Weill Cornell Medical College, New York City, NY, 10065, USA.}%
\thanks{Manuscript received April 19, 2005; revised August 26, 2015.}
}

\markboth{IEEE Transactions on Pattern Analysis and Machine Intelligence}%
{Shell \MakeLowercase{\textit{et al.}}: Bare Demo of IEEEtran.cls for Computer Society Journals}

\IEEEtitleabstractindextext{%
\begin{abstract}
Multi-task learning (MTL) refers to the paradigm of learning multiple related tasks together. In contrast, in single-task learning (STL) each individual task is learned independently. MTL often leads to better trained models because they can leverage the commonalities among related tasks. However, because MTL algorithms can ``leak" information from different models across different tasks, MTL poses a potential security risk. Specifically, an adversary may participate in the MTL process through one task and thereby acquire the model information for another task. The previously proposed privacy-preserving MTL methods protect data instances rather than models, and some of them may underperform in comparison with STL methods. In this paper, we propose a privacy-preserving MTL framework to prevent information from each model leaking to other models based on a perturbation of the covariance matrix of the model matrix. We study two popular MTL approaches for instantiation, namely, learning the low-rank and group-sparse patterns of the model matrix. Our algorithms can be guaranteed not to underperform compared with STL methods. We build our methods based upon tools for differential privacy, and privacy guarantees, utility bounds are provided, and heterogeneous privacy budgets are considered. The experiments demonstrate that our algorithms outperform the baseline methods constructed by existing privacy-preserving MTL methods on the proposed model-protection problem.
\end{abstract}

}

\maketitle

\IEEEdisplaynontitleabstractindextext
\IEEEpeerreviewmaketitle

\section{Introduction}\label{sec:introduction}

\IEEEPARstart{M}{ulti-task learning} (MTL)~\citep{caruana1997multitask} refers to the paradigm of learning multiple related tasks together. {In contrast, single-task learning (STL) refers to the paradigm of learning each individual task independently.} MTL often leads to better trained models because the commonalities among related tasks may assist in the learning process for each specific task. For example, an infant's ability to recognize a cat might help in developing the ability to recognize a dog. In recent years, MTL has received considerable interest in a broad range of application areas, including computer vision~\citep{li2010probabilistic,zhang2012robust}, natural language processing~\citep{almeida2013fast} and health informatics~\citep{sun2015linkage,wang2014exploring}. {The key to MTL is to relate learning tasks via a shared representation, which, in turn, benefits the tasks to be learned. Each possible shared representation encodes certain assumptions regarding task relatedness.}
Because MTL approaches explore and leverage the commonalities among related tasks within the learning process, either explicitly or implicitly, they pose a potential security risk.
Specifically, an adversary may participate in the MTL process through a participating task, thereby acquiring the model information for another task. {A predictive model may identify a causality between system inputs and outputs. Knowledge of the causality makes it possible or easier for an adversary to change a system input to trigger an irrational or even harmful output, which can be regarded as a \emph{generalized adversarial attack}. The system could be a predictive system for traffic-sign recognition or face identification, as studied by recent adversarial-attack approaches~\citep{szegedy2013intriguing,moosavi2016deepfool,kurakin2016adversarial,zantedeschi2017efficient,papernot2016practical,papernot2016transferability,mopuri2017fast}.
Pointed out by \citet{finlayson2019adversarial}, adversarial attacks on medical machine learning are increasingly rampant and easy to implement (e.g., by simply rotating a picture to upload to a specific angle), especially on medical fraud which is a \$250 billion industry. Therefore, model-information leakage during an MTL process could realize or escalate such adversarial attacks to increase irrational medical costs or insurances. In addition, the aforementioned system could well be a real human body. For example, consider \emph{personalized predictive modeling}~\citep{ng2015personalized,zhang2014towards}, which has become a fundamental methodology in health informatics. This type of modeling builds a custom-made model for each patient. In modern health informatics, such a model may include patient disease conditions/causes (e.g., which foods can induce an allergic reaction in a patient). If such information were to be leaked, an adversary might use the information to deliberately introduce the food into a patient meal to trigger an allergic reaction, which could be disastrous.}

Because of the concerns discussed above, a secure training strategy must be developed for MTL approaches to prevent information from each model leaking to other models. In this paper, we propose a model-protected multi-task learning (MP-MTL) approach that enables the joint learning of multiple related tasks while simultaneously preventing model leakage for each task. We use tools of \emph{differential privacy}~\citep{dwork2014algorithmic} which provides a strong, cryptographically motivated definition of privacy based on rigorous mathematical theory and has recently received significant research attention due to their robustness to known attacks~\citep{chaudhuri2011differentially,ganta2008composition}. This scheme is useful when one wishes to prevent potential attackers from acquiring information on any element of the input dataset based on a change in the output distribution.

\begin{figure}[t!]
\centering
\includegraphics[width=3.5in]{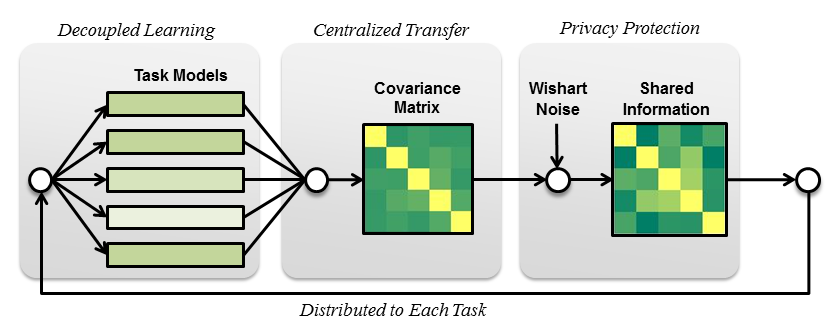}
\caption{{Model-protected multi-task learning framework. The solution process is a recursive two-step procedure. The first step is a decoupled learning procedure in which the model parameters for each task are estimated independently using the precomputed shared information among tasks. The second step is a centralized transfer procedure in which the information shared among tasks is extracted for distribution to each task for the decoupled learning procedure in the next step. The shared information is extracted from the tasks' covariance matrix, into which Wishart noise is introduced to ensure model security.}}\label{fig:framework}
\end{figure}

To focus on the main issue, our MP-MTL method is designed simply based on linear multi-task models~\citep{ji2009accelerated,liu2009multi}. We assume that the model parameters are learned by minimizing an objective that combines an average empirical prediction loss and a regularization term. The regularization term captures the commonalities among the different tasks and couples their model parameters. The solution process for this type of MTL method can be viewed as a recursive two-step procedure. The first step is a decoupled learning procedure in which the model parameters of each task are estimated independently using some precomputed shared information among tasks. The second step is a centralized transfer procedure in which some of the information shared among tasks is extracted for distribution to each task for the decoupled learning procedure in the next step. Our MP-MTL mechanism protects the models by adding perturbations during the second step. {Note that we assume a curator that collects models for joint learning but never needs to collect task data.} We develop a rigorous mathematical definition of the MP-MTL problem and propose an algorithmic framework to obtain the solution. We add perturbations to the covariance matrix of the parameter matrix because the tasks' covariance matrix is widely used as a fundamental source from which to extract useful knowledge to share among tasks~\citep{liu2009multi, zhang2010convex, ji2009accelerated,argyriou2007multi,zhou2011clustered,sun2015linkage}, {which is the key observation that enables the proposed framework.} The usage of the perturbed covariance matrix depends on the specific MTL method applied. Consequently, our technique can cover a wide range of MTL algorithms and {is generally applicable for many optimization schemes, such as proximal gradient methods~\citep{ji2009accelerated,liu2009multi}, alternating methods~\citep{argyriou2007multi} and Frank-Wolfe methods~\citep{jaggi2013revisiting}}. We introduce Wishart noise into the covariance matrix to ensure model security. Fig. \ref{fig:framework} illustrates the key ideas of the main framework.


We further develop two concrete approaches as instantiations of our framework, each of which transforms an existing MTL algorithm into a private version. Specifically, we consider two popular types of basic MTL models: 1) a model that learns a low-rank subspace by means of a trace norm penalty~\citep{ji2009accelerated} and 2) a model that performs shared feature selection by means of a group-$\ell_1$ ($\ell_{2,1}$ norm) penalty~\citep{liu2009multi}.
{We first choose to learn a low-rank subspace of the model matrix because it is typical to learn a shared representation, which is the key to relate tasks in MTL. In addition, it is also typical to learn correlated but different parameters for multiple models that share the same model structure, which is also commonly-encountered in MTL. In fact, it is a typical/mainstream approach in MTL, as stated by \citet{zhang2017survey} in their MTL survey as well as by \citet{su2018multi} and \citet{gu2016low}; (see, e.g.,~\citet{ando2005framework},~\citet{chen2009convex},~\citet{xu2012conditional},~\citet{han2016multi}, and~\citet{zhen2018multi}). On the other hand, learning a shared feature selection is also typical in MTL and can be regarded as learning a specific type of low-rank subspace.
In both cases, we instantiate our framework by approximating \emph{proximal gradient descent methods}, which were presented by~\citet{ji2009accelerated} and~\citet{liu2009multi}. The covariance matrix is used to build a linear transform matrix used to project the models into new feature subspaces; then, the most useful subspaces are selected. The projection matrix is related to the generalized inverse of the singular value matrix (or the diagonal matrix) of the perturbed covariance matrix for the instantiation with a trace norm penalty (or the instantiation with a group-$\ell_1$ penalty). Wishart noise is positive definite; thus, it renders the singular values of the perturbed covariance matrix ``larger'' and those of the generalized inverse ``smaller''. Consequently, under a high noise level, the projection matrix tends to be an identity matrix that shares no information between models but keeps the models intact. This means that our algorithms will not underperform in comparison with STL methods under high noise levels; hence, participation in the joint learning process has no negative effect on training any task model. Approximating the proximal operators with Wishart noise makes it possible for the added noise to destroy the covariance matrix without destroying the projected models, which is a key observation that enables the proposed instantiated algorithms.}

We provide privacy guarantees. Utility analyses are also presented for both convex and strongly convex prediction loss functions and for both the basic and accelerated proximal-gradient methods. Furthermore, we consider heterogeneous privacy budgets for different iterations of our algorithms and present a utility analysis for privacy-budget allocation. We also validate the effectiveness of our approach on both benchmark and real-world datasets.

{Our proposed MP-MTL algorithms fall into a larger scope of differentially private MTL algorithms (i.e., MTL algorithms with randomness added using differential privacy tools). Within this scope, to the best of our knowledge, we are the first to present a utility analysis for heterogeneous tasks. In contrast, the distributed tasks studied by~\citet{pathak2010multiparty} are homogeneous (i.e., the coding procedures for both features and targets are the same for different tasks). To our knowledge, we are also the first to provide privacy guarantees for heterogeneous tasks and for loss functions without closed-form solutions (e.g., logistic loss).}

{Since our method are the first to address the model-protected problem in MTL setting, we construct baseline MP-MTL methods for comparison by exploiting existing privacy-preserving MTL methods, which are referred to as instance-protected multi-task learning (IP-MTL) methods because they protect the security only of data instances rather than that of models. The IP-MTL methods are transformed into their respective MP-MTL methods by directly enforcing the \emph{group privacy}~\citep{dwork2014algorithmic} of the entire dataset coming from a single learning task. The experimental results demonstrate that our proposed algorithms outperform the constructed baseline MP-MTL methods.}

The contributions of this paper are highlighted as follows.
\begin{itemize}
\item {We are the first to propose and address the model-protection problem in an MTL setting.}
\item We develop a general algorithmic framework to solve the MP-MTL problem to obtain secure estimates of the model parameters. We derive concrete instantiations of our algorithmic framework for two popular types of MTL models, namely, models that learn the low-rank and group-sparse patterns of the model matrix. {By approximating the proximal operators with Wishart noise}, we can guarantee that our algorithms will not underperform in comparison with STL methods under high noise levels.
\item We provide privacy guarantees. We also present utility analyses for both convex and strongly convex prediction loss functions and for both the basic and accelerated proximal-gradient methods. Heterogeneous privacy budgets are considered for different iterations of our algorithms, and a utility analysis for privacy-budget allocation is presented.
\item {Within the larger scope of differentially private MTL algorithms, to the best of our knowledge, we are the first to provide privacy guarantees for heterogeneous tasks and for loss functions without closed-form solutions. We are also the first to present utility analyses for heterogeneous tasks and the first to present allocation strategies for heterogeneous privacy budgets.}
\item {For comparison, we construct baseline MP-MTL methods using IP-MTL methods (i.e., existing privacy-preserving MTL methods). The experiments demonstrate that our proposed algorithms significantly outperform the baseline methods.}
\end{itemize}

The remainder of this paper is organized as follows. {Section \ref{sec:related_work} discusses related works and definitions of differential privacy.} Section \ref{sec:pre} introduces the background on MTL problems and the definition of the proposed model-protection problem. The algorithmic framework and concrete instantiations of the proposed MP-MTL method are presented in Section \ref{sec:method}, {along with the analyses of utility and privacy-budget allocation}. Section \ref{sec:exp} presents an empirical evaluation of the proposed approaches, and Section \ref{sec:conclusion} provides conclusions.

\section{Related Works}\label{sec:related_work}

\subsection{Privacy-preserving MTL Approaches}

Few privacy-preserving MTL approaches have been proposed to date~\citep{mathew2012distributed,baytas2016asynchronous,pathak2010multiparty,shokri2015privacy,gupta2016differentially}. Moreover, such approaches protect only the security of data instances rather than that of models. A typical focus of research is distributed learning~\citep{mathew2012distributed,baytas2016asynchronous}, in which the datasets for different tasks are distributively located. The local task models are trained independently using their own datasets before being aggregated and injected with useful knowledge shared across tasks. Such a procedure mitigates the privacy problem by updating each local model independently. However, these methods do not provide theoretical privacy guarantees.

\citet{pathak2010multiparty} proposed a differentially private aggregation (DP-AGGR) method in a distributed learning scheme with privacy guarantees in which they first trained local models distributively and then averaged the models of the tasks before adding noise based on the output perturbation method of~\citep{dwork2006calibrating}. The final solution for each task is the averaged model. However, because this method performs only averaging, it has a limited ability to address {more complicated task relations such as low-rank~\citep{ando2005framework}, group-sparse~\citep{turlach2005simultaneous}, clustered~\citep{gu2009learning} or graph-based~\citep{zhang2010convex} task structures}.

\citet{gupta2016differentially} proposed a differentially private multi-task relationship learning (DP-MTRL) method to transform the multi-task relationship learning proposed by~\citet{zhang2010convex} into a differentially private version. They adopts the output perturbation method. {However, their method requires a closed-form solution (obtained by minimizing the least-square loss function) to achieve a theoretical privacy guarantee; thus, it cannot guarantee privacy for methods such as logistic regression, which requires iterative optimization procedures}. Moreover, their privacy-preserving MTL methods underperformed on their synthetic datasets compared with non-private STL methods (which can guarantee optimal privacy against cross-task information leakage), which suggests that there is no reason to use their proposed methods. {In addition, they did not study the additional privacy leakage that occurs due to the iterative nature of their algorithm (see~\citet{kairouz2017composition}), and they did not study the utility bound for their method}.

The approaches of both~\citet{pathak2010multiparty} and~\citet{gupta2016differentially} protect a single data instance instead of the model of each task, and they involve adding noise directly to the models, which is unnecessary to avoid information leakage across tasks---and may jeopardize the utility of the algorithms.

{

\subsection{Related Works of Differential Privacy}\label{subsec:related_def}

Several related definitions of privacy are listed as follows.

\noindent\textbf{\emph{\underline{Joint differential privacy}}}. In a game theory setting, \citet{kearns2015robust} and \citet{kearns2014mechanism} proposed to guarantee that for each player, the output to other players reveals little input information about that player. Except for the details of game theory, such as player type, this definition can be regarded as identical to the privacy constraint used in our proposed MP-MTL algorithms.

\noindent\textbf{\emph{\underline{One-analyst-to-many-analyst privacy}}}. In a database query setting, \citet{hsu2013differential} proposed a method for protecting the privacy of all the queries of one analyst from other analysts. They also proposed a related privacy called \emph{one-query-to-many-analyst privacy}. Both privacy definitions can be regarded as identical to the privacy constraints used in our proposed MP-MTL and our defined IP-MTL algorithms, respectively, except for the details of database query.

Roughly speaking, the above approaches and this paper adopt the same privacy constraints but work in different specific fields. This paper leverages these concepts to propose a method to handle the model-protection problem in an MTL setting.


\noindent\textbf{\emph{\underline{Differential privacy for streams}}}. This definition considers continual/iterative observations, and was proposed by \citet{chan2011private} and \citet{dwork2010differential}. Because machine learning algorithms are generally iterative, this paper also involves the concept of iteration in the definitions of MP-MTL and IP-MTL algorithms, and it simply aims to directly use composition theorems of differential privacy to bound the total privacy-budgets.

\noindent\textbf{\emph{\underline{Local private learning algorithms}}}. This definition was proposed by \citet{kasiviswanathan2011can} to characterize that each individual's data are added independent randomness before further processing. Algorithms that accomplish this task are referred to as input perturbation. The idea can be adopted to propose possible solutions to the MP-MTL problem. For example, independent randomness can be added to each task model. Both DP-MTRL and DP-AGGR can be regarded as examples, although they protect data instances rather than models. However, because local private learning algorithms have some limitations (e.g., they may require exponentially more data than do general private algorithms ~\citep{kasiviswanathan2011can}), we did not adopt this idea when constructing our MP-MTL method.

\noindent\textbf{\emph{\underline{Secure multi-party computation (SMC)}}}. In Section 1, we assume the use of a trusted curator that collects the task models, and this assumption raises privacy concerns in untrusted curator settings. Such concerns are related to the demand for SMC~\citep{pathak2010multiparty,goldreich1998secure}, {the purpose of which is to avoid the leakage of data instances to the curator}. We present an extended framework that considers SMC in the supplementary material.

In addition to the above related definitions, the sample-aggregate framework proposed by \citet{nissim2007smooth} is also related. This framework first randomly partitions a database into multiple small databases, executes the same algorithms on all the sub-databases, aggregates all the outputs, and finally adds randomness according to the smooth sensitivity of the aggregation function. For model-protection, this framework may be applicable for homogeneous tasks (which is the setting considered by DP-AGGR) to extend their method for empirical risk minimization: instead of data instances, tasks can be randomly partitioned into groups to perform the above procedures. However, applying this framework to heterogeneous tasks is not trivial. The framework may also be applied to improve the utility of DP-AGGR by smooth sensitivity.

}
\subsection{Methods that Privately Release the Covariance Matrix}
Several methods have been proposed to privately release the covariance matrix~\citep{jiang2016wishart,dwork2014analyze,blocki2012johnson}. Considering an additive noise matrix, based on our utility analysis, the overall utility of the MTL algorithm depends on the spectral norm of the noise matrix. A list of the bounds on the spectral norms of additive noise matrices can be found in~\citet{jiang2016wishart}. We choose to add Wishart noise~\citep{jiang2016wishart} to the covariance matrix for four reasons: (1) For a given privacy budget, this type of noise matrix has a better spectral-norm bound than does a Laplace noise matrix~\citep{jiang2016wishart}. (2) Unlike a Gaussian noise matrix, which enables an $(\epsilon,\delta)$-private method with a positive $\delta$, this approach enables an $(\epsilon,0)$-private method and can be used to build an iterative algorithm that is entirely $(\epsilon,0)$-private, which provides a stronger privacy guarantee. (3) Unlike the Gaussian and Laplace noise matrices cases, the Wishart noise matrix is positive definite and can be exploited to guarantee that our method will not underperform compared with STL methods under high noise levels. (4) This approach allows arbitrary changes to any task, unlike the method of~\citet{blocki2012johnson}. %

\section{Preliminaries and the Proposed Problem}\label{sec:pre}

{
In this section, we first introduce the MTL background and then introduce the definition of model-protection problems for MTL.
}

The notations and symbols that will be used throughout the paper are summarized in Table \ref{tab:notations}.

\begin{table}[htbp]
\centering
\caption{Notations and Symbols}
\label{tab:notations}
{\small
\begin{tabular}{l|p{20em}}
\hline\hline $[k]$& the index set $\{1,2,\cdots,k\}$\\
\hline
$[-i]$& the index set with index $i$ removed\\
\hline $\|\cdot\|_{*}$& the trace norm of a matrix (sum of the singular values of the matrix)\\
\hline $\|\cdot\|_{2,1}$& the $\ell_{2,1}$ norm of a matrix (sum of the $\ell_2$ norms of the row vectors of the matrix)\\
\hline ${\rm tr(\cdot)}$& the trace of a matrix (sum of the diagonal elements of the matrix)\\
\hline $\sigma_j(\cdot)$& the $j$-th largest singular value of a matrix, $j\in[m]$\\
\hline\hline
\end{tabular}}
\end{table}

Extensive MTL studies have been conducted on linear models using regularized approaches.
The basic MTL algorithm that we consider in this paper is as follows:
\begin{equation}\label{eq:MTL_ONE}
\widehat{\W} = \arg\min_{\W} \sum\nolimits_{i=1}^m\calL_i(\X_i\w_i,\y_i) + \lambda g(\W),
\end{equation}
where $m$ is the number of tasks. The datasets for the tasks are denoted by $\calD^m = (\X^m,\y^m) = \{(\X_1,\y_1),\ldots,(\X_m,\y_m)\}$, where for each $i\in[m]$, $\calD_i = (\X_i,\y_i)$, where $\X_i\in \mathbb{R}^{n_i \times d}$ and $\y_i\in \mathbb{R}^{n_i \times 1} $ denote the data matrix and target vector of the $i$-th task with $n_i$ samples and dimensionality $d$, respectively. $\calL_i$ is the prediction loss function for the $i$-th task. In this paper, we focus on linear MTL models in which $\w_i$ denotes the model/predictor for task $i$ and
$\W=[\w_1,\w_2,\cdots,\w_m]\in \mathbb{R}^{d \times m}$ is the model parameter matrix. $g(\cdot)$ is a regularization term that represents the structure of the information shared among the tasks, for which $\lambda$ is a pre-fixed hyper-parameter. As a special case, STL can be described by \eqref{eq:MTL_ONE} with $\lambda = 0$.

The key to MTL is to relate the tasks via a shared representation, which, in turn, benefits the tasks to be learned. Each possible shared representation encodes certain assumptions regarding task relatedness.

{A typical/mainstream assumption is that the tasks share a latent low-rank subspace, as stated by \citet{zhang2017survey} in their survey on MTL, and also by \citet{su2018multi} and \citet{gu2016low}; see, e.g.,~\citet{ando2005framework},~\citet{chen2009convex},~\citet{xu2012conditional},~\citet{han2016multi}, and~\citet{zhen2018multi}.} The formulation leads to a low-rank structure of the model matrix. Because optimization problems involving rank functions are intractable, a trace-norm penalty is typically used~\citep{amit2007uncovering, ji2009accelerated, pong2010trace}, as in the following problem, which will be referred to as the \emph{trace-norm-regularized MTL problem}.
\begin{equation}\label{eq:MTL_Trace}
\min_{\W} \ \sum\nolimits_{i=1}^m\calL_i(\X_i\w_i,\y_i) + \lambda \|\W\|_{*}.
\end{equation}

Another typical assumption is that all tasks share a subset of important features. Such task relatedness can be captured by imposing a group-sparse penalty on the predictor matrix to select shared features across tasks~\citep{turlach2005simultaneous,wright2009sparse,liu2009multi}. One commonly used group-sparse penalty is the group $\ell_1$ penalty~\citep{liu2009multi,nie2010efficient}, as in the following problem, which will be referred to as the \emph{group-$\ell_1$-regularized MTL problem}.
\begin{equation}\label{eq:MTL_L21}
\min_{\W} \ \sum\nolimits_{i=1}^m\calL_i(\X_i\w_i,\y_i) + \lambda \|\W\|_{2,1}.
\end{equation}

Next, we present a compact definition of the model-protection problem in the context of MTL and discuss the general approach without differential privacy. As \eqref{eq:MTL_ONE} shows, as a result of the joint learning process, $\widehat{\w}_j$ may contain some information on $\widehat{\w}_i$, for $i,j\in[m]$ and $i\neq j$, making it possible for the owner of task $j$ to use such information to attack task $i$. Thus, we define the model-protection problem as follows.
\begin{definition}[Model-protection Problem for MTL]\label{th:df_MP-MTL_problem}
The model-protection problem for MTL has three objectives:

1) to minimize the information on $\widehat{\w}_i$ that can be inferred from $\widehat{\w}_{[-i]}$, for all $i\in[m]$;

2) to maximize the prediction performance of $\widehat{\w}_i$, for all $i\in[m]$; and

3) to share useful predictive information among tasks.
\end{definition}

Now, consider such a procedure in which a trusted curator collects independently-trained models, denoted by $\w_1,\ldots,\w_m$, for all tasks \emph{without their associated data} to be used as input. After the joint learning process, the curator outputs the updated models, denoted by $\widehat{\W}$, and sends each updated model to each task privately. The model collection and joint learning processes are performed alternately.

{
We note that the \emph{trace-norm-regularized MTL problem} and the \emph{group-$\ell_1$-regularized MTL problem} are unified in the multi-task
feature learning framework, which is based on the covariance matrix of the tasks' predictors~\citep{argyriou2007spectral,evgeniou2007multi,argyriou2008convex}. Many other MTL methods also fall under this framework, such as learning clustered structures among tasks~\citep{gu2009learning,zhou2011clustered} and inferring task relations~\citep{zhang2010convex,fei2013structured,bonilla2007multi}. As such, we note that the tasks' covariance matrix constitutes a major source of shared knowledge in MTL methods; hence, it is regarded as the primary target for model protection.

Therefore, we address the model-protection problem by rephrasing the first objective in Definition \ref{th:df_MP-MTL_problem} as follows: to minimize the changes in $\widehat{\w}_{[-i]}$ and the tasks' covariance matrix ($\widehat{\W}\widehat{\W}\trans$ or $\widehat{\W}\trans\widehat{\W}$) when task $i$ participates in the joint learning process for all $i\in[m]$. Thus, the model for this new task is protected.

Then, we find that the concept of differential privacy (minimizing the change in the output distribution) can be adopted to further rephrase this objective as follows: to minimize the changes in \emph{the distribution of} $\widehat{\w}_{[-i]}$ and the tasks' covariance matrix when task $i$ participates in the joint learning process for all $i\in[m]$.
}

In differential privacy, algorithms are randomized by introducing some type of perturbation.
\begin{definition}[Randomized Algorithm]\label{th:df_rand_alg}
{A randomized algorithm} $\mathcal{A}: \calD \rightarrow \theta \in \calC$ is {built by introducing some type of perturbation into} some mapping $\calD \rightarrow \theta \in \calC$. Algorithm $\mathcal{A}$ outputs $\mathcal{A}(\calD) = \theta$ with a density $p(\mathcal{A}(\calD) = \theta)$ for each $\theta\in \calC$. The probability space is over the perturbation introduced into algorithm $\mathcal{A}$.
\end{definition}
In this paper, $\mathcal{A}$ denotes some randomized machine learning estimator, and $\mbox{\boldmath$\theta$}$ denotes the model parameters that we wish to estimate. Perturbations can be introduced into the original learning system via the (1) input data~\citep{Li2014,BCDKMT07}, (2) model parameters~\citep{chaudhuri2008privacy,Ji2014a}, (3) objective function~\citep{CMS11,Zhang2012}, or (4) optimization process~\citep{Song2013,wang2015privacy}.

The formal definition of differential privacy is as follows.

\begin{definition}[\citet{dwork2014algorithmic}]\label{th:df_dp_ori}
A randomized algorithm $\mathcal{A}$ provides $(\epsilon, \delta) $-differential privacy if, for any two adjacent datasets $\mathcal{D}$ and $\mathcal{D}'$ that differ by a single entry and for any set $\mathcal{S}$,
\begin{align*}
\mathbb{P}(\mathcal{A}(\mathcal{D})\in \mathcal{S})
\leq \exp(\epsilon)\mathbb{P}(\mathcal{A}(\mathcal{D}')\in \mathcal{S}) + \delta,
\end{align*}
where $\mathcal{A}(\mathcal{D})$ and $\mathcal{A}(\mathcal{D}')$ are the outputs of $\mathcal{A}$ on the inputs $\mathcal{D}$ and $ \mathcal{D}' $, respectively.
\end{definition}

The privacy loss pair $(\epsilon,\delta)$ is referred to as the privacy budget/loss, and it quantifies the privacy risk of algorithm $\mathcal{A}$. The intuition is that it is difficult for a potential attacker to infer whether a certain data point has been changed in (or added to) the dataset $\mathcal{D}$ based on a change in the output distribution. Consequently, the information of any single data point is protected.

Furthermore, note that differential privacy is defined in terms of application-specific adjacent input databases. In our setting, these are each task's model and dataset pair, which are treated as a ``single entry'' by Definition \ref{th:df_dp_ori}.

Several mechanisms exist for introducing a specific type of perturbation. A typical type is calibrated to the sensitivity of the original ``unrandomized'' machine learning estimator $f: \mathcal{D}\rightarrow \ttheta\in \mathbb{R}^d$. The sensitivity of an estimator is defined as the maximum change in its output due to a replacement of any single data instance. %

\begin{definition}[\citet{dwork2014algorithmic}]
The sensitivity of a function $f: \mathcal{D}\rightarrow \mathbb{R}^d$ is defined as
\begin{align*}
S(f) = \max_{\mathcal{D},\mathcal{D}'}\|f(\mathcal{D}) -f(\mathcal{D}') \|
\end{align*}
for all datasets $\mathcal{D}$ and $\mathcal{D}'$ that differ by at most one instance, where $\|\cdot \|$ { is specified by a particular mechanism. For example, the Gaussian mechanism~\citep{dwork2014analyze} requires the $\ell_2$ norm, and the Laplace mechanism~\citep{dwork2014algorithmic} requires the $\ell_1$ norm.}
\end{definition}

The use of additive noise such as Laplace ~\citep{dwork2014algorithmic} or Gaussian noise~\citep{dwork2014analyze} with a standard deviation proportional to $S(f)$ is a common practice for guaranteeing private learning. {In this paper, we adopt the Wishart noise for covariance matrices~\citep{jiang2016wishart}, which is defined as follows.
\begin{definition}[\citet{gupta1999matrix}]
A $d \times d$ random symmetric positive definite matrix $\E$ is said to have a
Wishart distribution $\E\sim W_d(\nu,\V)$ if its probability density function is
\begin{equation*}
p(\E) = \frac{|\E|^{(\nu-d-1)/2}\exp(-\rm tr(\V^{-1}\E)/2)}{2^{\frac{\nu d}{2}}|\V|^{1/2}\Gamma_d(\nu/2)},
\end{equation*}
where $\nu > d-1$ and $\V$ is a $d \times d$ positive definite matrix.
\end{definition}
}

Because machine learning schemes are usually presented as sequential paradigms with multiple iterations and usually output multiple variables simultaneously, several differential privacy properties are particularly useful for ensuring privacy in machine learning, such as post-processing immunity, group privacy, composition properties and adaptive composition. The details of these properties are introduced in the supplementary material.

\section{Methodology}\label{sec:method}

We present our methodology in this section: the modeling of and {rationale} for our MP-MTL framework, two instantiations and utility analyses. Regarding the theoretical results, we present only the main results; the detailed derivations are included in the provided supplementary material.

\subsection{The General MP-MTL Framework}
Consider an MTL algorithm $\calA$ with $T$ iterations. For $t = 1,\ldots,T$, a trusted curator collects the models of $m$ tasks, respectively, denoted by $\w_1^{(t-1)},\ldots,\w_m^{(t-1)}$. Then, a model-protection and shared-information-extraction procedure is performed, and the updated models $\hat{\w}_1^{(t)},\ldots,\hat{\w}_m^{(t)}$ are output and sent back to their respective tasks.
\begin{remark}\label{th:rm_nodata}
{Note that in each iteration, the curator collects only the models. The dataset for each task can be regarded as the input for the entire MTL algorithm, but it is not the input for the curator.}
\end{remark}

In such a setting, for each $i\in[m]$, we wish to protect the dataset $\calD_i = (\X_i,\y_i)$ of task $i$ and its \emph{entire input model-sequence} $(\w_{i}^{(0)},...,\w_{i}^{(T-1)})$ (denoted by $\w_{i}^{(0:T-1)}$ for short). {\emph{For the $i$-th task}, the \emph{entire output model-sequence} of other tasks, $\hat{\w}_{[-i]}^{(1:T)}$, is the \emph{view} of a potential adversary (i.e., the information that the adversary can acquire to infer the unique information of task $i$). Note that although the output model-sequence of each task is what we ultimately wish to protect, the unique information within each task is contained in the task's dataset and input model-sequence, which are actually protected.}

The idea for using differental privacy tools is as follows.
{For simpilicity, we assume that $T=1$ and omit the iteration-step indices.} Let $\widetilde{\calD} = \{(\w_1, \calD_1),\ldots,(\w_m, \calD_m)\}$ be an augmented dataset; i.e., let $(\w_i, \calD_i)$ be treated as the $i$-th ``data instance'' of the augmented dataset $\widetilde{\calD}$, for all $i\in[m]$. Thus, the $m$ datasets and $m$ models associated with the $m$ tasks are transformed into a single dataset $\widetilde{\calD}$ with $m$ ``data instances''. Then, we define $m$ outputs $\ttheta = (\theta_1,\ldots,\theta_m)$ such that for each $i\in[m]$, $\theta_i\in\calC_i$ denotes the \emph{view} of an adversary for task $i$, which includes $\hat{\w}_{[-i]}$. Thus, an $(\epsilon,\delta)$ - MP-MTL algorithm $\calA(\calB)$ should satisfy the following $m$ inequalities.
For each $i\in [m]$, for all neighboring datasets $\widetilde{\calD}$ and $\widetilde{\calD}'$ that differ by the $i$-th ``data instance'', and for any set $\calS_i\subseteq C_i$, we have
\begin{equation}\label{eq:MP_MTL-simple-dp}
\begin{split}
\bbP&( \theta_i \in \calS_i \mid \calB = \widetilde{\calD}) \leq e^{\epsilon} \bbP(\theta_i \in \calS_i \mid \calB=\widetilde{\calD}') + \delta.
\end{split}
\end{equation}

We formally define an MP-MTL algorithm as follows.
\begin{definition}[MP-MTL]\label{th:df_MP-MTL-iter}
Let $\calA$ be a randomized MTL algorithm with a number of iterations $T$. In the first iteration, $\calA$ performs the mapping $(\W^{(0)} \in\mathbb{R}^{d\times m},\calD^m)\rightarrow \theta_1\in \calC_1$, where $\theta_1$ includes $\widehat{\W}^{(1)} \in\mathbb{R}^{d\times m}$. For $t=2,\ldots,T$, in the $t$-th iteration, $\calA$ performs the mapping $(\W^{(t-1)} \in\mathbb{R}^{d\times m},\calD^m,\theta_1,\ldots,\theta_{t-1})\rightarrow \theta_t\in \calC_t$, where $\theta_t$ includes $\widehat{\W}^{(t)} \in\mathbb{R}^{d\times m}$. Then, $\calA$ is an \emph{$(\epsilon,\delta)$ - MP-MTL algorithm} if for all $i\in[m]$, for all $t\in[T]$, and for neighboring input pairs $(\W^{(t-1)},\calD^m)$ and $((\W')^{(t-1)},(\calD')^m)$ that differ only by the $i$-th task such that $\w_i^{(t-1)}\neq (\w'_i)^{(t-1)}$ or $\calD_i\neq\calD'_i$, the following holds for some constants $\epsilon,\delta\geq 0$ and for any set $\calS \subseteq \mathbb{R}^{d\times (m-1)\times T}$:
\begin{equation}\label{eq:MP_MTL-iter}
\begin{split}
\bbP( & \hat{\w}_{[-i]}^{(1:T)} \in \calS \mid \bigcap_{t=1}^T \calB_t = (\W^{(t-1)}, \calD^m, \ttheta_{1:t-1}) ) \\
\leq &e^{\epsilon} \bbP( \hat{\w}_{[-i]}^{(1:T)} \in \calS \mid \bigcap_{t=1}^T \calB_t = ((\W')^{(t-1)}, (\calD')^m, \ttheta_{1:t-1})) \\
&+ \delta,
\end{split}
\end{equation}
where for all $t\in[T]$, $\calB_t$ denotes the input for the $t$-th iteration and
\begin{numcases}{\ttheta_{1:t-1} = }
\nonumber \emptyset, & $t=1$\\
\nonumber \theta_1,\theta_2,\cdots,\theta_{t-1}, & $t\geq2$.
\end{numcases}
\end{definition}

{

Note that in Definition \ref{th:df_MP-MTL-iter}, we view the model sequence $\hat{\w}_{[-i]}^{(1:T)}$ as a single output of the algorithm for each task $i\in[m]$. The definition of neighboring inputs allows the model and dataset for any task to change in \emph{all rounds} of the iterative optimization rather than in only a single round.
Definition \ref{th:df_MP-MTL-iter} defines a privacy constraint for MTL algorithms. Roughly speaking, this privacy constraint can be regarded as identical to \emph{joint differential privacy} \citep{kearns2015robust,kearns2014mechanism} or to \emph{one-analyst-to-many-analyst privacy} \citep{hsu2013differential}, despite detailed differences between MTL, game theory, and database queries.}

STL can easily be shown to be optimal for avoiding information leakage across tasks because the individual task models are learned independently.\vspace{-0.5em}
{
\begin{claim}\label{th:lem_STL}
Any STL algorithm that learns each task independently is a $(0,0)$ - MP-MTL algorithm.
\end{claim}\vspace{-0.5em}}
From this claim, we learn that when no information sharing occurs across tasks, no leakage can occur across tasks.

Our MP-MTL framework is elaborated in Algorithm \ref{alg:MP-MTL}, which considers heterogeneous privacy budgets for different iteration steps. { To maintain the total privacy budget below than a specified value using the adaptive composition theorem provided by~\citet{kairouz2017composition}, we define a composition bound of a series of privacy budgets as follows (the equation is taken directly from Theorem 3.5 of~\citet{kairouz2017composition}):

\begin{definition}[Composition Bound of Privacy Budgets]\label{th:df_com_bound}
For an integer $T\geq1$, a series of privacy budgets, $\epsilon_1,\ldots,\epsilon_T\geq0$, a specified privacy loss $\delta\geq 0$, the composition bound of $\{\epsilon_t\}$ is defined as $CB(\{\epsilon_t\},\delta)$, which equals
\begin{equation}\label{eq:set_eps}
\begin{split}
&\min\biggl\{ \sum_{t=1}^{T}\epsilon_t,\sum_{t=1}^{T}\frac{(e^{\epsilon_t}-1)\epsilon_t}{(e^{\epsilon_t}+1)}
+\sqrt{\sum_{t=1}^{T}2\epsilon_t^2\log\biggl(\frac{1}{\delta}\biggr)},\\
&\sum_{t=1}^{T}\frac{(e^{\epsilon_t}-1)\epsilon_t}{(e^{\epsilon_t}+1)}
+\sqrt{\sum_{t=1}^{T}2\epsilon_t^2\log\biggl(e+\frac{\sqrt{\sum_{t=1}^{T}\epsilon_t^2}}{\delta}\biggr)}
\biggr\}.
\end{split}
\end{equation}
\end{definition}

}

In Algorithm \ref{alg:MP-MTL}, as mentioned in Section \ref{sec:pre}, we choose to protect the tasks' covariance matrix, which is denoted by $\SSigma= \W\W\trans$ or $\SSigma = \W\trans\W$, depending on the MTL method selected. As previously stated, Wishart noise~\citep{jiang2016wishart} is added. Fig. \ref{fig:framework} illustrates the key concepts of the framework. {In detail, Step 1 of Algorithm \ref{alg:MP-MTL} ensures that the total privacy budgets satisfy the specified values $\epsilon$ and $\delta$, respectively. The purpose of norm clipping in Step 3 is simply to render the models in a bounded space, which help us compute a proper noise scale to add to satisfy the privacy constraint defined in Definition \ref{th:df_MP-MTL-iter}. Step 4 extracts the shared information between tasks---the tasks' covariance matrix. Step 5 adds a perturbation into the shared information. Step 6 further extracts useful information from the perturbed covariance matrix. Step 7 sends the extracted useful information to each task to perform decoupled learning. If no noise is added, Steps 4--7 could be a proximal gradient descent step, i.e., first performing a proximal operator step and then taking a gradient descent step; see, e.g.,~\citet{ji2009accelerated} and \citet{liu2009multi}.} This framework is generally applicable for many optimization schemes, such as proximal gradient methods~\citep{ji2009accelerated,liu2009multi}, alternating methods~\citep{argyriou2007multi} and Frank-Wolfe methods~\citep{jaggi2013revisiting}.

{Note that we mainly provided theoretical and experimental results for the $\W\W\trans$ type of covariance matrix. Nonetheless, the $\W\trans\W$ type of covariance matrix can be regarded as a natural alternative, since it was successfully used by \citet{zhang2010convex} (as a non-private method), \citet{gupta2016differentially} (as a differentially private method), and all the subsequent MTL methods to learn relationships between tasks (see, e.g., the task-relation learning approaches introduced by \citet{zhang2017survey} in Section 2.4 of their survey on MTL). Therefore, we included it in our framework. }
{
\begin{remark}
In Algorithm \ref{alg:MP-MTL}, a curator who collects models and performs centralized transfer needs to run only Steps 4--6 and does not need to collect the datasets $(\X^m,\y^m)$, which are used only in STL algorithms.
\end{remark}
}

\begin{algorithm} [htb]
\caption{MP-MTL framework}
\label{alg:MP-MTL}{\small
\begin{algorithmic}[1]
\REQUIRE{Datasets $(\X^m,\y^m) = \{(\X_1,\y_1),\ldots,(\X_m,\y_m)\}$, where $\forall i\in[m], \ \X_i\in \mathbb{R}^{n_i \times d}$ and $\y_i\in \mathbb{R}^{n_i \times 1}$. Privacy loss $\epsilon , \delta \geq 0$. Number of iterations $T$. Initial shared information matrix $\M^{(0)}$. Initial task models $\W^{(0)}$, which can be acquired via arbitrary STL methods.}
\ENSURE{$\widehat{\W}^{(1:T)}$ .}
\STATE Set $\{\epsilon_t\}$ such that { $CB(\{\epsilon_t\},\delta)\leq \epsilon$, where $CB(\{\epsilon_t\},\delta)$ is the composition bound of $\{\epsilon_t\}$.}
\FOR{$t=1:T$}
{
\STATE Norm clipping: $\tilde{\w}_i^{(t-1)} = {\w}_i^{(t-1)}/\max(1,\frac{\|{\w}_i^{(t-1)}\|_2}{K})$, for all $i\in[m]$. Let $\widehat{\W}^{(0)} = \widetilde{\W}^{(0)}$.
}
\STATE $\widetilde{\SSigma}^{(t)} = \widetilde{\W}^{(t-1)}(\widetilde{\W}^{(t-1)})\trans$ (or $\widetilde{\SSigma}^{(t)} = (\widetilde{\W}^{(t-1)})\trans\widetilde{\W}^{(t-1)}$).\label{eq:proximal_op_start}\\
\STATE $\SSigma^{(t)} = \widetilde{\SSigma}^{(t)} + \E$, where $\E\sim W_d(d+1,\frac{K^2}{2\epsilon_t}\I_d)$ (or $\E\sim W_m(m+1,\frac{K^2}{2\epsilon_t}\I_m)$) is a sample from the Wishart distribution, $\I_d$ denotes the $d\times d$ identity matrix, and $\mbox{diag}(\cdot)$ transforms a vector into a diagonal matrix. \label{eq:add_wishart}\\
\STATE Perform an arbitrary mapping $f:\SSigma^{(1:t)}\rightarrow \M^{(t)}$, e.g., take the diagonal elements of $\SSigma^{(t)}$ or the singular value decomposition of $\SSigma^{(t)}$. \label{eq:proximal_op_end}\\
\STATE $\hat{\w}_i^{(t)}= \calA_{\mbox{st},i}(\M^{(t)}, \tilde{\w}_i^{(0:t-1)},\X_i,\y_i)$, for all $i\in[m]$, where $\calA_{\mbox{st},i}$ is an arbitrary STL algorithm for the $i$-th task and the $\tilde{\w}_i^{(0:t-1)}$ are used for initialization. \label{eq:STL_general} \\
\STATE Set the input for the next iteration: $\W^{(t)} = \widehat{\W}^{(t)}$.
\ENDFOR
\end{algorithmic}}
\end{algorithm}

\subsection{Instantiations of the MP-MTL Framework}
In this section, we instantiate our MP-MTL framework (described in Algorithm \ref{alg:MP-MTL}) by approximating the \emph{proximal gradient descent methods} presented by~\citet{ji2009accelerated} and~\citet{liu2009multi} for the \emph{trace-norm-regularized MTL problem} and the \emph{group-$\ell_1$-regularized MTL problem}, respectively.
{Both proximal gradient descent methods solve the respective MTL problems by alternately performing a proximal operator step and a gradient descent step. Taking the \emph{trace-norm-regularized MTL problem} as an example, the loss function, $\sum_i\calL_i$, is minimized by the gradient descent steps, while the regularization term, the trace-norm, is minimized by the proximal operator steps. The proximal operator minimizes the regularization term, keeping the variable near the result of a previous gradient descent step.
Specifically, we instantiate Steps 4--7 of Algorithm \ref{alg:MP-MTL} by approximating a proximal gradient descent step, i.e., first performing a proximal operator step and then taking a gradient descent step. It is similar for the \emph{group-$\ell_1$-regularized MTL problem} but the difference lies in the instantiations of Step 6 of Algorithm \ref{alg:MP-MTL} because different regularization terms lead to different optimal solutions for the proximal operators. Note that both instantiations use the $\W\W\trans$ type of covariance matrix, which is required by the optimal solutions~\citep{ji2009accelerated,liu2009multi}. }

First, we instantiate the MP-MTL framework for the \emph{trace-norm-regularized MTL problem}, as shown in Algorithm \ref{alg:MP-MTL-LR}. Generally speaking, the algorithm uses an accelerated proximal gradient method. Steps \ref{eq:prox_start_LR}--\ref{eq:feature_tran-LR} approximate the following proximal operator~\citep{ji2009accelerated}:
\begin{align}\label{eq:proximal_op_LR}
\widehat{\W}^{(t-1)} = \arg\min_{\W} \frac{1}{2\eta}\|\W - \widetilde{\W}^{(t-1)}\|_F^2 + \lambda\|\W\|_* ,
\end{align}
{where $\widetilde{\W}^{(t-1)}$ can be regarded as the result of the gradient descent step in the previous iteration, assuming $K$ is sufficiently large. In detail, Steps 6--8 of Algorithm \ref{alg:MP-MTL-LR} instantiate Step 6 of Algorithm \ref{alg:MP-MTL} by constructing a projection matrix, $\M^{(t)} = \U\S_{\eta\lambda}\U\trans$, from the result of singular vector decomposition of the perturbed covariance matrix. Steps 9--11 of Algorithm \ref{alg:MP-MTL-LR} instantiate Step 7 of Algorithm \ref{alg:MP-MTL} by first projecting the models (in Step 9) and then performing accelerated gradient descent.}
{
\begin{figure}[!t]\small
\centering
\subfigure[Before MTL]{\includegraphics[width=1.1in]{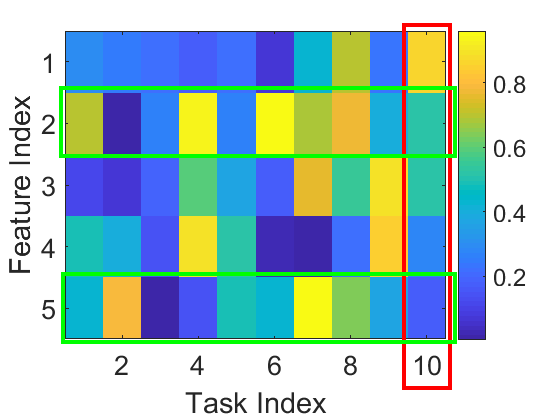}} %
\subfigure[After MTL with model leakage]{\includegraphics[width=1.1in]{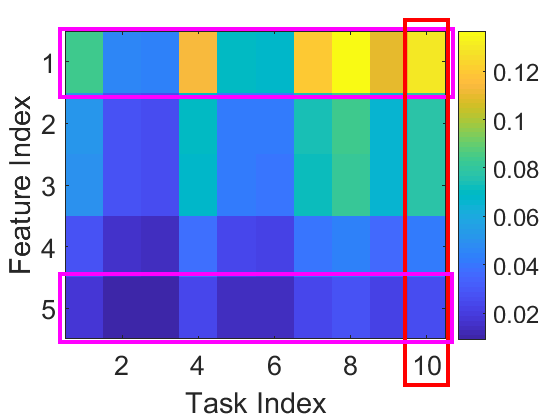}} %
\subfigure[After MTL with model protection]{\includegraphics[width=1.1in]{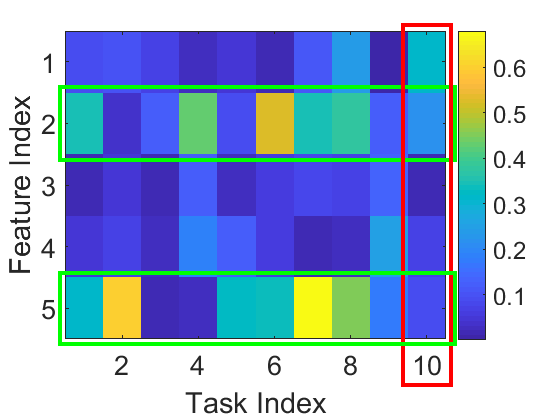}} %
\caption{Examples of model leakage and model protection showing model matrices, where columns correspond to tasks and rows correspond to features. The columns shown have been divided by their respective $\ell_2$ norms.}\label{fig:privacy_leakage}
\end{figure}
}

We provide a running example for model leakage and model protection under different settings of Algorithm \ref{alg:MP-MTL-LR}, as shown in Fig.~\ref{fig:privacy_leakage}. { We generate models for $m=10$ tasks, in which the data dimension is $d=5$. The 10th task (the rightmost one), is an anomaly task that requires privacy protection. In Fig.~\ref{fig:privacy_leakage} (a), the matrix denoted by $\W^{(0)}$ is first generated from an i.i.d. uniform distribution $\mathcal{U}(0,1)$. Then, the rightmost column is multiplied by $100$. For MTL with model leakage, we execute Algorithm \ref{alg:MP-MTL-LR}, setting $T=1, \eta=1,\epsilon_1 = \epsilon = 1e40, \delta = 0, K = 100\sqrt{5}$ and $\lambda = 50$. It can be regarded that the noise matrix $\E$ is not added, since $\epsilon \gg 1$. The output model matrix $\widehat{\W}^{(1)}$ is shown in Fig.~\ref{fig:privacy_leakage} (b), in which the 10th task results in significantly influences on the parameters of other models: other models' parameters are similar to those of the 10th task, e.g., for each task, the first feature is the biggest, and the fifth feature is the smallest. For MTL with model protection, we execute Algorithm \ref{alg:MP-MTL-LR} with the same setting as above except that we set $\epsilon_1 = \epsilon = 0.1$. The output model matrix $\widehat{\W}^{(1)}$ is shown in Fig.~\ref{fig:privacy_leakage} (c), in which the influences from the 10th task are not significant: other models' parameters are not similar to those of the 10th task. Meanwhile, for $\W^{(0)}$, shown in Fig.~\ref{fig:privacy_leakage} (a), for tasks 1-9, the $\ell_2$ norms of the second and the fifth rows are the two largest ones; these are clearly shown in Fig.~\ref{fig:privacy_leakage} (c). This result means the shared information between tasks is to use the second and the fifth features, which is successfully extracted by the MTL method with model protection.}

\begin{algorithm} [htb]
\caption{MP-MTL Low-rank Estimator}
\label{alg:MP-MTL-LR}{\small
\begin{algorithmic}[1]
\REQUIRE{Datasets $(\X^m,\y^m) = \{(\X_1,\y_1),\ldots,(\X_m,\y_m)\}$, where $\forall i\in[m], \ \X_i\in \mathbb{R}^{n_i \times d}$ and $\y_i\in \mathbb{R}^{n_i \times 1}$. Privacy loss $\epsilon , \delta \geq 0$. Number of iterations $T$. Step size $\eta$. Regularization parameter $\lambda>0$. Norm clipping parameter $K>0$. Acceleration parameters $\{\beta_t\}$. Initial task models $\W^{(0)}$.}
\ENSURE{$\widehat{\W}^{(1:T)}$ .}
\STATE Set $\{\epsilon_t\}$ such that { $CB(\{\epsilon_t\},\delta)\leq \epsilon$, where $CB(\{\epsilon_t\},\delta)$ is the composition bound of $\{\epsilon_t\}$.}
\FOR{$t=1:T$}
\STATE Norm clipping: $\tilde{\w}_i^{(t-1)} = {\w}_i^{(t-1)}/\max(1,\frac{\|{\w}_i^{(t-1)}\|_2}{K})$, for all $i\in[m]$. Let $\widehat{\W}^{(0)} = \widetilde{\W}^{(0)}$.
\STATE { $\widetilde{\SSigma}^{(t)} = \widetilde{\W}^{(t-1)}(\widetilde{\W}^{(t-1)})\trans$.}\label{eq:prox_start_LR}\\
\STATE $\SSigma^{(t)} = \widetilde{\SSigma}^{(t)} + \E$, where $\E\sim W_d(d+1,\frac{K^2}{2\epsilon_t}\I_d)$ is a sample from the Wishart distribution.\\
\STATE Perform singular vector decomposition: $\U\LLambda\U\trans = {\SSigma}^{(t)}$.\label{eq:trans_LR} %
\STATE Let $ \S_{\eta\lambda} $ be a diagonal matrix, and let $\S_{\eta\lambda,ii} = \max\{0, 1- \eta\lambda/\sqrt{\LLambda_{ii}} \}$, for $i=1,\ldots,\min\{d,m\}$.\\
\STATE {$\M^{(t)} = \U\S_{\eta\lambda}\U\trans$.}
\STATE Let $ \hat{\w}_i^{(t)} = \M^{(t)}\tilde{\w}_i^{(t-1)}$, for all $i\in[m]$.\label{eq:feature_tran-LR}\\
\STATE {Let $ {\z}_i^{(t)} = \hat{\w}_i^{(t)} + \beta_t (\hat{\w}_i^{(t)} - \hat{\w}_i^{(t-1)})$, for all $i\in[m]$.}\label{eq:acc-LR} \\
\STATE Let ${\w}_i^{(t)} = {\z}_i^{(t)} - \eta \frac{\partial \calL_i(\X_i{\z}_i^{(t)},\y_i)}{\partial {\z}_i^{(t)}}$, for all $i\in[m]$. \label{eq:gd-LR}\\
\ENDFOR
\end{algorithmic}}
\end{algorithm}

Second, we instantiate the MP-MTL framework for the \emph{group-$\ell_1$-regularized MTL problem} defined in \eqref{eq:MTL_L21}, as shown in Algorithm \ref{alg:MP-MTL-GS}. Steps \ref{eq:prox_start_GS}--\ref{eq:feature_select} approximate the following proximal operator~\citep{liu2009multi}:
\begin{align}\label{eq:proximal_op_GS}
\widehat{\W}^{(t-1)} = \arg\min_{\W} \frac{1}{2\eta}\|\W - \widetilde{\W}^{(t-1)}\|_F^2 + \lambda\|\W\|_{2,1}.
\end{align}
{The only difference between Algorithm \ref{alg:MP-MTL-GS} and Algorithm \ref{alg:MP-MTL-LR} is the way they obtain the projection matrix $\M^{(t)}$ for the models (see the differences between Steps 6--8 of Algorithm \ref{alg:MP-MTL-LR} and Steps 6--7 of Algorithm \ref{alg:MP-MTL-GS}). Because Algorithm \ref{alg:MP-MTL-GS} minimizes the group-sparse penalty, it focuses on only the diagonal elements of the perturbed covariance matrix.
}

The error bounds for the proximal operator approximations are provided in Section \ref{sec:utility}.

\begin{algorithm} [htb]
\caption{MP-MTL Group-sparse Estimator}
\label{alg:MP-MTL-GS}{\small
\begin{algorithmic}[1]
\REQUIRE{Datasets $(\X^m,\y^m) = \{(\X_1,\y_1),\ldots,(\X_m,\y_m)\}$, where $\forall i\in[m], \ \X_i\in \mathbb{R}^{n_i \times d}$ and $\y_i\in \mathbb{R}^{n_i \times 1}$. Privacy loss $\epsilon , \delta \geq 0$. Number of iterations $T$. Step size $\eta$. Regularization parameter $\lambda>0$. Norm clipping parameter $K>0$.
Acceleration parameters $\{\beta_t\}$. Initial task models $\W^{(0)}$.}
\ENSURE{$\widehat{\W}^{(1:T)}$ .}
\STATE Set $\{\epsilon_t\}$ such that { $CB(\{\epsilon_t\},\delta)\leq \epsilon$, where $CB(\{\epsilon_t\},\delta)$ is the composition bound of $\{\epsilon_t\}$.}
\FOR{$t=1:T$}
\STATE {Norm clipping: $\tilde{\w}_i^{(t-1)} = {\w}_i^{(t-1)}/\max(1,\frac{\|{\w}_i^{(t-1)}\|_2}{K})$, for all $i\in[m]$.} Let $\widehat{\W}^{(0)} = \widetilde{\W}^{(0)}$.
\STATE $\widetilde{\SSigma}^{(t)} = \widetilde{\W}^{(t-1)}(\widetilde{\W}^{(t-1)})\trans$.\label{eq:prox_start_GS} \\
\STATE $\SSigma^{(t)} = \widetilde{\SSigma}^{(t)} + \E$, where $\E\sim W_d(d+1,\frac{K^2}{2\epsilon_t}\I_d)$ is a sample of the Wishart distribution.\\
\STATE Let $ \S_{\eta\lambda} $ be a diagonal matrix, where for $i=1,\ldots,d$, $\S_{\eta\lambda,ii} = \max\{0, 1 - \eta\lambda/ \sqrt{|{\SSigma}_{ii}^{(t)}|} \}$. \label{eq:construct_Mt}
\STATE {$\M^{(t)} = \S_{\eta\lambda}$.}
\STATE Let $\hat{\w}_i^{(t)} = \M^{(t)} \tilde{\w}_i^{(t-1)}$, for all $i\in[m]$.\label{eq:feature_select}\\
\STATE {Let $ {\z}_i^{(t)} = \hat{\w}_i^{(t)} + \beta_t (\hat{\w}_i^{(t)} - \hat{\w}_i^{(t-1)})$, for all $i\in[m]$.}\label{eq:acc_select} \\
\STATE Let $ {\w}_i^{(t)} = {\z}_i^{(t)} - \eta \frac{\partial \calL_i(\X_i {\z}_i^{(t)},\y_i)}{\partial {\z}_i^{(t)}}$, for all $i\in[m]$. \label{eq:gd}
\ENDFOR
\end{algorithmic}}
\end{algorithm}

We use the following result to show that under high noise levels, our algorithms share no information between models but keep the models intact; thus, they degrade to STL methods but in such a way they do not underperform compared with STL methods.

\begin{proposition}\label{th:prop_better_than_STL}
For Algorithm \ref{alg:MP-MTL-LR}, the projection matrix $\U\S_{\eta\lambda}\U\trans$ degrades to an identity matrix, and the algorithm degrades to an STL algorithm with no random perturbation if the smallest singular value of $\E$ satisfies $\sigma_d(\E)= C \lambda^2$ for a sufficiently large $C>0$.

For Algorithm \ref{alg:MP-MTL-GS}, the projection matrix $\S_{\eta\lambda}$ degrades to an identity matrix, and the algorithm degrades to an STL algorithm with no random perturbation if the smallest diagonal element of $\E$ satisfies $\min_j\E_{jj}= C \lambda^2$ for sufficiently large $C>0$.
\end{proposition}

We also consider other complex MTL frameworks for instantiation. For example, \citet{gong2012robust},~\citet{chen2011integrating},~\citet{jalali2010dirty} and~\citet{chen2012learning} considered a decomposed parameter/model matrix to handle heterogeneities among tasks, e.g., detecting entry-wise outliers in the parameter matrix~\citep{jalali2010dirty,chen2012learning} and detecting anomalous tasks~\citep{gong2012robust,chen2011integrating}. These detection procedures are claimed to be beneficial for the knowledge sharing process in cases of heterogeneous tasks. Our MP-MTL framework can be naturally extended to such a model-decomposed setting because the additional procedures are still STL algorithms; hence, the privacy loss will not increase (see the supplementary material for additional details).

\subsection{Privacy Guarantees}

{The following two results show that our proposed framework and the two instantiated algorithms satisfy the privacy constraint defined in Definition \ref{th:df_MP-MTL-iter}.}
\begin{theorem}\label{th:th_MP-MTL}
Algorithm \ref{alg:MP-MTL} is an $(\epsilon,\delta)$ - MP-MTL algorithm.
\end{theorem}

\begin{corollary}\label{th:th_MP-MTL-LR}
Algorithm \ref{alg:MP-MTL-LR} and Algorithm \ref{alg:MP-MTL-GS} are $(\epsilon,\delta)$ - MP-MTL algorithms.
\end{corollary}

{
\subsection{Utility Analyses}\label{sec:utility}
{We build utility analyses specifically for our instantiations, for example, Algorithm \ref{alg:MP-MTL-LR} and \ref{alg:MP-MTL-GS} instead of Algorithm \ref{alg:MP-MTL}, because (1) Algorithm \ref{alg:MP-MTL} is a framework that allows the minimization of a variety of regularization terms and many optimization schemes. Specifically, Steps 6 and 7 of Algorithm \ref{alg:MP-MTL} include arbitrary mappings and arbitrary STL algorithms, respectively. Therefore, the analysis is not trivial and requires additional assumptions. (2) Algorithm \ref{alg:MP-MTL-LR} and \ref{alg:MP-MTL-GS} correspond to trace-norm and group-$\ell_1$-norm regularization, respectively, which correspond to two mainstream MTL approaches.

Our utility analyses are built upon the matrix perturbation error bounds of Wishart noise presented by~\citet{jiang2016wishart}, the error bounds with arbitrary heterogeneous residues of inexact proximal-gradient descent presented by~\citet{schmidt2011convergence}, and the two optimal solutions for proximal operators presented by~\citet{ji2009accelerated} and~\citet{liu2009multi}.
The following parts of the utility analyses are novel:
(1) the derivations of the approximation error bounds for both the proximal operators in \eqref{eq:proximal_op_LR} and \eqref{eq:proximal_op_GS};
(2) the derivations of runtime and utility bounds, considering three cases of composition bounds of privacy budgets, two privacy-budget allocation strategies, two specific regularization terms, both convex and strongly convex prediction loss functions, and both the basic and accelerated proximal-gradient methods, subject to the elaborate composition theorem of privacy;
(3) the optimizations of the utility bounds with respect to the parameters of privacy-budget allocation strategies.

We studied the utility bounds for three cases of the {composition bound of $\{\epsilon_t\}$ defined in \eqref{eq:set_eps}}. Here, we report the results for the following case, {because this case provides the minimum bound for small values of privacy budgets $\{\epsilon_t\}$ and $\delta$, such as for $\epsilon + e\delta\leq 1$; see~\citet{kairouz2017composition}.}
\begin{align*}
\epsilon = \sum_{t=1}^{T}\frac{(e^{\epsilon_t}-1)\epsilon_t}{(e^{\epsilon_t}+1)}
+\sqrt{\sum_{t=1}^{T}2\epsilon_t^2\log\biggl(e+\frac{\sqrt{\sum_{t=1}^{T}\epsilon_t^2}}{\delta}\biggr)}.
\end{align*}
The results for the other two cases {are similar} and can be found in the supplementary material.

}

{

First, we make some assumptions.
}

\noindent\textbf{\emph{\underline{Parameter space}}}. A bounded parameter space is assumed for model matrices:
\begin{align*}
\mathcal{W} = \{\W\in\mathbb{R}^{d\times m}: \max_{i\in[m]}\|\w_i\|_2\leq K\},
\end{align*}
where $K$ is the norm clipping parameter.

\noindent\textbf{\emph{\underline{Properties of objective functions}}}. We consider the loss function $f(\W) = \frac{1}{m}\sum\nolimits_{i=1}^m\calL_i(\X_i\w_i,\y_i)$ and assume that $mf(\W)$ is convex and has an $L$-Lipschitz-continuous gradient (as defined in \citet{schmidt2011convergence}).
Let $\W_* = \arg\min_{\W} mf(\W) + \lambda g(\W)$, where $g(\cdot) = \|\cdot\|_*$ for Algorithm \ref{alg:MP-MTL-LR} and $g(\cdot) = \|\cdot\|_{2,1}$ for Algorithm \ref{alg:MP-MTL-GS}. Without loss of generality, we assume that $\W_*\in \mathcal{W}$ and $f(\widetilde{\W}^{(0)})-f(\W_*) = O(K^2Lm)$. We adopt the notation $q = \min\{d,m\}$.

\noindent\textbf{\emph{\underline{The number of tasks}}}.
The number of tasks are assumed to be sufficient as follows.
\begin{assumption}\label{th:asm_m}
For Algorithm \ref{alg:MP-MTL-LR}, we assume that for sufficiently large $C>0$,
\begin{align*}
m > CK^2d^2\log^2(d) (\log(e+\epsilon/\sqrt{2}\delta)+2\epsilon)/\epsilon^2
.
\end{align*}

For Algorithm \ref{alg:MP-MTL-GS}, we assume that for sufficiently large $C>0$,
\begin{align*}
m > \left.C\log(d) \sqrt{\log(e+\epsilon/\sqrt{2}\delta)+2\epsilon}\middle/\epsilon\right..
\end{align*}

\end{assumption}

{Then, we present the results. Before reporting the utility bounds, we report two intermediate results: the approximation error bounds for proximal operators with trace-norm (low-rank) and group-$\ell_1$ (group-sparse) penalties, respectively. Note that for both results, the noise matrix $\E$ is allowed to be arbitrary.}
\begin{lemma}[Low rank]\label{th:lem_LR_iter_t}
Consider Algorithm \ref{alg:MP-MTL-LR}. For $t\in[T]$, in the $t$-th iteration, let $\C = \widetilde{\W}^{(t-1)}$. Let $r_c = \mbox{rank}(\C)\leq q$ be the rank of $\C$. Suppose that an index $k\leq q$ exists such that $\sigma_{k}(\C) > \eta\lambda$ and $\sigma_{k+1}(\C) \leq \eta\lambda.$
Assume that $2\sigma_1(\E)\leq \sqrt{\sigma_j(\C)}-\sqrt{\sigma_{j+1}(\C)}$ for $j \in [k]$.
Then, for any random matrix $\E\in \mathbb{R}^{d\times d}$, the following holds:
\begin{equation}\label{eq:bound_LR_iter_t}
\begin{split}
&\frac{1}{2\eta}\|\widehat{\W}^{(t)} - \C\|_F^2 + \lambda\|\widehat{\W}^{(t)}\|_* \\
& - \biggl\{\min_{\W} \frac{1}{2\eta}\|\W - \C\|_F^2 + \lambda\|\W\|_*\biggr\}\\
\leq&\frac{1}{\eta}\biggl(\frac{\sigma_1^2(\C)}{\eta\lambda}+ \sigma_1(\C)\biggl)\\
&\cdot\biggl[ \biggl(\frac{k^2}{
\eta\lambda} + 2k \biggr) \sigma_1(\E) + \max(0,r_c-k)\sqrt{\sigma_1(\E)}\biggr].
\end{split}
\end{equation}
\end{lemma}

\begin{lemma}[Group Sparse]\label{th:lem_GS_iter_t}
Consider Algorithm \ref{alg:MP-MTL-GS}. For $t\in[T]$, in the $t$-th iteration, let $\C = \widetilde{\W}^{(t-1)}$. Let the indices of the non-zero rows of $\C$ be denoted by $\mathcal{I}_c = \{j:\C^j\neq \mathbf{0}\}$, and let $r_{c,s} = |\mathcal{I}_c| \leq d$. Let $\SSigma_0 = \C\C\trans$. Suppose that an integer $k\leq d$ exists such that $\sum_{j=1}^{d}I(\sqrt{\SSigma_{jj,0}}\geq \eta\lambda) = k,$
where $I(\cdot)$ is the indicator function.
Then, for any random matrix $\E\in \mathbb{R}^{d\times d}$, the following holds:
\begin{equation}\label{eq:bound_GS_iter_t}
\begin{split}
&\frac{1}{2\eta}\|\widehat{\W}^{(t)} - \C\|_F^2 + \lambda\|\widehat{\W}^{(t)}\|_{2,1} \\
& - \biggl\{\min_{\W} \frac{1}{2\eta}\|\W - \C\|_F^2 + \lambda\|\W\|_{2,1}\biggr\}\\
\leq&\frac{1}{\eta}\biggl[\frac{r_{c,s}}{ \eta\lambda}\biggl(\max_{j\in[d]}\|\C^j\|_{2}\biggr)^2+ \biggl(\max_{j\in[d]}\|\C^j\|_{2}\biggr)\biggl]\\
&\cdot\biggl[ \frac{k}{2\eta\lambda} \max_{j\in[d]}|\E_{jj}| + \max(0,r_{c,s}-k) \max_{j\in[d]}\sqrt{|\E_{jj}|}\biggr].
\end{split}
\end{equation}
\end{lemma}

We find that the approximation error bounds both depend on
$\sigma_1(\E)$ (note that $\max_{j}|\E_{jj}|\leq \sigma_1(\E) $).

{Note that Lemma \ref{th:lem_LR_iter_t} requires $\eta\lambda$ to fall between the $k$-th and the $(k+1)$-th singular values in every iteration for the same $k$. Under Assumption \ref{th:asm_m} (the number of tasks is sufficiently large), when the initial task models in $W^{(0)}$ are acquired via proper STL methods, a significant margin will always exist between the $k$-th and the $(k+1)$-th singular values of the normalized model matrix $\C$ for the same $k$. Therefore, the above requirement is easily satisfied. It is similar for the requirement of $k$ in Lemma \ref{th:lem_GS_iter_t}.}

Now, we present guarantees regarding both utility and runtime.
In the following, $\E$ is assumed to be a Wishart random matrix in each iteration. The privacy budgets $\{\epsilon_t\}$ are considered heterogeneous, i.e., different with respect to $t\in[T]$.

{
We consider two cases for the loss function $f(\W)$: convex and strongly convex. For each case, we report the results of both Algorithms \ref{alg:MP-MTL-LR} (the low-rank estimator) and \ref{alg:MP-MTL-GS} (the group-sparse estimator). For each algorithm, we present the results for both the basic and the accelerated proximal gradient descent methods.

}

For the convex case of the loss function $f(\W)$, we set $\epsilon_t =\Theta(t^{\alpha})$ for $\alpha\in \mathbb{R}$ and $t\in[T]$. Define
\begin{align*}
M_0= \left.{\sqrt{\log(e+\epsilon/\sqrt{2}\delta)+2\epsilon} }\middle/{ \sqrt{|2\alpha+1|}\epsilon}\right.,
\end{align*}
which is used for both Theorems \ref{th:th_bound_LR_main_convex} and \ref{th:th_bound_GS_main_convex}.

\begin{theorem}[Low rank - Convexity]\label{th:th_bound_LR_main_convex}
Consider Algorithm \ref{alg:MP-MTL-LR}. For an index $k\leq q$ that satisfies the conditions given in Lemma \ref{th:lem_LR_iter_t} for all $t\in[T]$, $\eta =1/L$, and $\lambda = \Theta(LK\sqrt{m} )$, assume that $\epsilon_t\leq 4Kk^2d(\log d)/q^2$ for $t\in[T]$. Define
\begin{align*}
M = M_0Kkd\log d/\sqrt{m}.
\end{align*}
\noindent\textbf{No acceleration}: If we set $\beta_t = 0$ for $t\in [m]$ and then also set $T=\Theta(((\alpha/2 - 1)^2/M)^{\phi(\alpha)} )$
for $\mathcal{E} = f(\frac{1}{T}\sum_{t=1}^{T}\widehat{\W}^{(t)}) - f(\W_*)$, we have, with high probability,
\begin{equation}\label{eq:bound_LR_main_basic_convex}
\mathcal{E}= O( K^2L(M/(\alpha/2 - 1)^2)^{\phi(\alpha)} ),
\end{equation}
where
\begin{equation}\label{eq:bound_LR_main_basic_convex_phi}
\begin{split}
\phi(\alpha) =
\left\{
\begin{array}{ll}
2/(2\alpha+1), & {\alpha>2;} \\
2/5, & {-1/2<\alpha<2;} \\
1/(2-\alpha), & {\alpha<-1/2.}
\end{array}
\right.
\end{split}
\end{equation}

\noindent\textbf{Use acceleration}: If we set $\beta_t = {(t-1)}/{(t+2)}$ for $t\in [m]$ and then also set $T=\Theta(((\alpha/2 - 2)^2/M)^{\phi(\alpha)/2} )$
for $\mathcal{E} = f(\widehat{\W}^{(T)}) - f(\W_*)$, we have, with high probability,
\begin{equation}\label{eq:bound_LR_main_acc_convex}
\mathcal{E}= O( K^2L(M/(\alpha/2 - 2)^2)^{\phi(\alpha)} ),
\end{equation}
where
\begin{equation}\label{eq:bound_LR_main_acc_convex_phi}
\begin{split}
\phi(\alpha) =
\left\{
\begin{array}{ll}
4/(2\alpha+1), & {\alpha>4;} \\
4/9, & {-1/2<\alpha<4;} \\
2/(4-\alpha), & {\alpha<-1/2.}
\end{array}
\right.
\end{split}
\end{equation}
\end{theorem}

\begin{theorem}[Group sparse - Convexity]\label{th:th_bound_GS_main_convex}
Consider Algorithm \ref{alg:MP-MTL-GS}. For an index $k\leq d$ that satisfies the condition given in Lemma \ref{th:lem_GS_iter_t} for all $t\in[T]$, $\eta =1/L$, and $\lambda = \Theta(LKd\sqrt{m} )$, assume that $\epsilon_t\leq {k^2\log(d)}/{4Kd(d-k)^2m}$ for $t\in[T]$. Define
\begin{align*}
M = M_0k\log d/m.
\end{align*}
\noindent\textbf{No acceleration}: If we set $\beta_t = 0$ for $t\in [m]$and then also set $T=\Theta(((\alpha/2 - 1)^2/M)^{\phi(\alpha)} )$
for $\mathcal{E} = f(\frac{1}{T}\sum_{t=1}^{T}\widehat{\W}^{(t)}) - f(\W_*)$, we have, with high probability,
\begin{equation}\label{eq:bound_GS_main_basic_convex}
\mathcal{E}= O( K^2L(M/(\alpha/2 - 1)^2)^{\phi(\alpha)} ),
\end{equation}
where $\phi(\alpha)$ is defined in \eqref{eq:bound_LR_main_basic_convex_phi}.

\noindent\textbf{Use acceleration}: If we set $\beta_t = {(t-1)}/{(t+2)}$ for $t\in [m]$ and then also set $T=\Theta(((\alpha/2 - 2)^2/M)^{\phi(\alpha)/2} )$
for $\mathcal{E} = f(\widehat{\W}^{(T)}) - f(\W_*)$, we have, with high probability,
\begin{equation}\label{eq:bound_GS_main_acc_convex}
\mathcal{E}= O( K^2L(M/(\alpha/2 - 2)^2)^{\phi(\alpha)} ),
\end{equation}
where $\phi(\alpha)$ is defined in \eqref{eq:bound_LR_main_acc_convex_phi}.
\end{theorem}

Next, we assume that $mf(\W)$ is $\mu$-strongly convex and has an $L$-Lipschitz-continuous gradient, where $\mu<L$.
In this case, we set $\epsilon_t =\Theta(Q^{-t})$ for $Q>0$ and $t\in[T]$ and define
\begin{align*}
M'_0= \left.{ \sqrt{\log(e+\epsilon/\sqrt{2}\delta)+2\epsilon}}\middle/{ \sqrt{|1-Q^2|} \epsilon}\right.,
\end{align*}
which is used for both Theorems \ref{th:th_bound_LR_main_strong_convex} and \ref{th:th_bound_GS_main_strong_convex}.

\begin{theorem}[Low rank - Strong convexity]\label{th:th_bound_LR_main_strong_convex}
Consider Algorithm \ref{alg:MP-MTL-LR}. For an index $k\leq q$ that satisfies the conditions given in Lemma \ref{th:lem_LR_iter_t} for all $t\in[T]$, $\eta =1/L$, and $\lambda = \Theta(LK\sqrt{m} )$, assume that $\epsilon_t\leq 4Kk^2d(\log d)/q^2$ for $t\in[T]$, denoted by
\begin{align*}
M = M'_0Kkd\log d/\sqrt{m}.
\end{align*}
\noindent\textbf{No acceleration}: If we set $\beta_t = 0$ for $t\in [m]$ and then let $Q_0 = 1-\mu/L$ and set $T=\Theta(\log_{1/\psi(Q,Q_0^2)}((Q_0/\sqrt{Q} - 1)^2/M) )$
for $\mathcal{E} = \frac{1}{\sqrt{m}}\|\widehat{\W}^{(T)}-\W_*\|_F$, we have, with high probability,
\begin{equation}\label{eq:bound_LR_main_basic_strong_convex}
\mathcal{E}= O(K(M/(Q_0/\sqrt{Q} - 1)^2)^{\log_{\psi(Q,Q_0^2)}Q_0}) ,
\end{equation}
where for any $\widetilde{Q}\in(0,1)$,
\begin{equation}\label{eq:bound_LR_main_basic_strong_convex_psi}
\begin{split}
\psi(Q,\widetilde{Q}) =
\left\{
\begin{array}{ll}
Q, & {0<Q<\widetilde{Q};} \\
\widetilde{Q}, & {\widetilde{Q}<Q<1;} \\
\widetilde{Q}/Q, & {Q>1.}
\end{array}
\right.
\end{split}
\end{equation}

\noindent\textbf{Use acceleration}: If we set $\beta_t = {(1-\sqrt{\mu/L})}/{(1+\sqrt{\mu/L})}$ for $t\in [m]$ and then let $Q'_0 = 1-\sqrt{\mu/L}$ and set $T=\Theta(\log_{1/\psi(Q,Q'_0)}((\sqrt{Q'_0}/\sqrt{Q} - 1)^2/M) )$
for $\mathcal{E} = f(\widehat{\W}^{(T)}) - f(\W_*)$, we have, with high probability,
\begin{equation}\label{eq:bound_LR_main_acc_strong_convex}
\mathcal{E}= O(K(M/(\sqrt{Q'_0}/\sqrt{Q} - 1)^2)^{\log_{\psi(Q,Q'_0)}Q'_0}),
\end{equation}
where $\psi(\cdot,\cdot)$ is defined in \eqref{eq:bound_LR_main_basic_strong_convex_psi}.
\end{theorem}

\begin{theorem}[Group sparse - Strong convexity]\label{th:th_bound_GS_main_strong_convex}
Consider Algorithm \ref{alg:MP-MTL-GS}. For an index $k\leq d $ that satisfies the condition given in Lemma \ref{th:lem_GS_iter_t} for all $t\in[T]$, $\eta =1/L$, and $\lambda = \Theta(LKd\sqrt{m} )$, assume that $\epsilon_t\leq {k^2\log(d)}/{4Kd(d-k)^2m}$ for $t\in[T]$. Define
\begin{align*}
M = M'_0k\log d/m.
\end{align*}
\noindent\textbf{No acceleration}: If we set $\beta_t = 0$ for $t\in [m]$ and then let $Q_0 = 1-\mu/L$ and set $T=\Theta(\log_{1/\psi(Q,Q_0^2)}((Q_0/\sqrt{Q} - 1)^2/M) )$
for $\mathcal{E} = \frac{1}{\sqrt{m}}\|\widehat{\W}^{(T)}-\W_*\|_F$, we have, with high probability,
\begin{equation}\label{eq:bound_GS_main_basic_strong_convex}
\mathcal{E}= O(K(M/(Q_0/\sqrt{Q} - 1)^2)^{\log_{\psi(Q,Q_0^2)}Q_0}),
\end{equation}
where $\psi(\cdot,\cdot)$ is defined in \eqref{eq:bound_LR_main_basic_strong_convex_psi}.

\noindent\textbf{Use acceleration}: If we set $\beta_t = {(1-\sqrt{\mu/L})}/{(1+\sqrt{\mu/L})}$ for $t\in [m]$ and then let $Q'_0 = 1-\sqrt{\mu/L}$ and set $T=\Theta(\log_{1/\psi(Q,Q'_0)}((\sqrt{Q'_0}/\sqrt{Q} - 1)^2/M) )$
for $\mathcal{E} = f(\widehat{\W}^{(T)}) - f(\W_*)$, we have, with high probability,
\begin{equation}\label{eq:bound_GS_main_acc_strong_convex}
\mathcal{E}= O(K(M/(\sqrt{Q'_0}/\sqrt{Q} - 1)^2)^{\log_{\psi(Q,Q'_0)}Q'_0}),
\end{equation}
where $\psi(\cdot,\cdot)$ is defined in \eqref{eq:bound_LR_main_basic_strong_convex_psi}.
\end{theorem}

\subsection{Privacy Budget Allocation}\label{subsec:Budget}
{In this section, we optimize the utility bounds presented in Theorems \ref{th:th_bound_LR_main_convex}--\ref{th:th_bound_GS_main_strong_convex} with respect to $\alpha$ and $Q$, respectively, which results in optimized privacy-budget allocation strategies. Then, we discuss the optimized results.}

\begin{theorem}\label{th:th_budget}
Consider Algorithms \ref{alg:MP-MTL-LR} and \ref{alg:MP-MTL-GS}.

For a \emph{convex} $f$, we use Theorems \ref{th:th_bound_LR_main_convex} and \ref{th:th_bound_GS_main_convex}.

(1) \emph{No acceleration}: The bounds for the low-rank and group-sparse estimators both reach their respective minima w.r.t. $\alpha$ at $\alpha = 0$. Meanwhile, $\phi(\alpha) = 2/5$.

(2) \emph{Use acceleration}: The bounds for low-rank and group-sparse estimators both reach their respective minima w.r.t. $\alpha$ at $\alpha = 2/5$. Meanwhile, $\phi(\alpha) = 4/9$.

For a \emph{strongly convex} $f$, we use Theorems \ref{th:th_bound_LR_main_strong_convex} and \ref{th:th_bound_GS_main_strong_convex}.

(1) \emph{No acceleration}: The bounds for the low-rank and group-sparse estimators both reach their respective minima w.r.t. $Q$ at $Q = Q_0^{2/5}$. Meanwhile, $\log_{\psi(Q,Q_0^2)}Q_0 = 1/2$.

(2) \emph{Use acceleration}: The bounds for low-rank and group-sparse estimators both reach their respective minima w.r.t. $Q$ at $Q = (Q'_0)^{1/5}$. Meanwhile, $\log_{\psi(Q,Q'_0)}Q'_0 = 1$.
\end{theorem}

The results corresponding to the optimized privacy-budget allocation strategies (with $\delta>0$) are summarized in Table \ref{tab:utility_results}, where the terms with respect to $K,L,k$, and $\sqrt{\log(e+\epsilon/\sqrt{2}\delta)+2\epsilon}$ are omitted, and the results associated with the setting $\epsilon = \sum_{t=1}^{T}\epsilon_t$ (and $\delta=0$) are included, providing $(\epsilon,0)$ - MP-MTL algorithms.

\setlength{\tabcolsep}{2pt}
\begin{table}[htbp]\small
\centering
\renewcommand{\multirowsetup}{\centering}
\caption{Utility results.}\label{tab:utility_results}
\begin{tabular}{c|c|c|c|c}
\hline
\hline
& & &Low rank& Group sparse\\
\hline
\multirow{4}{*}{$\delta=0$}&\multirow{2}{1.6cm}{\footnotesize No Acceleration}&Convex& $\mathcal{O}((\frac{d\log(d)}{\sqrt{m}\epsilon})^{\frac{1}{3}})$ &$\mathcal{O}((\frac{\log(d)}{{m}\epsilon})^{\frac{1}{3}})$ \\
\cline{3-5}
&&Strong convex&$\mathcal{O}((\frac{d\log(d)}{\sqrt{m}\epsilon})^{\frac{1}{2}})$ &$\mathcal{O}((\frac{\log(d)}{{m}\epsilon})^{\frac{1}{2}})$ \\
\cline{2-5}
&\multirow{2}{1.6cm}{\footnotesize Use Acceleration}&Convex&$\mathcal{O}((\frac{d\log(d)}{\sqrt{m}\epsilon})^{\frac{2}{5}})$ & $\mathcal{O}((\frac{\log(d)}{{m}\epsilon})^{\frac{2}{5}})$\\
\cline{3-5}
&&Strong convex&$\mathcal{O}(\frac{d\log(d)}{\sqrt{m}\epsilon})$&$\mathcal{O}(\frac{\log(d)}{{m}\epsilon})$\\
\hline
\multirow{4}{*}{$\delta>0$}&\multirow{2}{1.6cm}{\footnotesize No Acceleration}&Convex& $\mathcal{O}((\frac{d\log(d)}{\sqrt{m}\epsilon})^{\frac{2}{5}})$ &$\mathcal{O}((\frac{\log(d)}{{m}\epsilon})^{\frac{2}{5}})$ \\
\cline{3-5}
&&Strong convex&$\mathcal{O}((\frac{d\log(d)}{\sqrt{m}\epsilon})^{\frac{1}{2}})$ &$\mathcal{O}((\frac{\log(d)}{{m}\epsilon})^{\frac{1}{2}})$ \\
\cline{2-5}
&\multirow{2}{1.6cm}{\footnotesize Use Acceleration}&Convex&$\mathcal{O}((\frac{d\log(d)}{\sqrt{m}\epsilon})^{\frac{4}{9}})$ & $\mathcal{O}((\frac{\log(d)}{{m}\epsilon})^{\frac{4}{9}})$\\
\cline{3-5}
&&Strong convex&$\mathcal{O}(\frac{d\log(d)}{\sqrt{m}\epsilon})$&$\mathcal{O}(\frac{\log(d)}{{m}\epsilon})$\\
\hline
\hline
\end{tabular}
\end{table}

{We learn from Theorem \ref{th:th_budget} that (1) for all four settings, a non-decreasing series of $\{\epsilon_t\}$ results in a good utility bound, since the best $\alpha = 0, 2/5 \geq 0$ for $\epsilon_t=\Theta(t^{\alpha})$ and the best $Q = Q_0^{2/5}, (Q'_0)^{1/5} < 1$ for $\epsilon_t=\Theta(Q^{-t})$. Intuitively, this means adding non-increasing noise over the iterations---which is reasonable because the initial iterations may move quickly in the parameter space while the last iterations may only fine-tune the model slightly. (2) Both the strong-convexity condition and the acceleration strategy improve the utility bounds: both increase the powers of those bounds that are far less than $1$ under Assumption \ref{th:asm_m}. (3) By setting $\alpha$ and $Q$ to their optimized values, the acceleration strategy improves the runtime, as shown in Claim \ref{claim:acc_runtime}.

\begin{claim}\label{claim:acc_runtime}
Assume Assumption \ref{th:asm_m}. Consider Theorem \ref{th:th_bound_LR_main_convex}-\ref{th:th_bound_GS_main_strong_convex} and set $\alpha$ and $Q$ to the optimized values in Theorem \ref{th:th_budget}, respectively. Assume $\mu/L<0.3819$. The values for $T$ are smaller when using the acceleration strategy compared to those with no acceleration.
\end{claim}

}

Now, we introduce our concrete strategy in the following to set $\{\epsilon_t\}$ in both Algorithm \ref{alg:MP-MTL-LR} and Algorithm \ref{alg:MP-MTL-GS}. We assume that $T$, $\epsilon$ and $\delta$ are given. {Note that this strategy is optimal if $\alpha$ and $Q$ are set according to the optimal settings stated by Theorem \ref{th:th_budget}.}

For a \emph{convex} $f$, if no acceleration is to be used, then set $\beta_t = 0$ for $t\in [m]$ and set $\alpha \in \mathbb{R}$ (e.g., $\alpha=0$); otherwise, set $\beta_t = {(t-1)}/{(t+2)}$ for $t\in [m]$ and set $\alpha \in \mathbb{R}$ (e.g., $\alpha = 2/5$). Then, for $t\in[T]$, let $\epsilon_t = \epsilon_0t^{\alpha}$ and find the largest $\epsilon_0$ that satisfies $CB(\{\epsilon_t\},\delta)\leq \epsilon$, where $CB(\{\epsilon_t\},\delta)$ is the composition bound of $\{\epsilon_t\}$ defined in \eqref{eq:set_eps}.

For a \emph{$\mu$-strongly convex} $mf(\W)$ with a known value of $\mu$ (e.g., $\frac{\mu}{2}\|\w_i\|_2^2$ is added to each $\mathcal{L}_i$), if no acceleration is to be used, then set $\beta_t = 0$ for $t\in [m]$ and set $Q > 0$ (e.g., $Q=(1-\mu/L)^{2/5}$, if $L$ is known); otherwise, if $L$ is known, set $\beta_t = {(1-\sqrt{\mu/L})}/{(1+\sqrt{\mu/L})}$ for $t\in [m]$ and set $Q>0$ (e.g., $Q=(1-\sqrt{\mu/L})^{1/5}$). Then, for $t\in[T]$, let $\epsilon_t = \epsilon_0Q^{-t}$ and find the largest $\epsilon_0$ that satisfies $CB(\{\epsilon_t\},\delta)\leq \epsilon$, where $CB(\{\epsilon_t\},\delta)$ is the composition bound of $\{\epsilon_t\}$ defined in \eqref{eq:set_eps}.

}

\subsection{Baseline MP-MTL Constructed by IP-MTL}\label{subsec:SMC}

IP-MTL algorithms \emph{prevent a single data instance} in one task from leaking to other tasks and are formally defined as follows.

\begin{definition}[IP-MTL]\label{th:df_DP-MTL-iter}
Let $\calA$ be a randomized MTL algorithm with a number of iterations $T$. In the first iteration, $\calA$ performs the mapping $(\W^{(0)} \in\mathbb{R}^{d\times m},\calD^m)\rightarrow \theta_1\in \calC_1$, where $\theta_1$ includes $\widehat{\W}^{(1)} \in\mathbb{R}^{d\times m}$. For $t=2,\ldots,T$, in the $t$-th iteration, $\calA$ performs the mapping $(\W^{(t-1)} \in\mathbb{R}^{d\times m},\calD^m,\theta_1,\ldots,\theta_{t-1})\rightarrow \theta_t\in \calC_t$, where $\theta_t$ includes $\widehat{\W}^{(t)} \in\mathbb{R}^{d\times m}$. Here, $\calA$ is an $(\epsilon,\delta)$ - \emph{IP-MTL} algorithm if---for all $i\in[m]$ and for all neighboring datasets $\calD^m$ and $(\calD')^m$ that differ by a single data instance for the $i$-th task---the following holds for some constants $\epsilon,\delta\geq 0$ and for any set $\calS \subseteq \mathbb{R}^{d\times (m-1)\times T}$:

\begin{equation}\label{eq:DP_MTL-iter}
\begin{split}
\bbP( & \hat{ \w}_{[-i]}^{(1:T)} \in \calS \mid \bigcap_{t=1}^T \calB_t = (\W^{(t-1)}, \calD^m, \ttheta_{1:t-1}) ) \\
\leq &e^{\epsilon} \bbP( \hat{ \w}_{[-i]}^{(1:T)} \in \calS \mid \bigcap_{t=1}^T \calB_t = ((\W')^{(t-1)}, (\calD')^m, \ttheta_{1:t-1})) \\
&+ \delta,
\end{split}
\end{equation}
where for all $t\in[T]$, $\calB_t$ denotes the input for the $t$-th iteration,
\begin{numcases}{\ttheta_{1:t-1} = }
\nonumber \emptyset, & $t=1$\\
\nonumber \theta_1,\theta_2,\cdots,\theta_{t-1}, & $t\geq2$,
\end{numcases}
and $(\W')^{(t-1)}$ is associated with the case where a single data instance for the $i$-th task has been replaced.
\end{definition}

As examples, the methods of \citet{pathak2010multiparty} and~\citet{gupta2016differentially} both fall into this category.
\begin{proposition}\label{th:DP-MP-iter_example}
The methods of both \citet{pathak2010multiparty} and~\citet{gupta2016differentially} are IP-MTL algorithms with $T=1$ and $T\geq 1$, respectively.
\end{proposition}

Now we can construct baseline MP-MTL methods by IP-MTL methods based on result of Proposition \ref{th:prop_DP-MP-iter}: to guarantee an $(\epsilon,\delta)$ - MP-MTL algorithm, one can use an $(\epsilon/n,\delta/(n\exp(\epsilon))$ - IP-MTL algorithm.
\begin{proposition}\label{th:prop_DP-MP-iter}
For task sample sizes of $n_1,\ldots,n_m$, any $(\epsilon,\delta)$ - IP-MTL algorithm is a $(n\epsilon,n\exp(n\epsilon)\delta)$ - MP-MTL algorithm when $n=\max_{i\in[m]}n_i$.
\end{proposition}
{The proof of Proposition \ref{th:prop_DP-MP-iter} can be found in the supplementary material, directly following the proof of the group privacy Lemma stated by Lemma 2.2 of~\citet{vadhan2016complexity}. Therefore, Proposition \ref{th:prop_DP-MP-iter} be regarded as the group privacy property of differential privacy applied to a
``group'' of the entire dataset for a single task. }

\section{Experiments}\label{sec:exp}

In this section, we evaluate the proposed MP-MTL method. We evaluate two instantiations of our method, Algorithm \ref{alg:MP-MTL-LR} and Algorithm \ref{alg:MP-MTL-GS} with respect to their ability to capture the low-rank and group-sparse patterns, respectively, in the model matrix.
We use both synthetic and real-world datasets to evaluate these algorithms. All the algorithms were implemented in
MATLAB.

\subsection{Methods for Comparison}
We use least-square loss and logistic loss for the least-square regression and binary classification problems, respectively.

For each setting, we evaluate three types of methods: 1) non-private STL methods, in which each task is learned independently without the introduction of any perturbation; 2) {MP-MTL methods, including our proposed MP-MTL methods and baseline MP-MTL methods constructed by IP-MTL methods}; and 3) non-private MTL methods, which correspond to the original MTL methods without the introduction of any perturbation.

To select the IP-MTL methods for constructing the baseline MP-MTL methods, because few such approaches have been proposed, we first consider the DP-MTRL method proposed by~\citet{gupta2016differentially}. The authors of this method did not consider privacy-loss increase resulting from their iterative update procedure. We solve this problem in our comparison by using the same composition technique as in our method.

We also modified the DP-MTRL method to consider the Lipschitz constants of the loss functions when computing the sensitivities in the 4th step of the algorithm, which were omitted in the Algorithm 1 presented in the cited paper. For all $i\in[m]$, the Lipschitz constant $L_i$ of the loss function $\calL_i$ is estimated as $L_i = \max_{j\in[n_i]}|\calL'_i(\x_{ij}\w_i,y_{ij})|$, which is smaller than the true value. Thus, intuitively, we add less noise to their algorithm than
would otherwise be added. %

{For the binary classification case, we still let DP-MTRL minimize the least-square loss}, because in each of its outer iterations (which alternately compute the parameter matrix and its covariance matrix) DP-MTRL requires a closed-form solution to guarantee the theoretical privacy results. However, in the logistic loss case, iterative optimization is required in each outer iteration; consequently, the requirement of a closed-form solution cannot be satisfied. Therefore, DP-MTRL provides no privacy guarantee for logistic loss. Moreover, it is not trivial to modify the DP-MTRL algorithm for loss functions that require iterative optimization in each outer iteration because additional leakage will occur in each inner iteration.

The DP-AGGR method proposed by~\citet{pathak2010multiparty} which outputs an averaged model as the final solution, is also considered to be an IP-MTL method that transforms into a baseline MP-MTL method.

\begin{remark}
{We continue to refer to the baseline MP-MTL methods constructed by IP-MTL methods (DP-MTRL and DP-AGGR) using their respective names.}
\end{remark}

Differentially private STL methods are not considered because 1) empirically, they are always outperformed by non-private STL methods~\citep{chaudhuri2011differentially, wang2015privacy}, and 2) our MP-MTL method always outperforms STL methods, as will be demonstrated later.

\subsection{Experimental Setting}

For the non-private methods, the regularization parameters and the numbers of iterations were optimized via $5$-fold cross-validation on the training data
{, and acceleration was used without considering the strong convexity of the loss function $f$}. {For the private methods, the regularization parameters, the number of iterations, the optimization strategy (whether to use acceleration and whether to consider strong convexity via adding $\ell_2$ norm penalties), and the privacy-budget allocation hyper-parameters ($\alpha$ and $Q$) under each privacy loss $\epsilon$ were optimized via $5$-fold cross-validation on the training data. In the case considering strong convexity, $\frac{\mu}{2}\|\w_i\|_2^2$ was added to each $\mathcal{L}_i$ with $\mu = 1e-3$.}

{Note that the parameter tuning step using cross-validation was not included in the privacy budget for the algorithms. In this paper, we regarded the hyper-parameters generated by cross-validation as given not only for our methods but also for the baseline methods (DP-AGGR and DP-MTRL). We plan to explore an effective cross-validation method using the minimum privacy budget with the optimum utility in future work.}

For all the experiments, the $\delta$ values in the MP-MTL algorithms were set to {$1/m\log(m)$}, where $m$ is the number of tasks as suggested by~\citet{abadi2016deep}, and the $\delta$ values in the baseline MP-MTL methods, i.e., DP-MTRL and DP-AGGR, were set in accordance with Proposition \ref{th:prop_DP-MP-iter}.

All the experiments were replicated 100 times under each model setting.

\subsection{Evaluation Metrics}

{We adopt the evaluation metrics commonly-encountered in MTL approaches.} For least-square regression, we use nMSE~\citep{chen2011integrating,gong2012robust}, which is defined as the mean squared error (MSE) divided by the variance of the target vector. For binary classification, we use the average AUC~\citep{chen2012learning}, which is defined as the mean value of the area under the ROC curve for each task.

\subsection{Simulation}

We created a synthetic dataset as follows. The number of tasks was {$m = 320$}, the number of training samples for each task was {$n_i = 30$},
and the feature dimensionality of the training samples was {$d = 30$}.
The entries of the training data $\X_i \in \mathbb{R}^{ n_i\times d}$ (for the $i$-th
task) were randomly generated from the normal distribution $\mathcal{N}(0, 1)$ before being normalized such that the $\ell_2$ norm of each sample was one.

\begin{figure}[t!]
\centering
\subfigure[True]{\includegraphics[width=1.1in]{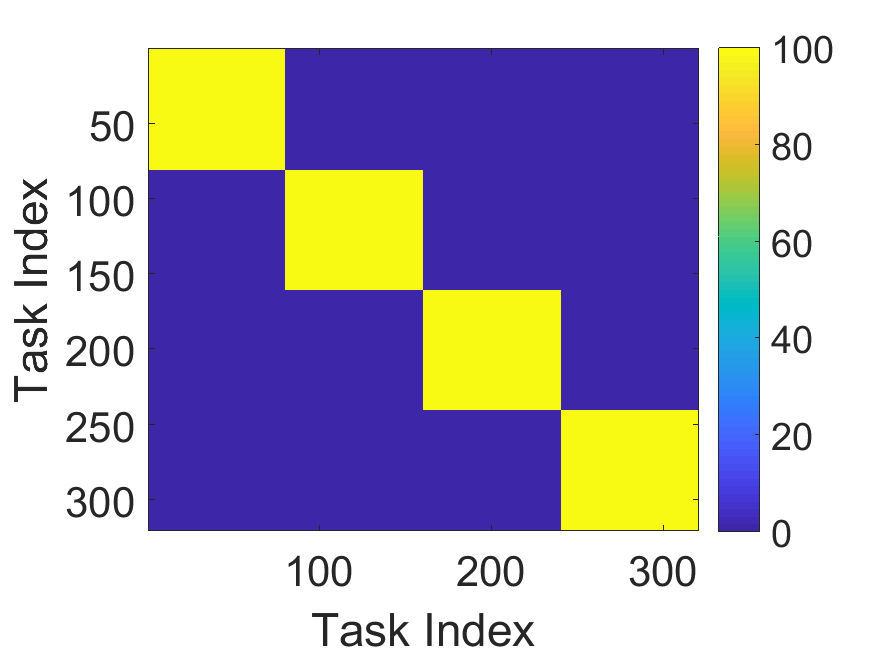}}
\subfigure[DP-MTRL]{\includegraphics[width=1.1in]{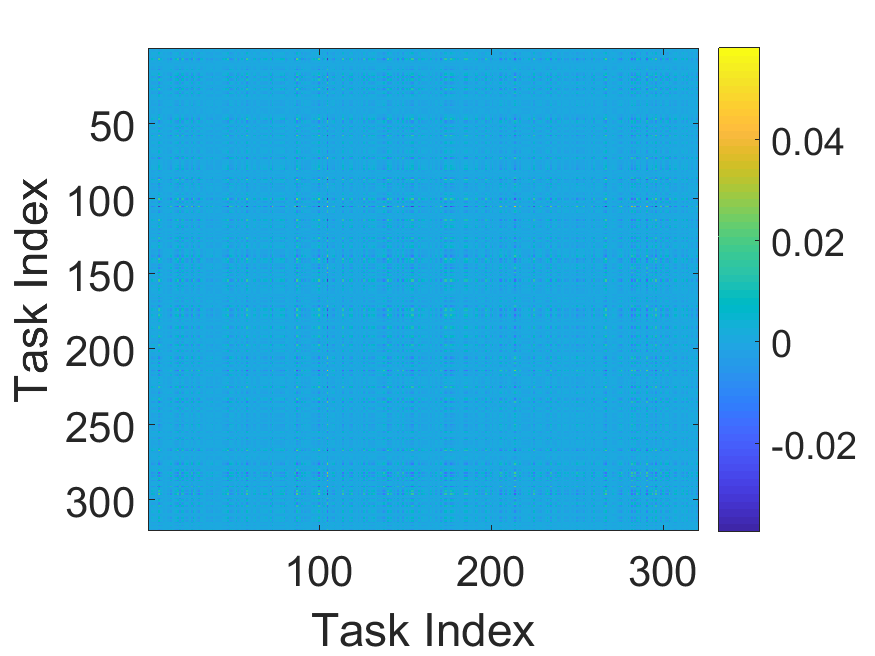}}
\subfigure[MP-MTL]{\includegraphics[width=1.1in]{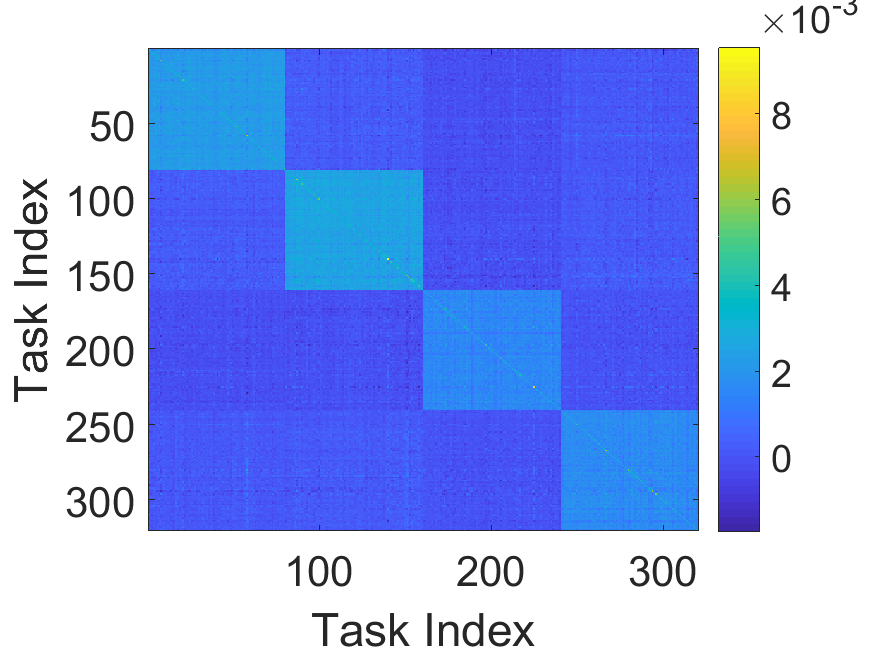}}
\subfigure[True]{\includegraphics[width=1.1in]{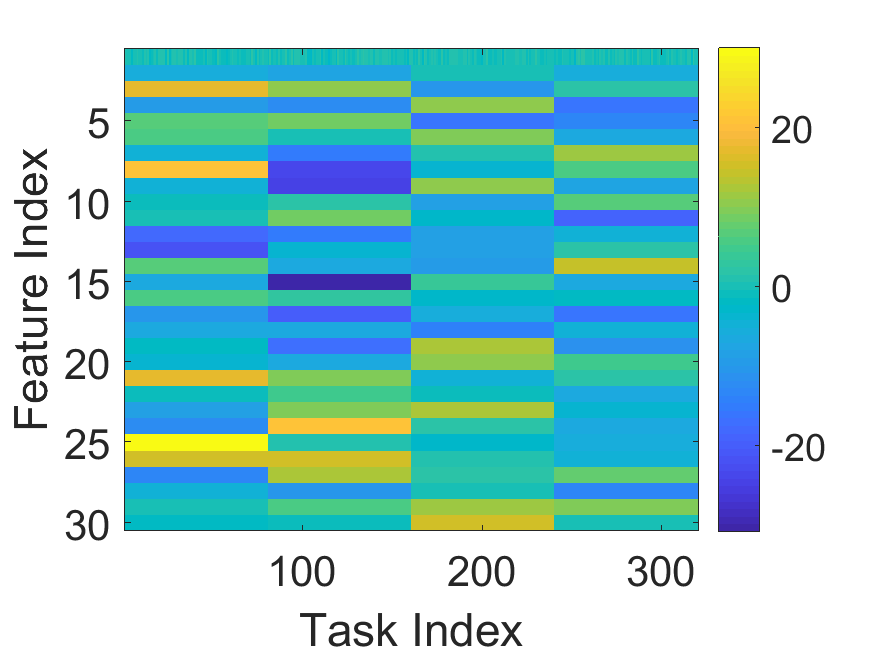}}
\subfigure[DP-MTRL]{\includegraphics[width=1.1in]{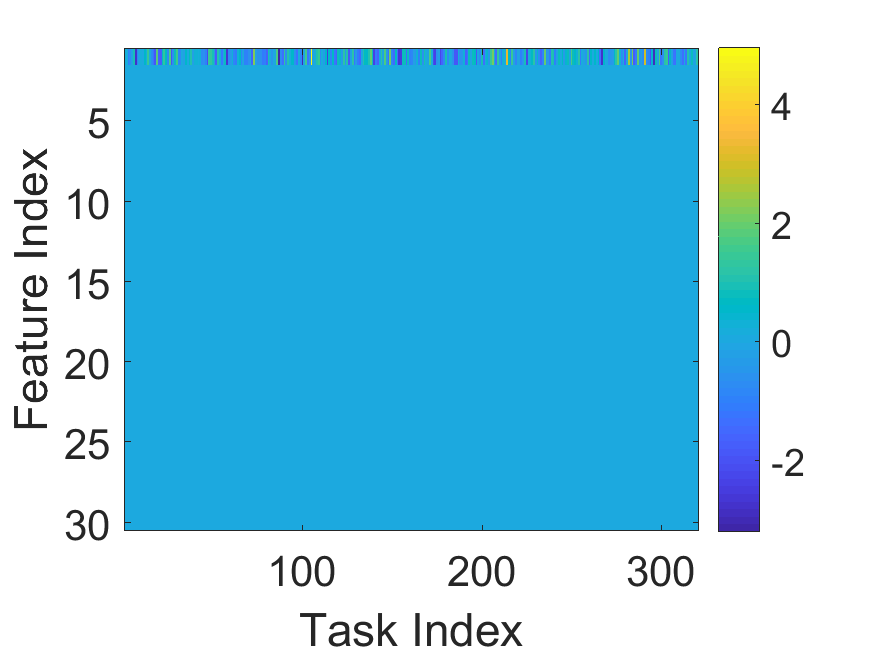}}
\subfigure[MP-MTL]{\includegraphics[width=1.1in]{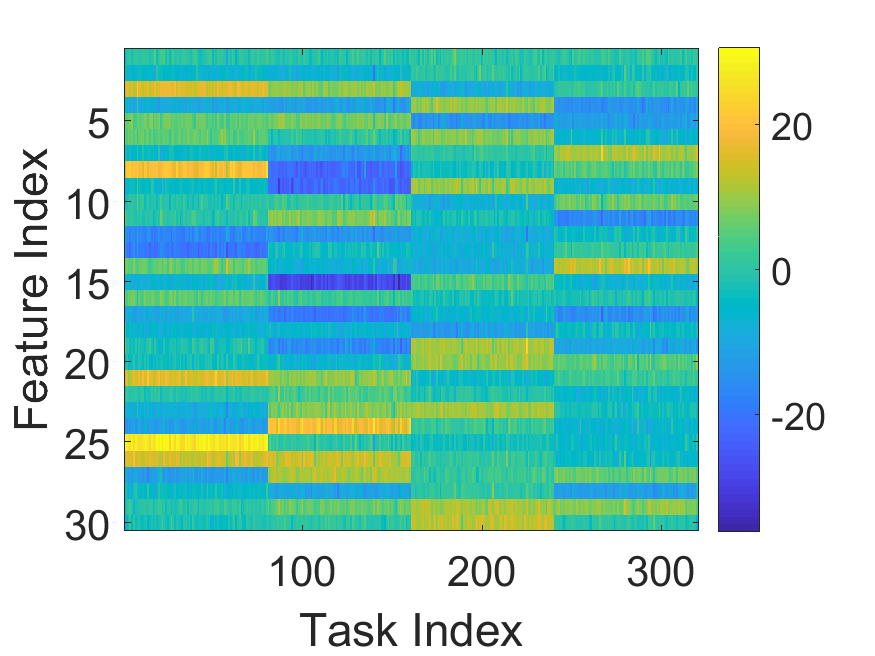}}
\caption{Task relationships and output model matrices for the synthetic data experiments: (a), (b) and (c) are task relationship matrices, (d), (e) and (f) are the output model matrices. The results shown are the averages of $100$ runs with $\epsilon = 0.1$.}\label{fig:syn_relation}
\end{figure}

{To obtain a low-rank pattern, we first generated a covariance matrix $\SSigma \in \mathbb{R}^{m\times m}$ as shown in Fig. \ref{fig:syn_relation} (a). Then, the model parameter matrix $\W\in\mathbb{R}^{d\times m}$ (see Fig. \ref{fig:syn_relation} (d)) was generated from a matrix variate normal (MVN) distribution~\citep{gupta1999matrix}, i.e., $\W\sim MVN(\mathbf{0},\I,\SSigma)$.}

To obtain a group-sparse pattern, we generated the model parameter matrix $\W\in\mathbb{R}^{d\times m}$ such that the first $4$ rows were nonzero. The values of the nonzero entries were generated from a uniform distribution in the range $[-50, -1]\bigcup [1, 50]$. %

Without loss of generality, we consider only the simulation of least-square regression. The results for logistic regression are similar. The response (target) vector for each task was $\y_i = \X_i\w_i + \varepsilon_i \in \mathbb{R}^{n_i} (i \in [m])$, where each entry in the vector $\varepsilon_i$ was randomly generated from $\mathcal{N} (0, 1)$.

The test set was generated in the same manner; the number of test samples was $9n_i$.

{
\subsubsection{Privacy Budget Allocation}\label{sec:exp_ba_simu}
The privacy-budget allocation strategies in Section \ref{subsec:Budget} were evaluated based on the synthetic data associated with the low-rank model matrix. The results shown in Fig. \ref{fig:syn_budget_allocation} are from a $5$-fold cross-validation on the training data. The prediction performances increase when acceleration is used, and achieve local optima at small positive values of the horizontal axes, which is consistent with our utility analyses. A local optimum exists in the negative horizontal axis in Fig. \ref{fig:syn_budget_allocation} (b) when acceleration is used---perhaps because $m$ is not sufficiently large as assumed in Assumption \ref{th:asm_m}.
\begin{figure}[t!]
\centering
\subfigure[Convexity]{\includegraphics[width=1.7in]{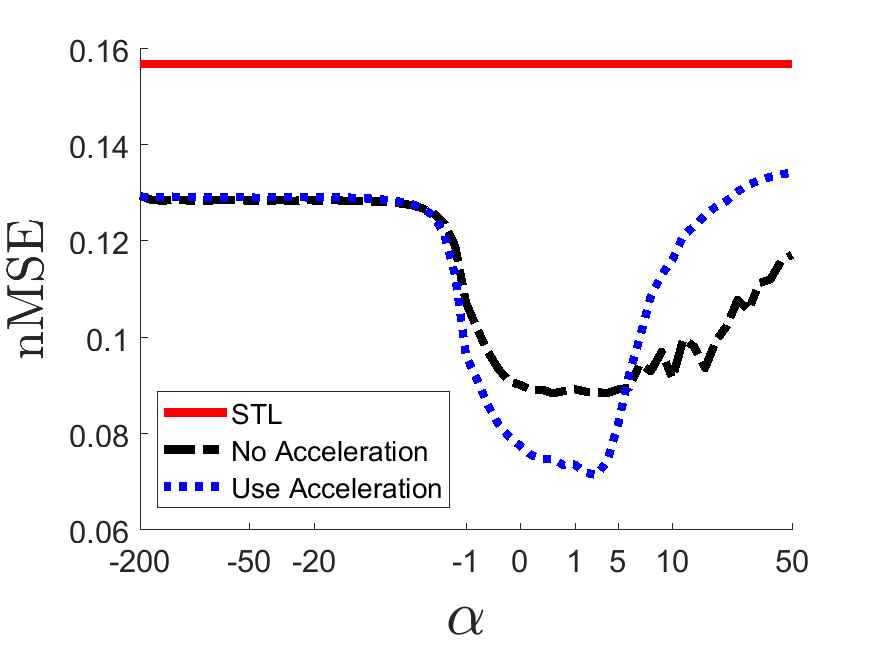}}
\subfigure[Strong convexity]{\includegraphics[width=1.7in]{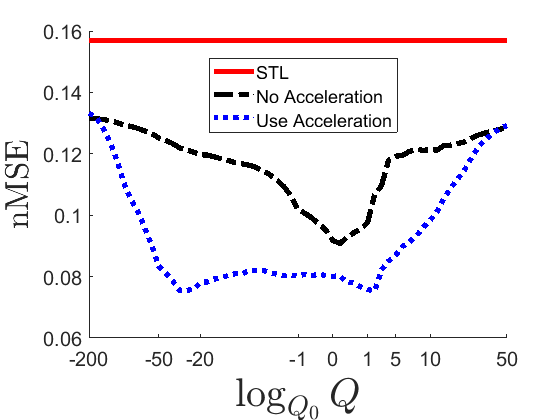}}
\caption{Evaluations for privacy-budget allocation strategies. In (a), we set $\epsilon_t = \Theta(t^{\alpha})$, for $t\in[T]$; in (b), we set $\epsilon_t = \Theta(Q^{-t})$, for $t\in[T]$. $Q_0 = 1-\sqrt{\mu} \approx 0.9684$. The results shown are averages of $100$ runs with $\epsilon = 0.1$. For the non-private MTL method, the nMSE was $0.0140$. }\label{fig:syn_budget_allocation}
\end{figure}
}
{
\subsubsection{Noise-to-Signal Ratio}
Based on the setting in Section \ref{sec:exp_ba_simu}, the noise-to-signal ratios under the best privacy-budget allocation strategy (using acceleration and considering basic convexity) are shown in Fig. \ref{fig:syn_noise_ratio}, {in which we executed Algorithm 2 on the synthetic data set with the low-rank pattern.} In contrast, for DP-MTRL, $\log_{10}(\|\E\|_F/\|\widetilde{\SSigma}^{(t)}\|_F) = 0.2670\pm 0.0075$ under the best iteration number $T=1$. The output model matrices of DP-MTRL and our method are shown in Fig. \ref{fig:syn_relation} (e) and (f), and their respective covariance matrices are shown in Fig. \ref{fig:syn_relation} (b) and (c), respectively. These plots suggest that the high levels of noises added in our method had little influence on the output model matrix and the pattern in its covariance matrix, because our method adds noise only to the knowledge-sharing process and our method degrades to an STL method under high noise levels (as shown in Proposition \ref{th:prop_better_than_STL}). {In contrast, in DP-MTRL, the output model matrix and the pattern in the covariance matrix are significantly affected or even destroyed because the noise was added directly to the model matrix, resulting in negative side-effects. This result may also have occurred because DP-MTRL is \emph{a local private learning algorithm}, which needs a much larger $m$ to achieve acceptable utility (see the discussion in Section \ref{subsec:related_def}).}
\begin{figure}[t!]
\centering
\subfigure{\includegraphics[width=1.7in]{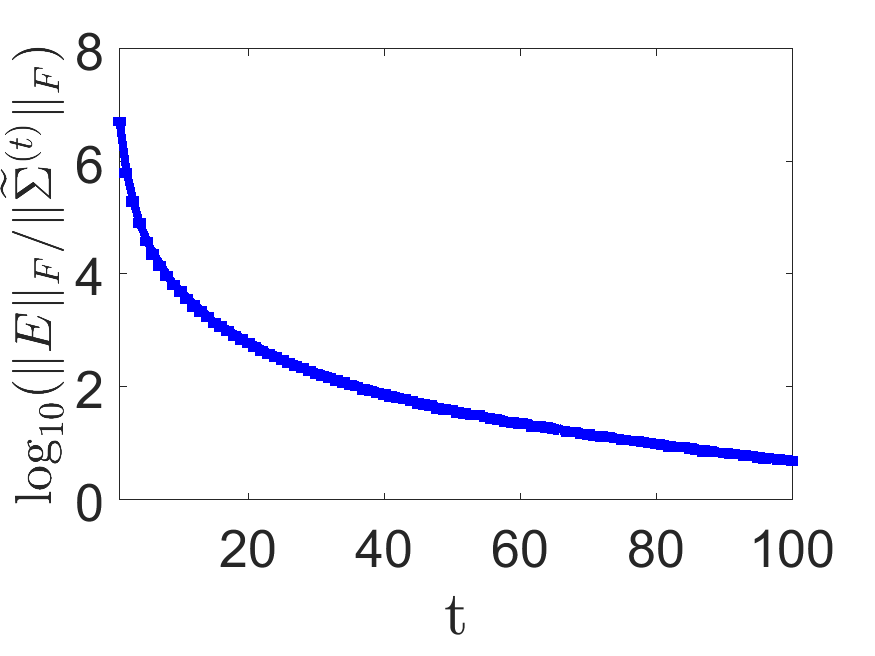}}
\caption{Noise-to-signal ratios over the iterations of Algorithm \ref{alg:MP-MTL-LR}. The results shown are averages of $100$ runs with $\epsilon = 0.1$.}\label{fig:syn_noise_ratio}
\end{figure}
}

{
\subsubsection{Privacy-Accuracy Tradeoff}
In Fig. \ref{fig:syn}, the performances of both of our MP-MTL algorithms (i.e., Algorithms \ref{alg:MP-MTL-LR} and \ref{alg:MP-MTL-GS}) fall between those of the non-private STL and non-private MTL methods, suggesting that our methods are useful as MTL methods but may be affected by the introduced noise. In Fig. \ref{fig:syn} (a), Algorithm \ref{alg:MP-MTL-GS} underperforms compared with Algorithm \ref{alg:MP-MTL-LR}, because the true model matrix is not group-sparse. DP-MTRL outperforms the STL method and our Algorithm \ref{alg:MP-MTL-GS} when $\epsilon$ is large because it suits the true model matrix, in which the relatedness among tasks is modeled by a graph. In Fig. \ref{fig:syn} (b), the true model matrix is group-sparse and is not suitable for DP-MTRL; hence, DP-MTRL underperforms compared with the STL method even when $\epsilon$ is large. Algorithm \ref{alg:MP-MTL-LR} rivals Algorithm \ref{alg:MP-MTL-GS} because the true model matrix is also low-rank. In both panels of Fig. \ref{fig:syn}, Algorithm \ref{alg:MP-MTL-LR} rivals the non-private MTL when $\epsilon = 10$.

\begin{figure}[t!]
\centering
\subfigure[Low-rank pattern]{\includegraphics[width=2.7in]{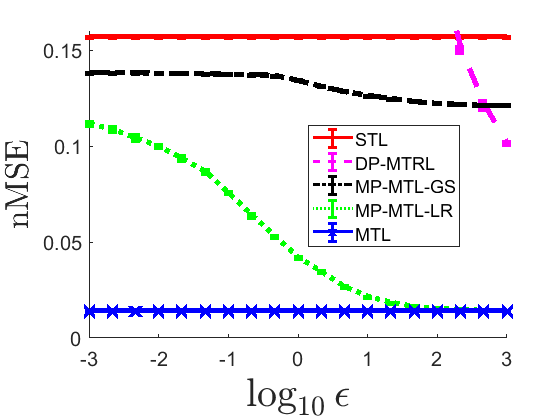}}
\subfigure[Group-sparse pattern]{\includegraphics[width=2.7in]{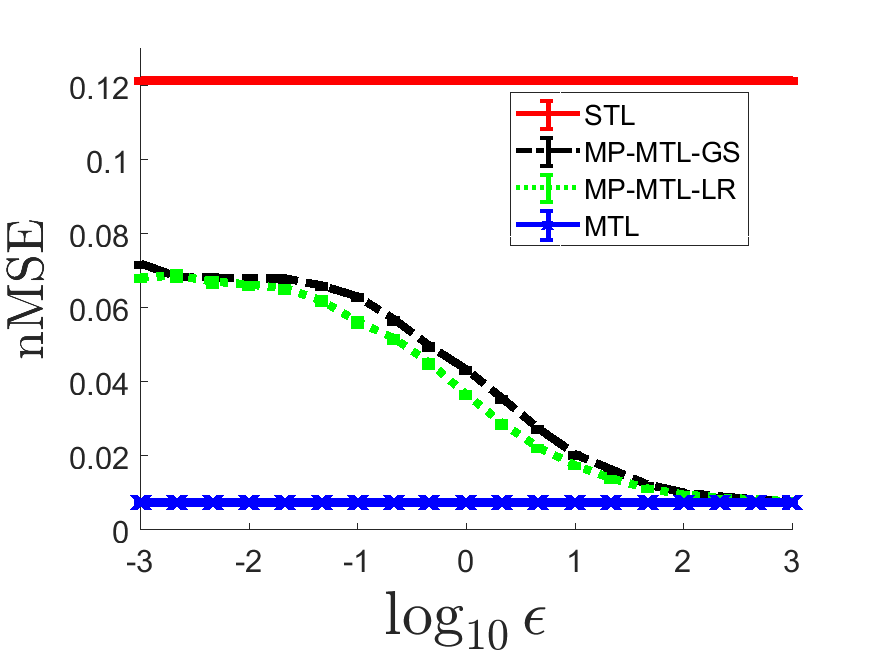}}
\caption{Privacy-accuracy tradeoff on synthetic datasets. For (a), the data associated with the low-rank model matrix were used; for (b), the data associated with the group-sparse model matrix were used. MP-MTL-LR denotes Algorithm \ref{alg:MP-MTL-LR}, MP-MTL-GS denotes Algorithm \ref{alg:MP-MTL-GS}, and STL denotes the $\ell_2$-norm-penalized STL method. In both panels, STL and MTL denote non-private methods. In (b), the nMSEs of DP-MTRL are above $0.16$; in both panels, the nMSEs of DP-AGGR are above $0.78$. More detailed performances of DP-MTRL and DP-AGGR are presented in Fig. \ref{fig:syn_detail_baseline}.}\label{fig:syn}
\end{figure}
}

{
Fig. \ref{fig:syn_detail_baseline} shows the detailed performances for DP-MTRL and DP-AGGR corresponding to those in Fig. \ref{fig:syn}. Fig. \ref{fig:syn_detail_baseline} (c) is used to show that the accuracy of DP-AGGR grows with $\epsilon$ under the same setting as in Fig. \ref{fig:syn_detail_baseline} (b). As discussed previously, DP-AGGR performs only model averaging, which is not suitable for the true model matrices in both settings of Fig. \ref{fig:syn_detail_baseline} (a) and (b); hence, the accuracies of DP-AGGR are much worse than those of the respective STL methods.

\begin{figure}[t!]
\subfigure[Low-rank pattern]{\includegraphics[width=1.7in]{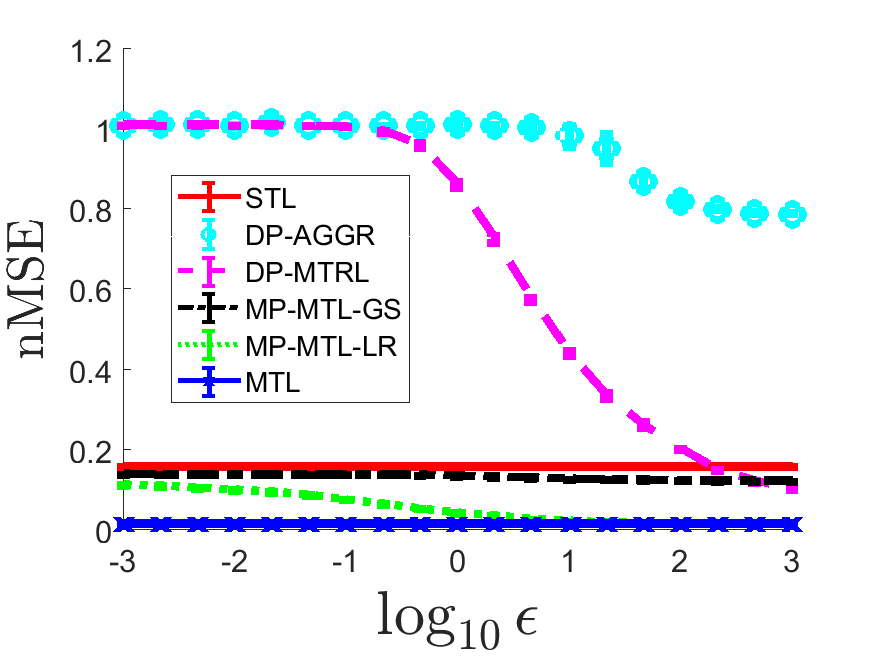}}
\subfigure[Group-sparse pattern]{\includegraphics[width=1.7in]{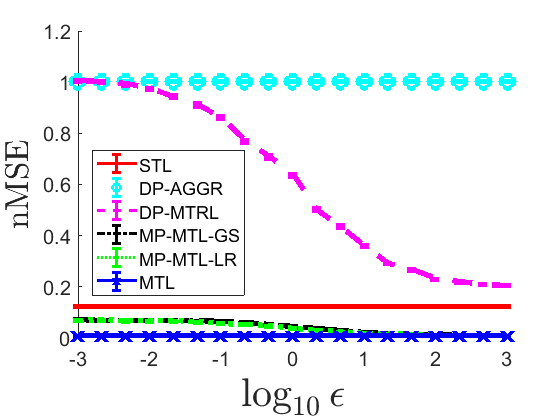}}
\subfigure[Group-sparse pattern]{\includegraphics[width=1.7in]{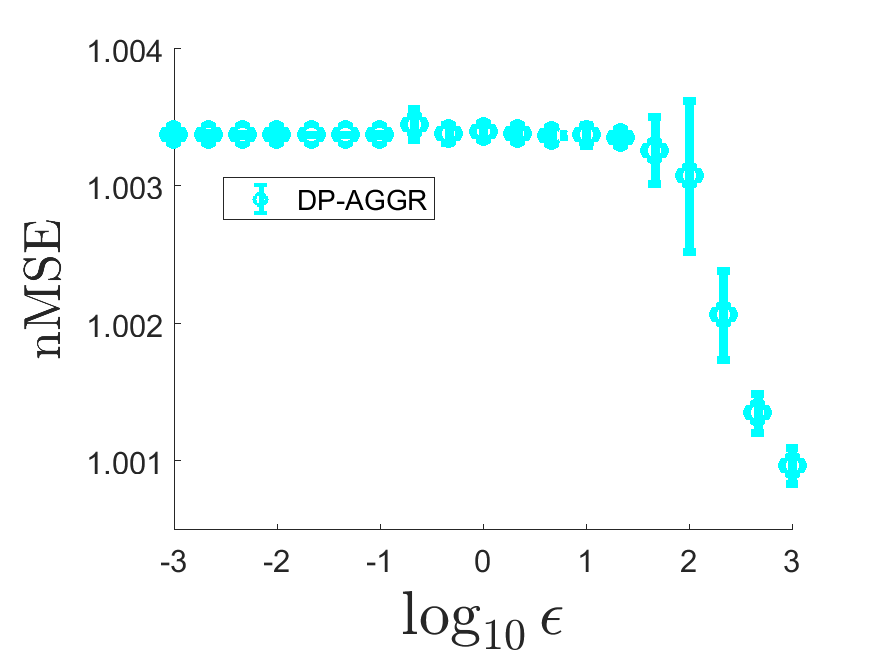}}
\caption{Detailed privacy-accuracy tradeoff on synthetic datasets for DP-MTRL and DP-AGGR. For (a), the data associated with the low-rank model matrix were used; for (b) and (c), the data that associated with the group-sparse model matrix were used. In (c), the plot shows the same performances of DP-AGGR as those in (b) but with a finer vertical axis. Other settings are the same as those used for Fig. \ref{fig:syn}.}\label{fig:syn_detail_baseline}
\end{figure}
}

{
\subsubsection{Varing the Number of Tasks}
Based on the setting in Section \ref{sec:exp_ba_simu}, the average performances of the first $20$ of the $320$ total tasks are shown in Fig. \ref{fig:syn_multim} under different numbers of training tasks. The accuracy increases with the number of tasks involved, which is consistent with our utility analyses. {The standard deviation of each result is plotted in Fig. \ref{fig:syn_multim}, showing that the increments are statistically significant.}

\begin{figure}[t!]
\centering
\subfigure{\includegraphics[width=2.7in]{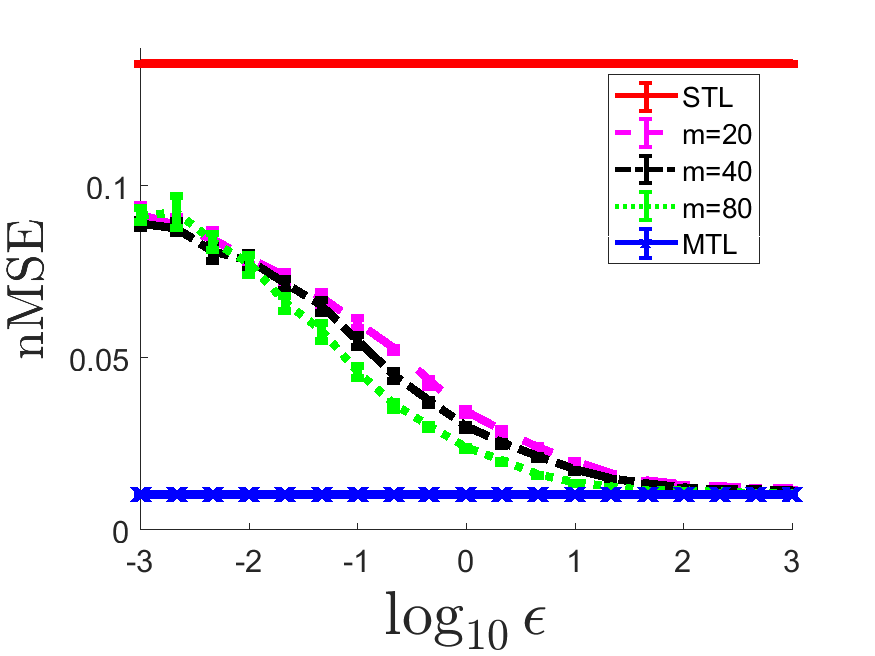}}
\caption{Behaviors based on the number of tasks $m$ used for training. We used $320$ tasks for MTL training.}\label{fig:syn_multim}
\end{figure}
}

\subsection{Application}

\subsubsection{Data Description}

We also evaluate the considered methods on the following two real datasets.

\noindent\textbf{\emph{\underline{School Data}}}. The School dataset\footnote{http://www.cs.ucl.ac.uk/staff/a.argyriou/code/} is a popular dataset for MTL~\citep{gong2012robust} that consists of the exam scores of 15,362 students
from 139 secondary schools. Each student is described by 27 attributes, including both school-specific information and student-specific information such as gender and ethnic group.
The problem of predicting exam scores for the students can be formulated as an MTL problem: the number of tasks is $m=139$, the data dimensionality is $d=27$, and the number of data samples is $\sum_in_i = 15,362$.

\noindent\textbf{\emph{\underline{LSOA II Data}}}. These data are from the Second Longitudinal Study of Aging (LSOA II) \footnote{https://www.cdc.gov/nchs/lsoa/lsoa2.htm.}. LSOA II was a collaborative study conducted by the National Center for Health Statistics (NCHS) and the National Institute of Aging from 1994 to 2000. A national representative sample of $9,447$ subjects of $70$ years of age and older were selected and interviewed. Three separate interviews were conducted with each subject, one each during the periods of 1994--1996, 1997--1998, and 1999--2000, referred to as WAVE 1, WAVE 2, and WAVE 3, respectively. Each wave of interviews included multiple modules covering a wide range of assessments. We used data from WAVE 2 and WAVE 3, which include a total of $4,299$ sample subjects and 44 targets (each subject corresponded to 44 targets). We extracted $188$ features from the WAVE 2 interviews.
{
The targets include $m=41$ binary outcomes used in this study. These outcomes fall into several categories: 7 measures of fundamental daily activity, 13 of extended daily activity, 5 of social involvement, 8 of medical condition, 4 of cognitive ability, and 4 of sensation condition.

The features include demographic, family structure, daily personal care, medical history, social activity, health opinions, behavior, nutrition, health insurance and income and asset attributes, the majority of which are binary values.

Both the targets and the features have missing values due to non-responsed and questionnaire filtering. The average missing value rates of the targets and features are 13.7\% and 20.2\%, respectively. To address the missing values among the features, we adopted the following preprocessing procedure. For the continuous features, missing values were imputed with the sample mean. For binary features, it is better to treat the missing values as a third category because the absence of a value may also carry important information. Therefore, two dummy variables were created for each binary feature with missing values (no third variable is necessary in such a case) resulting in a total of $d=295$ features. To address the missing values among the targets, we included the samples associated with the observed targets for each task, resulting in $\max_{i\in[m]}n_i = 3,473$.

}

For both the real-world datasets, We randomly selected 30\% of the samples from each task to form the training set and used the remaining samples as the test set. For all the tasks, each data point was normalized to have a unit length.

\subsubsection{Privacy-Accuracy Tradeoff}
\begin{figure}[t!]
\centering
\subfigure[School Data]{\includegraphics[width=1.7in]{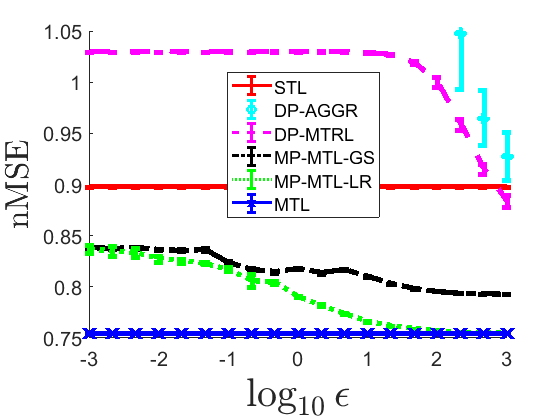}}
\subfigure[LSOA II Data]{\includegraphics[width=1.7in]{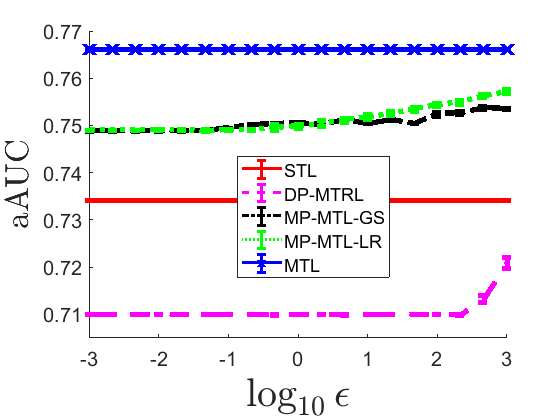}}
\caption{Privacy-accuracy tradeoff on real-world datasets. In both panels, MTL denotes the method with the best performance among the four non-private MTL methods proposed by~\citet{ji2009accelerated},~\citet{liu2009multi},~\citet{zhang2010convex} and DP-AGGR without perturbations; MP-MTL-LR denotes Algorithm \ref{alg:MP-MTL-LR}, whereas MP-MTL-GS denotes Algorithm \ref{alg:MP-MTL-GS}; STL denotes the method with the better performance between the $\ell_1$- and $\ell_2$-regularized methods. In (b), the aAUCs of DP-AGGR are below $0.66$. The detailed performances of DP-AGGR are presented in Fig. \ref{fig:real_detail_baseline}.}\label{fig:real}
\end{figure}

From Fig. \ref{fig:real}, we can observe results similar to those seen in Fig. \ref{fig:syn}. In addition, our MP-MTL algorithms outperform the baseline MP-MTL methods, DP-MTRL and DP-AGGR, especially when $\epsilon$ is small.
{DP-AGGR underperforms compared with the STL method because its model averaging approach assumes that the tasks are homogeneous. In Fig. \ref{fig:real} (b), the
aAUC values of DP-MTRL and our Algorithms (\ref{alg:MP-MTL-LR} and \ref{alg:MP-MTL-GS}) increase slowly because the feature dimension is large and the number of tasks is insufficient, which is consistent with our utility analyses.}

{
Fig. \ref{fig:real_detail_baseline} shows the detailed performances of DP-AGGR. In Fig. \ref{fig:real_detail_baseline} (b), because the dimension is large and the number of tasks is insufficient, the accuracy of DP-AGGR barely grows with $\epsilon$.

\begin{figure}[t!]
\centering
\subfigure[School Data]{\includegraphics[width=1.7in]{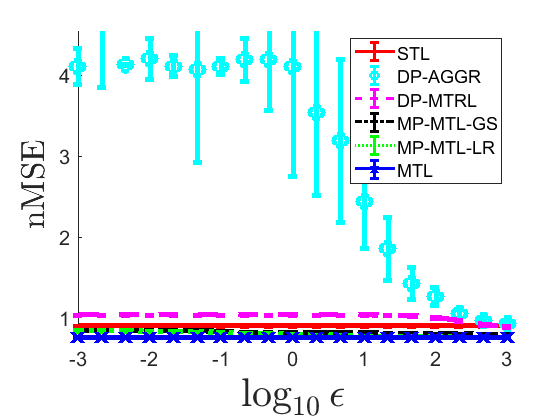}}
\subfigure[LSOA II Data]{\includegraphics[width=1.7in]{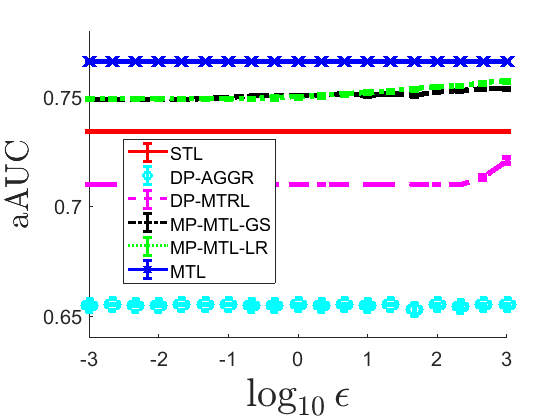}}
\caption{Detailed privacy-accuracy tradeoff on real-world datasets for DP-AGGR. All the settings are the same as those in Fig. \ref{fig:real}. }\label{fig:real_detail_baseline}
\end{figure}
}

{
Because the MTL behavior may change when the training-data percentage (the size of the training data divided by the size of the entire dataset) changes, we evaluated the methods on both real-world datasets at different training-data percentages and achieved similar results; see the supplementary material for more details.}

\section{Conclusions}
\label{sec:conclusion}
In this paper, we discussed the potential security risks of multi-task learning approaches and presented a rigorous mathematical formulation of the model-protected multi-task learning (MP-MTL) problem. We proposed an algorithmic framework for implementing MP-MTL along with two concrete framework instantiations that learn the low-rank and group-sparse patterns in the model matrix. {We demonstrated that our algorithms are guaranteed not to underperform compared with single-task learning methods under high noise levels. Privacy guarantees were provided. The utility analyses suggested that both the strong-convexity condition and the acceleration strategy improve the utility bounds, { and that the acceleration strategy also improves the runtime}. A utility analysis for privacy-budget allocation yielded a recommendation for privacy budgets that are non-decreasing over the iterations. The experiments demonstrated that our algorithms significantly outperform baseline methods constructed by existing privacy-preserving MTL methods on the proposed model-protection problem.} Some interesting future research directions include developing concrete MP-MTL algorithms for other MTL approaches and other optimization schemes.
\ifCLASSOPTIONcompsoc
\section*{Acknowledgments}
\else
\section*{Acknowledgment}
\fi

This research was supported in part by the National Science Foundation (NSF) under grants IIS-1565596, III-1615597 and IIS-1650723, in part by the Office of Naval Research (ONR) under grant number N00014-14-1-0631, and in part by the National Institutes of Health (NIH) under grants R00LM011392 and R21LM012060. In addition, we would like to thank Yuxiang Wang from the School of Computer Science at Carnegie Mellon University for his valuable comments on how to improve the properties of iterative MP-MTL algorithms.

\ifCLASSOPTIONcaptionsoff
\newpage
\fi

{\small
\bibliographystyle{abbrvnat}
\bibliography{sigproc} %
}

\begin{IEEEbiography}[{\includegraphics[width=1in,height=1.25in,clip,keepaspectratio]{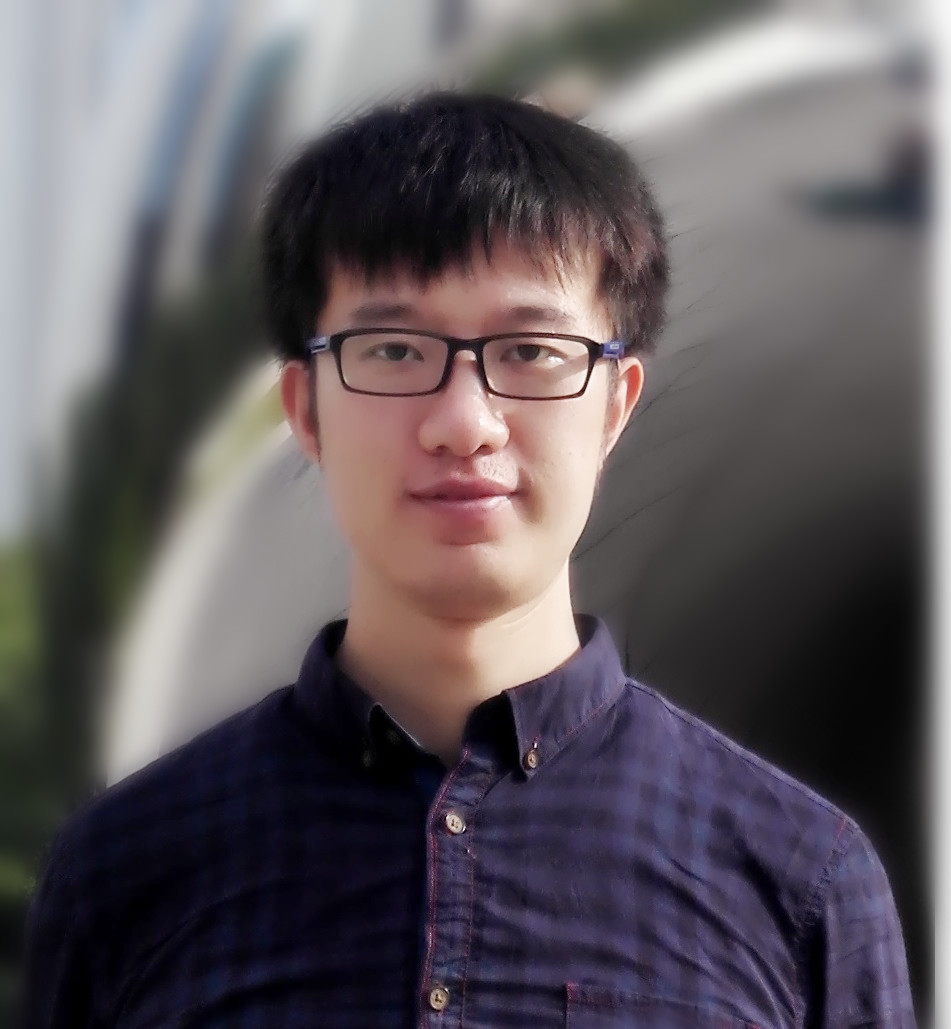}}]{Jian Liang}
received his B.S. degree in automation from the Huazhong University of Science and Technology, Wuhan, China, in 2012. He is currently working toward his Ph.D. degree in the Department of Automation at Tsinghua University, Beijing. His research interests lie in the areas of multi-task learning and time-series data mining. He won the Best Short Paper Award at the 2016 IEEE International Conference on Healthcare Informatics (ICHI).
\end{IEEEbiography}

\begin{IEEEbiography}[{\includegraphics[width=1in,height=1.25in,clip,keepaspectratio]{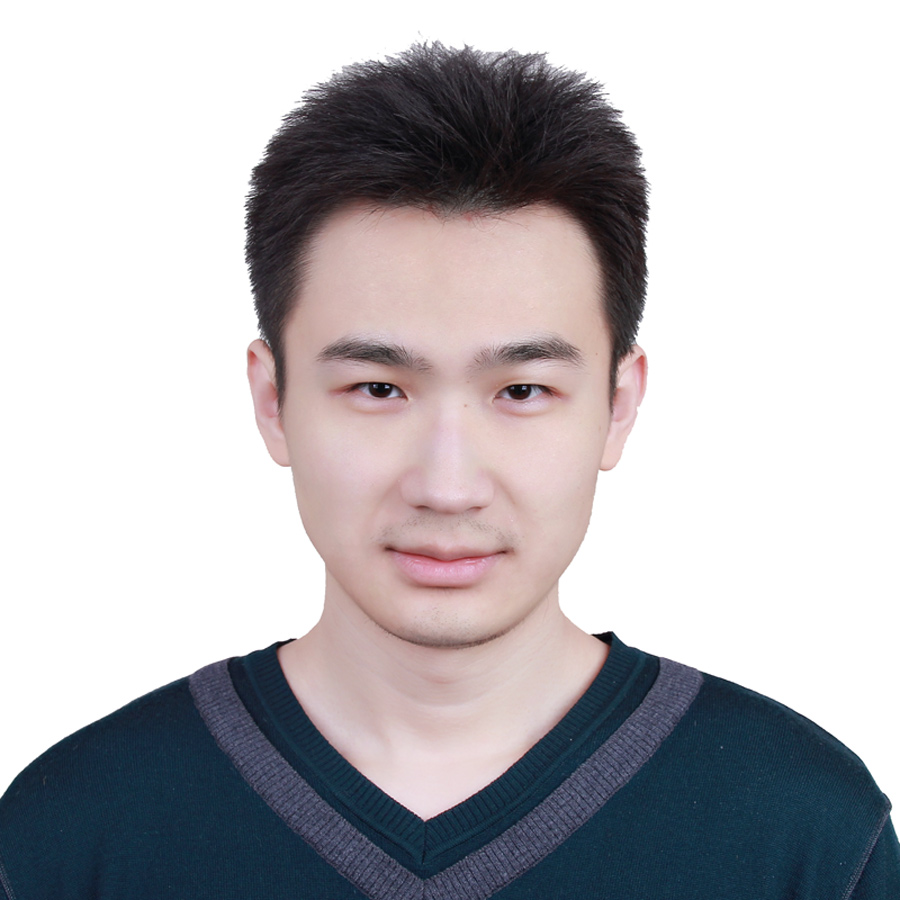}}]{Ziqi Liu}
Ziqi Liu is currently a Ph.D. student at Xi’an Jiaotong University under advisor Professor Qinghua Zheng. His research interests lie in scalable machine learning and nonparametric modeling. The major awards he has received include the WSDM’16 Best Research Paper Award.
\end{IEEEbiography}

\begin{IEEEbiography}[{\includegraphics[width=1in,height=1.25in,clip,keepaspectratio]{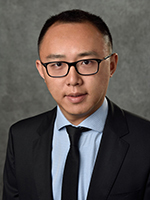}}]{Jiayu Zhou}
Jiayu Zhou is currently an assistant professor in the Department of Computer Science and Engineering at Michigan State University. He received his Ph.D. degree in computer science from Arizona State University in 2014. He has broad research interests in the fields of large-scale machine learning and data mining as well as biomedical informatics. He has served as a technical program committee member for premier conferences such as NIPS, ICML, and SIGKDD. His papers have received the Best Student Paper Award at the 2014 IEEE International Conference on Data Mining (ICDM), the Best Student Paper Award at the 2016 International Symposium on Biomedical Imaging (ISBI) and the Best Paper Award at IEEE Big Data 2016.
\end{IEEEbiography}

\begin{IEEEbiography}[{\includegraphics[width=1in,height=1.25in,clip,keepaspectratio]{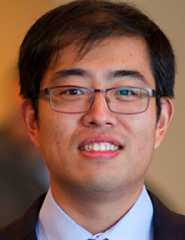}}]{Xiaoqian Jiang}
is an assistant professor in the Department of Biomedical Informatics at the University of California, San Diego. He received his Ph.D. in computer science from Carnegie Mellon University. He is an associate editor of BMC Medical Informatics and Decision Making and serves as a member of the editorial board of the Journal of the American Medical Informatics Association. He works primarily in the fields of health data privacy and predictive models in biomedicine. Dr. Jiang was a recipient of the NIH K99/R00 award and won the Distinguished Paper Award from the American Medical Informatics Association Clinical Research Informatics (CRI) Summit in 2012 and 2013.
\end{IEEEbiography}

\begin{IEEEbiography}[{\includegraphics[width=1in,height=1.25in,clip,keepaspectratio]{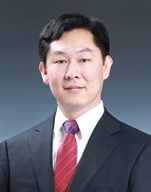}}]{Changshui Zhang}
received his B.S. degree from Peking University, Beijing, China, in 1986 and his Ph.D. degree from Tsinghua University, Beijing, China, in 1992. He is currently a professor in the Department of Automation at Tsinghua University. He is a member of the editorial board of Pattern Recognition and an IEEE Fellow. His research interests include artificial intelligence, image processing, pattern recognition, machine learning and evolutionary computation.
\end{IEEEbiography}

\begin{IEEEbiography}[{\includegraphics[width=1in,height=1.25in,clip,keepaspectratio]{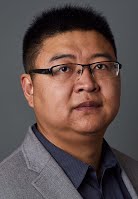}}]{Fei Wang}
is an assistant professor in the Division of Health Informatics of the Department of Healthcare Policy and Research at Cornell University. His major research interest lies in data analytics and its applications in health informatics. His papers have received over 3,700 citations, with an H-index of 33. He won the Best Student Paper Award at ICDM 2015, received a Best Research Paper nomination at ICDM 2010, and the Marco Romani Best Paper nomination at AMIA TBI 2014, and his papers were selected as Best Paper finalists at SDM 2011 and 2015. Dr. Wang is an active editor of the journal Data Mining and Knowledge Discovery, an associate editor of the Journal of Health Informatics Research and Smart Health, and a member of the editorial boards of Pattern Recognition and the International Journal of Big Data and Analytics in Healthcare. Dr. Wang is also the vice chair of the KDD working group at AMIA.
\end{IEEEbiography}

\newpage
\begin{center}
\textbf{\large Supplemental Materials: Model-Protected Multi-Task Learning}
\end{center}
\appendices

\setcounter{section}{0}

\small


\section{Model-Decomposed MP-MTL Methods}\label{sec:Model-Decomposed}


In this section, we consider the extension of our MP-MTL framework for MTL methods using the decomposed parameter/model matrix. Specifically, we focus on the following problem, where the trace norm is used for knowledge sharing across tasks and the $\|\cdot\|_1$ norm (sum of the $\ell_1$ norm for vectors) is used for entry-wise outlier detection, as described in Algorithm \ref{alg:MP-MTL-LR-SP}.

\begin{equation}\label{eq:MTL_Sparse_Trace}
\begin{split}
   \min_{\W} \ &\sum\nolimits_{i=1}^m\calL_i(\X_i\w_i,\y_i) + \lambda_1 \|\P\|_{*} + \lambda_2 \|\Q\|_{1}\\
   s.t. \ &\W = \P+\Q,
\end{split}
\end{equation}
where $\P,\Q\in\mathbb{R}^{d \times m}$.

\begin{algorithm} [htb]
\caption{Model-Protected low-rank and group-sparse (MP-LR-SP) Estimator}
\label{alg:MP-MTL-LR-SP}
{\small
\begin{algorithmic}[1]
\REQUIRE{Datasets $(\X^m,\y^m) = \{(\X_1,\y_1),\ldots,(\X_m,\y_m)\}$, where $\forall i\in[m], \ \X_i\in \mathbb{R}^{n_i \times d}$ and $\y_i\in \mathbb{R}^{n_i \times 1}$. Privacy loss $\epsilon , \delta  \geq 0$. Number of iterations $T$. Step size $\eta$.  Regularization parameter $\lambda_1,\lambda_2>0$. Norm clipping parameter $K>0$.
Acceleration parameters $\{\beta_t\}$.
Initial models of tasks $\W^{(0)}$.}
\ENSURE{$\widehat{\W}^{(1:T)}$ .}
\STATE Set $\{\epsilon_t\}$ such that { $CB(\{\epsilon_t\},\delta)\leq \epsilon$, where $CB(\{\epsilon_t\},\delta)$ is the composition bound of $\{\epsilon_t\}$.}
\STATE Let $\P^{(0)}= \Q^{(0)} = \widehat{\Q}^{(0)}=\W^{(0)}$.
\FOR{$t=1:T$}
\STATE {Norm clipping: $\tilde{\p}_i^{(t-1)} = {\p}_i^{(t-1)}/\max(1,\frac{\|{\p}_i^{(t-1)}\|_2}{K})$ for all $i\in[m]$.} Let $\widehat{\P}^{(0)} = \widetilde{\P}^{(0)}$.
\STATE  $\widetilde{\SSigma}^{(t)} = \widetilde{\P}^{(t-1)}(\widetilde{\P}^{(t-1)})\trans$.\\
\STATE  $\SSigma^{(t)} = \widetilde{\SSigma}^{(t-1)} + \E$, where $\E\sim W_d(d+1,\frac{K^2}{2\epsilon_t}\I_d)$ is a sample of the Wishart distribution.\\
\STATE Perform SVD decomposition: $\U\LLambda\U\trans =  {\SSigma}^{(t)}$.
\STATE Let $  \S_{\eta\lambda_1} $ be a diagonal matrix and let $\S_{\eta\lambda_1,ii} = \max\{0,   1- \eta\lambda_1/\sqrt{\LLambda_{ii}}  \}$ for $i=1,\ldots,\min\{d,m\}$.\\
\STATE {$\M^{(t)}=\U\S_{\eta\lambda_1}\U\trans$.}
\STATE Let $\hat{\p}_i^{(t)} = \M^{(t)}\tilde{\p}_i^{(t-1)}$ for all $i\in[m]$.\\
\STATE Let $\hat{\q}_i^{(t)} = \mbox{sign}(\q_i^{(t-1)}) \circ \max\{0, |\q_i^{(t-1)}| - \eta\lambda_2\} $ for all $i\in[m]$, where $\circ$ denotes the entry-wise product.\\
\STATE Let $\widehat{\W}^{(t)} = \widehat{\P}^{(t)} + \widehat{\Q}^{(t)}$.\\{
\STATE Let $ {\z}_{i,p}^{(t)} =   \hat{\p}_i^{(t)} + \beta_t (\hat{\p}_i^{(t)} - \hat{\p}_i^{(t-1)})$ for all $i\in[m]$. \\
\STATE Let $ {\z}_{i,q}^{(t)} =   \hat{\q}_i^{(t)} + \beta_t (\hat{\q}_i^{(t)} - \hat{\q}_i^{(t-1)})$ for all $i\in[m]$. }\\
\STATE ${\p}_i^{(t)}= {\z}_{i,p}^{(t)} - \eta \frac{\partial \mathcal{L}_i(\X_i({\z}_{i,p}^{(t)}+{\z}_{i,q}^{(t)}),\y_i)}{\partial \hat{\p}_i^{(t)}}$ for all $i\in[m]$.\\
\STATE ${\q}_i^{(t)}= {\z}_{i,q}^{(t)} - \eta \frac{\partial \mathcal{L}_i(\X_i({\z}_{i,p}^{(t)}+{\z}_{i,q}^{(t)}),\y_i)}{\partial \hat{\q}_i^{(t)}}$ for all $i\in[m]$.\\
\ENDFOR
\end{algorithmic}}
\end{algorithm}

We note that in Algorithm \ref{alg:MP-MTL-LR-SP}, the role of $\P$ is the same as the role of $\W$ in Algorithm \ref{alg:MP-MTL-LR}, and the additional procedures introduced to update $\Q$ are still STL algorithms. As such, we have the result in Corollary \ref{th:th_MP-MTL-LR-SP}.

\begin{corollary}\label{th:th_MP-MTL-LR-SP}
Algorithm \ref{alg:MP-MTL-LR-SP} is an $(\epsilon,\delta)$ MP-MTL algorithm.
\end{corollary}
\begin{proof}
For simplicity, we omit the symbol $\calB$ to denote the input in the conditional events in some equations.

 Use Corollary \ref{th:th_MP-MTL-LR} and Theorem \ref{th:th_MP-MTL}. Given $t\in[T]$, the algorithm $(\widetilde{\P}^{(t-1)},\SSigma^{(1:t-1)})\rightarrow (\M^{(t)},\SSigma^{(t)})$ is an $(\epsilon_t,0)$-differentially private algorithm, where $\M^{(t)} = \U\S_{\eta\lambda}\U\trans$.

 Now, for all $i\in[m]$, applying the \emph{Post-Processing immunity} property (Property \ref{th:lem_post}) for the mapping $f: (\M^{(t)}, \tilde{\p}_{[-i]}^{(t-1)})\rightarrow \hat{\p}_{[-i]}^{(t-1)} $, which does not touch any unperturbed sensitive information of the $i$-th task, we have for any set $\calS \subseteq \mathbb{R}^{d\times(m-1)}$ that
 \begin{align*}
 \bbP( \hat{\p}&_{[-i]}^{(t-1)} \in \calS \mid \widetilde{\P}^{(t-1)} ,\SSigma^{(1:t-1)} ) \\
 \leq &e^{\epsilon_t}\bbP(\hat{\p}_{[-i]}^{(t-1)} \in \calS \mid (\widetilde{\P}')^{(t-1)},\SSigma^{(1:t-1)}  ),
\end{align*}
where $\widetilde{\P}^{(t-1)}$ and $(\widetilde{\P}')^{(t-1)}$ differ only in the $i$-th column.

Then, because in the $t$-th iteration the mapping $ \Q^{(t-1)} \rightarrow \widehat{\Q}^{(t-1)}$ is a deterministic STL algorithm, we have for any set $\calS \subseteq \mathbb{R}^{d\times(m-1)}$ that
 \begin{align*}
 \bbP( \hat{\q}&_{[-i]}^{(t-1)} \in \calS \mid \Q^{(t-1)} ) \\
 = & \bbP( \hat{\q}_{[-i]}^{(t-1)} \in \calS \mid \q_{[-i]}^{(t-1)}  ,\q_{i}^{(t-1)}  ) \\
 = & \bbP( \hat{\q}_{[-i]}^{(t-1)} \in \calS \mid \q_{[-i]}^{(t-1)} ,(\q'_{i})^{(t-1)} ) \\
 = & e^0\bbP(\hat{\q}_{[-i]}^{(t-1)} \in \calS \mid (\Q')^{(t-1)} ) +0 ,
\end{align*}
where $\Q^{(t-1)}$ and $(\Q')^{(t-1)}$ differ only in the $i$-th column.

Then applying \emph{Combination} property (Property \ref{th:lem_comb}), we have for any set $\calS \subseteq \mathbb{R}^{d\times(m-1)}\times \mathbb{R}^{d\times(m-1)}$
 \begin{align*}
 \bbP( (\hat{\p}&_{[-i]}^{(t-1)} ,\hat{\q}_{[-i]}^{(t-1)}) \in \calS \mid \P^{(t-1)} ,\SSigma^{(1:t-1)}, \Q^{(t-1)} ) \\
\leq &e^{\epsilon_t}\bbP((\hat{\p}_{[-i]}^{(t-1)} ,\hat{\q}_{[-i]}^{(t-1)}) \in \calS \mid (\P')^{(t-1)},\SSigma^{(1:t-1)} ,(\Q')^{(t-1)} ),
\end{align*}

Because the mapping$(  \widehat{\P}^{(t-1)}, \widehat{\Q}^{(t-1)}, \calD^m)\rightarrow (\widehat{\P}^{(t)}, \widehat{\Q}^{(t)})$ is a deterministic STL algorithm, applying the \emph{Post-Processing immunity} property (Property \ref{th:lem_post}), we further have for any set $\calS \subseteq \mathbb{R}^{d\times(m-1)}\times \mathbb{R}^{d\times(m-1)}$ that
\begin{align*}
 \bbP(& (\hat{\p}_{[-i]}^{(t)} ,\hat{\q}_{[-i]}^{(t)}) \in \calS \mid \P^{(t-1)} ,\SSigma^{(1:t-1)}, \Q^{(t-1)}, \calD^m ) \\
\leq &e^{\epsilon_t}\bbP((\hat{\p}_{[-i]}^{(t)} ,\hat{\q}_{[-i]}^{(t)}) \in \calS \mid (\P')^{(t-1)},\SSigma^{(1:t-1)} ,(\Q')^{(t-1)} ,(\calD')^m),
\end{align*}
where $(\calD')^m$ differs from $\calD^m$ in the entire dataset of the $i$-th task.

Now, using Theorem \ref{th:th_MP-MTL}, for $t=1,\ldots,T$, we again take the $t$-th dataset $\widetilde{\calD}_t = \{(\p_1^{(t-1)},\q_1^{(t-1)}, \calD_1),\ldots,(\p_m^{(t-1)},\q_m^{(t-1)} \calD_m)\}$ and denote $\vartheta_{t,i} = (\hat{\q}_{[-i]}^{(t)},\hat{\q}_{[-i]}^{(t)}, \M^{(t)},\SSigma^{(t)})\in \calC_{t,i}$. Given the fact that $\P^{(t)} = \widehat{\P}^{(t)}$ and $\Q^{(t)} = \widehat{\Q}^{(t)}$ for all $t\in[T]$, we have for any set $\calS_{t,i}\subseteq\calC_{t,i} $ that
\begin{align*}
 \bbP( \vartheta&_{t,i} \in \calS_{t,i}\mid \widetilde{\calD}_t, \vvartheta_{1:t-1} ) \\
 \leq &e^{\epsilon_t}\bbP( \vartheta_{t,i} \in \calS_{t,i}\mid \widetilde{\calD}'_t, \vvartheta_{1:t-1} ),
\end{align*}
where $\widetilde{\calD}_t$ and $\widetilde{\calD}'_t$ are two adjacent datasets that differ in a single entry, the $i$-th ``data instance'' $(\p_i^{(t-1)},\q_i^{(t-1)}, \calD_i = (\X_i,\y_i))$, and
\begin{numcases}{\vvartheta_{1:t-1} = }
\nonumber \emptyset, & $t=1$\\
\nonumber (\vartheta_{1,1},\dots,\vartheta_{1,m}) \ldots,(\vartheta_{t-1,1},\dots,\vartheta_{t-1,m}), & $t\geq2$.
\end{numcases}

This renders the algorithm in the $t$-th iteration as an $(\epsilon_t,0)$-differentially private algorithm.

Then, again by the \emph{Adaptive composition} property (Property \ref{th:lem_Ad_comp}), for all $i\in[m]$ and for any set $\calS'\subseteq\bigotimes_{t=1}^{T}\mathcal{C}_{t_i}$, we have
\begin{align*}
  \mathbb{P} ((\vartheta&_{1,i},\cdots,\vartheta_{T,i})\in \mathcal{S}'  \mid \bigcap_{t=1}^T(\calB_t = (\widetilde{\calD}_t, \vvartheta_{1:t-1}))  )\\
  \leq & e^{\tilde{\epsilon}}\mathbb{P}((\vartheta_{1,i},\cdots,\vartheta_{T,i})\in \mathcal{S}'\mid \bigcap_{t=1}^T(\calB_t = (\widetilde{\calD}'_t, \vvartheta_{1:t-1}) ))\\
   &+ \delta,
\end{align*}
where for all $t\in[T]$, $\calB_t$ denotes the input for the $t$-th iteration.

Finally, for all $t\in[T]$, taking $\theta_t = (\vartheta_{t,1},\dots,\vartheta_{t,m})$  and given the fact that $\widehat{\W}^{(t)} = \widehat{\P}^{(t)} + \widehat{\Q}^{(t)}$, we have for any set $\calS\subseteq \mathbb{R}^{d\times(m-1)\times T}$ that
\begin{align*}
 \bbP( &  \hat{\w}_{[-i]}^{(1:T)}  \in \calS  \mid \bigcap_{t=1}^T \calB_t = (\W^{(t-1)}, \calD^m, \ttheta_{1:t-1}) ) \\
\leq   &e^{\epsilon} \bbP(  \hat{\w}_{[-i]}^{(1:T)} \in \calS \mid \bigcap_{t=1}^T \calB_t = ((\W')^{(t-1)}, (\calD')^m, \ttheta_{1:t-1})) \\
&+ \delta,
\end{align*}
where $(\W')^{(t-1)}$ are associated with the setting in which the $i$-th task has been replaced.
\end{proof}

\begin{remark}
Based on Algorithm \ref{alg:MP-MTL-LR-SP}, this will result in a similar procedure and identical theoretical results with respect to privacy by replacing the trace norm with the $\ell_{2,1}$ norm to force group sparsity in $\P$ or by replacing the $\|\cdot\|_1$ norm with the $\ell_{1,2}$ norm (sum of the $\ell_2$ norm of column vectors) or $\|\cdot\|_F^2$ (square of the Frobenius norm).
\end{remark}

\section{MP-MTL Framework with Secure Multi-Party Computation}\label{sec:SMC_framework}
{\citet{pathak2010multiparty} considered the demand for secure multi-party computation (SMC): protecting data instances from leaking to the curator and leaking between tasks during joint learning.} However, by Proposition \ref{th:prop_DP-MP-iter}, the method of \citet{pathak2010multiparty} may introduce excess noise to protect both the data instances and the models simultaneously. To avoid unnecessary noise, we consider a divide-and-conquer strategy to ensure privacy for a single data instance and the model separately. Specifically, in each iteration of the MP-MTL algorithms, we perform private sharing after introducing the perturbation to the parameter matrix to protect a single data instance, as described in Algorithm \ref{alg:MP-MTL-SMC}, {where a noise vector is added in Step \ref{eq:SMC-perturb-DP} to the model vector based on sensitivity of replacing a single data instance.}

\begin{algorithm} [htb]
\caption{MP-MTL framework with Secure Multi-party Computation (SMC)}
\label{alg:MP-MTL-SMC}{\small
\begin{algorithmic}[1]
\REQUIRE{Datasets $(\X^m,\y^m) = \{(\X_1,\y_1),\ldots,(\X_m,\y_m)\}$, where $\forall i\in[m], \ \X_i\in \mathbb{R}^{n_i \times d}$ and $\y_i\in \mathbb{R}^{n_i \times 1}$. Privacy loss for model protection $\epsilon_{\mbox{mp}}, \delta_{\mbox{mp}}  \geq 0$. Privacy loss for single data instance protection $\epsilon_{\mbox{dp}}  \geq 0$. Number of iterations $T$. Shared information matrices $\M^{(0)}$. Initial models of tasks $\W^{(0)}$.}
\ENSURE{$\widehat{\W}^{(1:T)}$ .}
\STATE Set $\{\epsilon_{\mbox{mp},t}\}$ such that { $CB(\{\epsilon_{\mbox{mp},t}\},\delta_{\mbox{mp}})\leq \epsilon$, where $CB(\{\epsilon_{\mbox{mp},t}\},\delta_{\mbox{mp}})$ is the composition bound of $\{\epsilon_{\mbox{mp},t}\}$.}, taking $\epsilon_t = \epsilon_{\mbox{mp},t},\epsilon=\epsilon_{\mbox{mp}},\delta = \delta_{\mbox{mp}}$.\\
\STATE Set $\{\epsilon_{\mbox{dp},t}\}$ such that $CB(\{\epsilon_{\mbox{dp},t}\},0)\leq \epsilon_{\mbox{dp}}$, taking $\epsilon_t = \epsilon_{\mbox{dp},t},\epsilon=\epsilon_{\mbox{dp}},\delta = \delta_{\mbox{dp}}$.\\
\FOR{$t=1:T$}
{
\STATE $\tilde{\w}_i^{(t-1)} = {\w}_i^{(t-1)}/\max(1,\frac{\|{\w}_i^{(t-1)}\|_2}{K}) + \b_i$, where $\b_i$ is a sample with the density function of $$p(\b_i)\propto\exp\biggl(-\frac{2K}{\epsilon_{\mbox{dp},t}}\|\b_i\|_2\biggr),$$
 for all $i\in[m]$. \label{eq:SMC-perturb-DP}
Let $\widehat{\W}^{(0)} = \widetilde{\W}^{(0)}$.
}
\STATE  $\widetilde{\SSigma}^{(t)} = \widetilde{\W}^{(t-1)}(\widetilde{\W}^{(t-1)})\trans$ (or $\widetilde{\SSigma}^{(t)} = (\widetilde{\W}^{(t-1)})\trans\widetilde{\W}^{(t-1)}$).\\
\STATE  $\SSigma^{(t)} = \widetilde{\SSigma}^{(t)} + \E$, where $\E\sim W_d(d+1,\frac{K^2}{2\epsilon_{\mbox{mp},t}}\I_d)$ (or $\E\sim W_m(m+1,\frac{K^2}{2\epsilon_{\mbox{mp},t}}\I_m)$) is a sample of the Wishart distribution.\\
\STATE Perform an arbitrary mapping $f:\SSigma^{(1:t)}\rightarrow \M^{(t)}$. \\
\STATE $\hat{\w}_i^{(t)}= \calA_{\mbox{st},i}(\M^{(t)},  \tilde{\w}_i^{(0:t-1)},\X_i,\y_i)$ for all $i\in[m]$, where $\w_i^{(0:t-1)}$ are for the initialization. \\
\STATE Set the input for the next iteration: $\W^{(t)} = \widehat{\W}^{(t)}$.
\ENDFOR
\end{algorithmic}}
\end{algorithm}

The results in Proposition \ref{th:prop_SMC} show that we can simultaneously protect a single data instance and the model using such a divide-and-conquer strategy. Because it is not necessary to protect all the data instances in each task using data-protected algorithms, the perturbation for data-instance protection can be reduced.
\begin{proposition}\label{th:prop_SMC}
Use Property \ref{th:lem_post} and Theorem \ref{th:th_MP-MTL}. Algorithm \ref{alg:MP-MTL-SMC} is an $(\epsilon_{\mbox{mp}},\delta_{\mbox{mp}})$ - MP-MTL algorithm and an $(\epsilon_{\mbox{dp}},\delta_{\mbox{dp}})$ - IP-MTL algorithm.
\end{proposition}
\begin{proof}
For simplicity, we omit the symbol $\calB$ used to denote the input in the conditional events in some equations.

First, the \ref{eq:SMC-perturb-DP}-th step is a standard output perturbation of~\citet{chaudhuri2011differentially} with the $\ell_2$ sensitivity $\max_{\tilde{\w}_i,\tilde{\w}'_i}\|\tilde{\w}_i-\tilde{\w}'_i\|_2\leq \|\tilde{\w}_i\|_2+\|\tilde{\w}'_i\|_2\leq 2K$, where $\tilde{\w}_i$ and $\tilde{\w}'_i$ are arbitrary vectors with the maximum norm of $K$; thus, we have for all $i\in[m]$, for all neighboring datasets $\calD^m$ and $(\calD')^m$ that differ in a single data instance of the $i$-th task, and for any set $\calS\in \mathbb{R}^d$,
 \begin{align*}
 \bbP ( \tilde{\w}&_{i}^{(t-1)}  \in \calS  \mid    \tilde{\w}_i^{(0:t-2)}, \calD^m, \M^{(t-1)}   ) \\
\leq& \exp(\epsilon_{\mbox{dp},t}) \bbP( \tilde{\w}_{i}^{(t-1)}  \in \calS  \mid  \tilde{\w}_i^{(0:t-2)}, (\calD')^m, \M^{(t-1)}   ),
\end{align*}
where $\tilde{\w}_i^{(0:t-2)} = \emptyset$ when $t=1$.


Then, because the mapping $(\widetilde{\W}^{(t-1)}, \SSigma^{(1:t-1)}) \rightarrow  \theta_{t} = ( \SSigma^{(t)}, \M^{(t)}, \widetilde{\W}^{(t-1)} ) \in  \calC_{t}$ does not touch any unperturbed sensitive information of $(\X_i,\y_i,\w_i^{(0:t-1)})$, the \emph{Post-Processing immunity} property (Property \ref{th:lem_post}) can be applied such that we have for any set $\calS' \subseteq  \calC_{t}$ that
 \begin{align*}
 \bbP ( \theta&_{t}  \in \calS'  \mid  \widetilde{\W}^{(0:t-2)}, \calD^m, \M^{(t-1)}   ) \\
\leq& \exp(\epsilon_{\mbox{dp},t}) \bbP(\theta_{t} \in \calS'  \mid  \widetilde{\W}^{(0:t-2)}, (\calD')^m, \M^{(t-1)} ),
\end{align*}
which means that
 \begin{align*}
 \bbP ( \theta&_{t}  \in \calS'  \mid    \calD^m,  \ttheta_{1:t-1}  ) \\
\leq& \exp(\epsilon_{\mbox{dp},t}) \bbP(\theta_{t} \in \calS'  \mid     (\calD')^m,  \ttheta_{1:t-1}  ),
\end{align*}
where
\begin{numcases}{\ttheta_{1:t-1} = }
\nonumber \emptyset, & $t=1$\\
\nonumber \theta_1,\theta_2,\cdots,\theta_{t-1}, & $t\geq2$.
\end{numcases}

Then, by  the \emph{Adaptive composition} property (Property \ref{th:lem_Ad_comp}), we have for any set $\calS'' \subseteq  \bigotimes_{t=1}^{T}\calC_{t}$ that
\begin{align*}
 \bbP (  \theta&_{1:T}  \in \calS''  \mid  \bigcap_{t=1}^T(\calB_t =(  \calD^m,  \ttheta_{1:t-1}))  ) \\
\leq& \exp(\epsilon_{\mbox{dp}}) \bbP(  \theta_{1:T}  \in \calS''  \mid    \bigcap_{t=1}^T(\calB_t=( (\calD')^m,  \ttheta_{1:t-1} ) ))\\
&+\delta_{\mbox{dp}}.
\end{align*}

Because the mapping $(\theta_{t},\calD_{[-i]}, \tilde{\w}_{[-i]}^{(0:t-2)}, {\W}^{(t-1)}  ) \rightarrow \hat{\w}_{[-i]}^{(t)} $ does not touch any unperturbed sensitive information of $(\X_i,\y_i,\w_i^{(0:t-1)})$  for all $t\in[T]$ (${\W}^{(t-1)}$ is actually not used in the mapping), the \emph{Post-Processing immunity} property (Property \ref{th:lem_post}) can be applied such that we have for any set $\calS_0 \subseteq  \mathbb{R}^{d\times (m-1)\times T }$ that
\begin{align*}
 \bbP &( \hat{\w}_{[-i]}^{(1:T)} \in \calS_0  \mid  \bigcap_{t=1}^T(\calB_t =(  \calD^m,  \ttheta_{1:t-1}, \W^{(t-1)}))  ) \\
\leq& e^{\epsilon_{\mbox{dp},t}} \bbP( \hat{\w}_{[-i]}^{(1:T)} \in \calS_0  \mid    \bigcap_{t=1}^T(\calB_t=( (\calD')^m,  \ttheta_{1:t-1} ,(\W')^{(t-1)}) ))\\
&+\delta_{\mbox{dp}},
\end{align*}
where $(\W')^{(t-1)}$ is associated with the setting in which a single data instance of the $i$-th task has been replaced.

Therefore, Algorithm \ref{alg:MP-MTL-SMC} is an $(\epsilon_{\mbox{dp}},\delta_{\mbox{dp}})$ - IP-MTL algorithm.

Next, for the conditional density of $\SSigma^{(t)}$ given $\W^{(t-1)}$, we have
\begin{align*}
  p( \SSigma&^{(t)} \mid \W^{(t-1)}) \\
  =& \int_{\widetilde{\W}^{(t-1)}}p( \SSigma^{(t)} \mid \W^{(t-1)}, \widetilde{\W}^{(t-1)})\\
  &p(\widetilde{\W}^{(t-1)}\mid \W^{(t-1)} )d\widetilde{\W}^{(t-1)}\\
=&\int_{\widetilde{\W}^{(t-1)}}p( \SSigma^{(t)} \mid  \widetilde{\W}^{(t-1)}) p(\widetilde{\W}^{(t-1)}\mid \W^{(t-1)} )d\widetilde{\W}^{(t-1)}\\
=&\int_{\widetilde{\W}^{(t-1)}}p( \SSigma^{(t)} \mid  \widetilde{\W}^{(t-1)}) \prod_{i=1}^{m}p(\tilde{\w}_i^{(t-1)}\mid \w_i^{(t-1)} )d\widetilde{\W}^{(t-1)}.\\
\end{align*}

Because, for all $i\in[m]$ and some constant $c=\frac{\tilde{s}_i^{(t-1)}}{\epsilon_{\mbox{dp},t}}$, we have
\begin{align*}
p(\tilde{\w}_i^{(t-1)}\mid \w_i^{(t-1)} ) \propto \exp\biggl(-c\|\tilde{\w}_i^{(t-1)} - \w_i^{(t-1)} \|_2\biggr),
\end{align*}
given $(\W')^{(t-1)}$ such that for some $i\in[m]$, $(\w'_i)^{(t-1)}\neq \w_i^{(t-1)}$,
letting $(\tilde{\w}'_i)^{(t-1)} = \tilde{\w}_i^{(t-1)} - \w_i^{(t-1)} + (\w'_i)^{(t-1)}$, we have
\begin{align*}
\|(\tilde{\w}'_i)^{(t-1)} -  (\w'_i)^{(t-1)}\|_2 &= \|\tilde{\w}_i^{(t-1)} - \w_i^{(t-1)}\|_2  \\
\Rightarrow p((\tilde{\w}'_i)^{(t-1)}\mid \w_i^{(t-1)} )  &= p(\tilde{\w}_i^{(t-1)}\mid \w_i^{(t-1)} ),
\end{align*}
and $d(\tilde{\w}'_i)^{(t-1)}  = d\tilde{\w}_i^{(t-1)} $.

Furthermore, based on the proof of Theorem \ref{th:th_MP-MTL} in Section \ref{sec:proof_th_MP-MTL}, we know that for neighboring matrices $\widetilde{\W}^{(t-1)}$ and $(\widetilde{\W}')^{(t-1)}$ that differ in the $i$-th column, we have
\begin{align*}
p( \SSigma^{(t)} \mid  \widetilde{\W}^{(t-1)}) \leq \exp(\epsilon_{\mbox{mp},t}) p( \SSigma^{(t)} \mid  (\widetilde{\W}')^{(t-1)}).
\end{align*}

Therefore, for all $i\in[m]$, given $(\W')^{(t-1)}$ such that $(\w'_i)^{(t-1)}\neq \w_i^{(t-1)}$, under the choice for $(\tilde{\w}'_i)^{(t-1)}$, we have
\begin{align*}
  p(& \SSigma^{(t)} \mid \W^{(t-1)}) \\
=&\int_{\widetilde{\W}^{(t-1)}}p( \SSigma^{(t)} \mid  \widetilde{\W}^{(t-1)}) \prod_{j=1}^{m}p(\tilde{\w}_j^{(t-1)}\mid \w_j^{(t-1)} )d\widetilde{\W}^{(t-1)}\\
\leq&\int_{(\widetilde{\W}')^{(t-1)}}e^{\epsilon_{\mbox{mp},t}} p( \SSigma^{(t)} \mid  (\widetilde{\W}')^{(t-1)}) p((\tilde{\w}'_i)^{(t-1)}\mid (\w'_i)^{(t-1)} )\\
&\prod_{j\in[m],j\neq i}p(\tilde{\w}_j^{(t-1)}\mid \w_j^{(t-1)} )d(\widetilde{\W}')^{(t-1)}.\\
=&\int_{(\widetilde{\W}')^{(t-1)}}\exp(\epsilon_{\mbox{mp},t})p( \SSigma^{(t)} \mid  (\widetilde{\W}')^{(t-1)})\\
& p((\widetilde{\W}')^{(t-1)}\mid (\W')^{(t-1)} )d(\widetilde{\W}')^{(t-1)}\\
=&\exp(\epsilon_{\mbox{mp},t}) p( \SSigma^{(t)} \mid (\W')^{(t-1)}),
\end{align*}
which renders the mapping $\W^{(t-1)} \rightarrow \SSigma^{(t)}$ as an $(\exp(\epsilon_{\mbox{mp},t}), 0)$ - differentially private algorithm.

Then, according to the proof of Theorem \ref{th:th_MP-MTL} in Section \ref{sec:proof_th_MP-MTL}, Algorithm \ref{alg:MP-MTL-SMC} is an $(\epsilon_{\mbox{mp}},\delta_{\mbox{mp}})$ - MP-MTL algorithm.
\end{proof}

{
\section{Results of Utility Analyses Under Other Two Settings}\label{sec:result_utility_other_two}

Here we consider the other two settings of $\{\epsilon_t\}$.
\subsection{Setting No.1}\label{sec:result_utility_other_two_setting_1}
In this setting, we have
  \begin{align*}
 \epsilon =\sum_{t=1}^{T}\epsilon_t.
 \end{align*}

First, we consider heterogeneous privacy budgets and set $\epsilon_t =\Theta(t^{\alpha})$ for $\alpha\in \mathbb{R}$ and $t\in[T]$ for the convex case. In this case, denote by
\begin{align*}
   M_0= 1/{ |\alpha+1|\epsilon}.
\end{align*}

\begin{theorem}[Low rank - Convexity - Setting No.1]\label{th:th_bound_LR_main_convex_setting_1}
Consider Algorithm \ref{alg:MP-MTL-LR}. For an index $k\leq q$ that suffices the definition in Lemma \ref{th:lem_LR_iter_t} for all $t\in[T]$, $\eta =1/L$, $\lambda = \Theta(LK\sqrt{m} )$, assume $\epsilon_t\leq 4Kk^2d(\log d)/q^2$ for $t\in[T]$. Denote by
\begin{align*}
M = M_0Kkd\log d/\sqrt{m}.
\end{align*}
\noindent\textbf{No acceleration}: If we set $\beta_t = 0$ for $t\in [m]$, then setting $T=\Theta(((\alpha/2 - 1)^2/M)^{\phi(\alpha)})$
for $\mathcal{E} = f(\frac{1}{T}\sum_{t=1}^{T}\widehat{\W}^{(t)}) - f(\W_*)$, we have with high probability,
\begin{equation}\label{eq:bound_LR_main_basic_convex_setting_1}
\mathcal{E}= O(K^2L(M/(\alpha/2 - 1)^2)^{\phi(\alpha)}),
\end{equation}
where
\begin{equation}\label{eq:bound_LR_main_basic_convex_phi_setting_1}
\begin{split}
\phi(\alpha) =
\left\{
  \begin{array}{ll}
    1/(\alpha+1), &  {\alpha>2;} \\
    1/3, &  {-1<\alpha<2;} \\
    1/(2-\alpha), &  {\alpha<-1.}
  \end{array}
\right.
\end{split}
\end{equation}

\noindent\textbf{Use acceleration}: If we set $\beta_t = {(t-1)}/{(t+2)}$ for $t\in [m]$, then setting $T=\Theta(((\alpha/2 - 2)^2/M)^{\phi(\alpha)/2})$
for $\mathcal{E} = f(\widehat{\W}^{(T)}) - f(\W_*)$, we have with high probability,
\begin{equation}\label{eq:bound_LR_main_acc_convex_setting_1}
\mathcal{E}= O(K^2L(M/(\alpha/2 - 2)^2)^{\phi(\alpha)}),
\end{equation}
where
\begin{equation}\label{eq:bound_LR_main_acc_convex_phi_setting_1}
\begin{split}
\phi(\alpha) =
\left\{
  \begin{array}{ll}
    2/(\alpha+1), &  {\alpha>4;} \\
    2/5, &  {-1<\alpha<4;} \\
    2/(4-\alpha), &  {\alpha<-1.}
  \end{array}
\right.
\end{split}
\end{equation}
\end{theorem}
\begin{proof}
First, consider the case with no acceleration. We first use Proposition 1 of \citet{schmidt2011convergence} by regarding procedures from Step \ref{eq:prox_start_LR} to Step \ref{eq:feature_tran-LR} as approximation for the proximal operator in \eqref{eq:proximal_op_LR}. Note that the norm clipping only bounds the parameter space and does not affect the results of \citet{schmidt2011convergence}. Then for $\varepsilon_t$ defined in Lemma \ref{th:lem_bound_vareps_t_LR} for $t\in[T]$, we have
\begin{align*}
\mathcal{E} =& \frac{2L}{m(T+1)^2}\biggl(\|\widetilde{\W}^{(0)}-\W_*\|_F \\
&+ 2\sum_{t=1}^{T}t\sqrt{\frac{2\varepsilon_t}{L}}+\sqrt{2\sum_{t=1}^{T}t^2\frac{\varepsilon_t}{L}}\biggr)^2.
\end{align*}

Meanwhile, by Lemma \ref{th:lem_bound_vareps_t_LR}, we have
\begin{align*}
\varepsilon_t = O\biggl( \frac{\kappa}{\epsilon_t}\biggr),
\end{align*}
where $\kappa = \frac{K^3\sqrt{m}kd\log d}{\eta}$.

On the other hand, let
\begin{align*}
c_1 = \epsilon = \sum_{t=1}^{T}\epsilon_t,
\end{align*}
then by Lemma \ref{th:lem_bound_error_poly}, we have
\begin{align*}
\sum_{t=1}^{T}\sqrt{\varepsilon_t} =
\left\{
  \begin{array}{ll}
    O\biggl( \sqrt{\frac{\kappa T^{\alpha+1} }{c_1(\alpha/2 - 1)^2(\alpha+1)  }}\biggr), &  {\alpha>2;} \\
    O\biggl( \sqrt{\frac{\kappa T^{3} }{c_1(\alpha/2 - 1)^2(\alpha+1)  }}\biggr), &  {-1<\alpha<2;} \\
    O\biggl( \sqrt{\frac{\kappa T^{2-\alpha} }{c_1(\alpha/2 - 1)^2(-\alpha-1)  }}\biggr), &  {\alpha<-1,}
  \end{array}
\right.
\end{align*}

Because $\widetilde{\W}^{(0)}$ is the result of the norm clipping, we have $\widetilde{\W}^{(0)} \in \mathcal{W}$.

Finally, taking $c_3 = \phi(\alpha)$ defined in \eqref{eq:bound_LR_main_basic_convex_phi} and $c_4 = \frac{\kappa}{c_2(\alpha/2 - 1)^2|\alpha+1| }$, under the assumption that $\W_*\in \mathcal{W}$, using Lemma \ref{th:lem_pseudo_bound_basic_convex}, we have the results for the case with no acceleration.

For the accelerated case, we use Proposition 2 of \citet{schmidt2011convergence} to have
\begin{align*}
\mathcal{E} =& \frac{2L}{m(T+1)^2}\biggl(\|\widetilde{\W}^{(0)}-\W_*\|_F \\
&+ 2\sum_{t=1}^{T}t\sqrt{\frac{2\varepsilon_t}{L}}+\sqrt{2\sum_{t=1}^{T}t^2\frac{\varepsilon_t}{L}}\biggr)^2.
\end{align*}
Then one can prove similarly combining Lemma \ref{th:lem_bound_vareps_t_LR}, Lemma \ref{th:lem_bound_sum_eps_case_2_and_3}, Lemma \ref{th:lem_bound_error_poly} and Lemma \ref{th:lem_pseudo_bound_acc_convex}.

\end{proof}

\begin{theorem}[Group sparse - Convexity - Setting No.1]\label{th:th_bound_GS_main_convex_setting_1}
Consider Algorithm \ref{alg:MP-MTL-GS}. For an index $k\leq d$ that suffices the definition in Lemma \ref{th:lem_GS_iter_t} for all $t\in[T]$, $\eta =1/L$, $\lambda = \Theta(LKd\sqrt{m} )$, assume $\epsilon_t\leq {k^2\log(d)}/{4Kd(d-k)^2m}$ for $t\in[T]$. Denote by
\begin{align*}
M = M_0k\log d/m.
\end{align*}
\noindent\textbf{No acceleration}: If we set $\beta_t = 0$ for $t\in [m]$, then setting $T=\Theta(((\alpha/2 - 1)^2/M)^{\phi(\alpha)})$
for $\mathcal{E} = f(\frac{1}{T}\sum_{t=1}^{T}\widehat{\W}^{(t)}) - f(\W_*)$, we have with high probability,
\begin{equation}\label{eq:bound_GS_main_basic_convex_setting_1}
\mathcal{E}= O(K^2L(M/(\alpha/2 - 1)^2)^{\phi(\alpha)}),
\end{equation}
where $\phi(\alpha)$ is defined in \eqref{eq:bound_LR_main_basic_convex_phi_setting_1}.

\noindent\textbf{Use acceleration}: If we set $\beta_t = {(t-1)}/{(t+2)}$ for $t\in [m]$, then setting $T=\Theta(((\alpha/2 - 2)^2/M)^{\phi(\alpha)/2})$
for $\mathcal{E} = f(\widehat{\W}^{(T)}) - f(\W_*)$, we have with high probability,
\begin{equation}\label{eq:bound_GS_main_acc_convex_setting_1}
\mathcal{E}= O(K^2L(M/(\alpha/2 - 2)^2)^{\phi(\alpha)}),
\end{equation}
where $\phi(\alpha)$ is defined in \eqref{eq:bound_LR_main_acc_convex_phi_setting_1}.
\end{theorem}
\begin{proof}
First, consider the case with no acceleration. We use Proposition 1 of \citet{schmidt2011convergence} and prove similarly as in the proof for Theorem \ref{th:th_bound_LR_main_convex_setting_1}, combining Lemma \ref{th:lem_bound_vareps_t_GS}, Lemma \ref{th:lem_bound_sum_eps_case_2_and_3}, Lemma \ref{th:lem_bound_error_poly} and Lemma \ref{th:lem_pseudo_bound_basic_convex}.

For the accelerated case, we use Proposition 2 of \citet{schmidt2011convergence} and prove similarly as in the proof for Theorem \ref{th:th_bound_LR_main_convex_setting_1},  combining Lemma \ref{th:lem_bound_vareps_t_GS}, Lemma \ref{th:lem_bound_sum_eps_case_2_and_3}, Lemma \ref{th:lem_bound_error_poly} and Lemma \ref{th:lem_pseudo_bound_acc_convex}.
\end{proof}

Now we further assume that $mf(\W)$ is $\mu$-strongly convex and has $L$-Lipschitz-continuous gradient, where $\mu<L$.
We set $\epsilon_t =\Theta(Q^{-t})$ for $Q>0$ and $t\in[T]$ for this case. In this case, denote by
\begin{align*}
   M'_0= 1/{ |1-Q|\epsilon}.
\end{align*}

\begin{theorem}[Low rank - Strong convexity - Setting No.1]\label{th:th_bound_LR_main_strong_convex_setting_1}
Consider Algorithm \ref{alg:MP-MTL-LR}. For an index $k\leq q$ that suffices the definition in Lemma \ref{th:lem_LR_iter_t} for all $t\in[T]$, $\eta =1/L$, $\lambda = \Theta(LK\sqrt{m} )$, assume $\epsilon_t\leq 4Kk^2d(\log d)/q^2$ for $t\in[T]$. Denote by
\begin{align*}
   M = M'_0Kkd\log d/\sqrt{m}.
\end{align*}
\noindent\textbf{No acceleration}: If we set $\beta_t = 0$ for $t\in [m]$, then denoting $Q_0 = 1-\mu/L$ and setting $T = \Theta(\log_{1/\psi(Q,Q_0^2)}((Q_0/\sqrt{Q} - 1)^2/M))$
for $\mathcal{E} = \frac{1}{\sqrt{m}}\|\widehat{\W}^{(T)}-\W_*\|_F$, we have with high probability,
\begin{equation}\label{eq:bound_LR_main_basic_strong_convex_setting_1}
\mathcal{E}= O(K(M/(Q_0/\sqrt{Q} - 1)^2)^{\log_{\psi(Q,Q_0^2)}Q_0}),
\end{equation}
where $\psi(\cdot,\cdot)$ is defined in \eqref{eq:bound_LR_main_basic_strong_convex_psi}.

\noindent\textbf{Use acceleration}: If we set $\beta_t = {(1-\sqrt{\mu/L})}/{(1+\sqrt{\mu/L})}$ for $t\in [m]$, then denoting $Q'_0 = 1-\sqrt{\mu/L}$ and setting $T = \Theta(\log_{1/\psi(Q,Q'_0)}((\sqrt{Q'_0}/\sqrt{Q} - 1)^2/M))$
for $\mathcal{E} = f(\widehat{\W}^{(T)}) - f(\W_*)$, we have with high probability,
\begin{equation}\label{eq:bound_LR_main_acc_strong_convex_setting_1}
\mathcal{E}= O(K(M/(\sqrt{Q'_0}/\sqrt{Q} - 1)^2)^{\log_{\psi(Q,Q'_0)}Q'_0}),
\end{equation}
where $\psi(\cdot,\cdot)$ is defined in \eqref{eq:bound_LR_main_basic_strong_convex_psi}.
\end{theorem}
\begin{proof}
First, consider the case with no acceleration. We use Proposition 3 of \citet{schmidt2011convergence} to have
\begin{align*}
\mathcal{E}& =\frac{Q_0^T}{\sqrt{m}}\biggl(\|\widetilde{\W}^{(0)}-\W_*\|_F + 2\sum_{t=1}^{T}Q_0^{-t}\sqrt{\frac{2\varepsilon_t}{L}}\biggr).
\end{align*}
Then one can prove similarly as in the proof for Theorem \ref{th:th_bound_LR_main_convex_setting_1}, combining Lemma \ref{th:lem_bound_vareps_t_LR}, Lemma \ref{th:lem_bound_sum_eps_case_2_and_3}, Lemma \ref{th:lem_bound_error_poly} and Lemma \ref{th:lem_pseudo_bound_basic_strong_convex}.

For the accelerated case, we use Proposition 4 of \citet{schmidt2011convergence} to have
\begin{align*}
\mathcal{E} =&\frac{(Q_0)^T}{m}\biggl(\sqrt{2(f(\widehat{\W}^{(0)}) - f(\W_*))} + 2\sqrt{\frac{L}{\mu}}\sum_{t=1}^{T}\sqrt{\varepsilon_t(Q_0)^{-t}} \\
&+ \sqrt{\sum_{t=1}^{T}\varepsilon_t(Q_0)^{-t}}\biggr)^2 .
\end{align*}
Then one can prove similarly as in the proof for Theorem \ref{th:th_bound_LR_main_convex_setting_1}, using the assumption that $f(\widetilde{\W}^{(0)})-f(\W_*) = O(K^2Lm)$,  combining Lemma \ref{th:lem_bound_vareps_t_LR}, Lemma \ref{th:lem_bound_sum_eps_case_2_and_3}, Lemma \ref{th:lem_bound_error_poly} and Lemma \ref{th:lem_pseudo_bound_acc_strong_convex}.
\end{proof}

\begin{theorem}[Group sparse - Strong convexity - Setting No.1]\label{th:th_bound_GS_main_strong_convex_setting_1}
Consider Algorithm \ref{alg:MP-MTL-GS}. For an index $k\leq d $that suffices the definition in Lemma \ref{th:lem_GS_iter_t} for all $t\in[T]$, $\eta =1/L$, $\lambda = \Theta(LKd\sqrt{m} )$, assume $\epsilon_t\leq {k^2\log(d)}/{4Kd(d-k)^2m}$ for $t\in[T]$. Denote by
\begin{align*}
M = M'_0k\log d/{m}.
\end{align*}
\noindent\textbf{No acceleration}: If we set $\beta_t = 0$ for $t\in [m]$, then denoting $Q_0 = 1-\mu/L$ and setting $T = \Theta(\log_{1/\psi(Q,Q_0^2)}((Q_0/\sqrt{Q} - 1)^2/M))$
for $\mathcal{E} = \frac{1}{\sqrt{m}}\|\widehat{\W}^{(T)}-\W_*\|_F$, we have with high probability,
\begin{equation}\label{eq:bound_GS_main_basic_strong_convex_setting_1}
\mathcal{E}= O(K(M/(Q_0/\sqrt{Q} - 1)^2)^{\log_{\psi(Q,Q_0^2)}Q_0} ),
\end{equation}
where $\psi(\cdot,\cdot)$ is defined in \eqref{eq:bound_LR_main_basic_strong_convex_psi}.

\noindent\textbf{Use acceleration}: If we set $\beta_t = {(1-\sqrt{\mu/L})}/{(1+\sqrt{\mu/L})}$ for $t\in [m]$, then denoting $Q'_0 = 1-\sqrt{\mu/L}$ and setting $T = \Theta(\log_{1/\psi(Q,Q'_0)}((\sqrt{Q'_0}/\sqrt{Q} - 1)^2/M))$
for $\mathcal{E} = f(\widehat{\W}^{(T)}) - f(\W_*)$, we have with high probability,
\begin{equation}\label{eq:bound_GS_main_acc_strong_convex_setting_1}
\mathcal{E}= O(K(M/(\sqrt{Q'_0}/\sqrt{Q} - 1)^2)^{\log_{\psi(Q,Q'_0)}Q'_0}),
\end{equation}
where $\psi(\cdot,\cdot)$ is defined in \eqref{eq:bound_LR_main_basic_strong_convex_psi}.
\end{theorem}
\begin{proof}
First, consider the case with no acceleration. We use Proposition 3 of \citet{schmidt2011convergence} and prove similarly as in the proof for Theorem \ref{th:th_bound_LR_main_convex_setting_1}, combining Lemma \ref{th:lem_bound_vareps_t_GS}, Lemma \ref{th:lem_bound_sum_eps_case_2_and_3}, Lemma \ref{th:lem_bound_error_poly} and Lemma \ref{th:lem_pseudo_bound_basic_strong_convex}.

For the accelerated case, we use Proposition 4 of \citet{schmidt2011convergence} and prove similarly as in the proof for Theorem \ref{th:th_bound_LR_main_convex_setting_1}, using the assumption that $f(\widetilde{\W}^{(0)})-f(\W_*) = O(K^2Lm)$,  combining Lemma \ref{th:lem_bound_vareps_t_GS}, Lemma \ref{th:lem_bound_sum_eps_case_2_and_3}, Lemma \ref{th:lem_bound_error_poly} and Lemma \ref{th:lem_pseudo_bound_acc_strong_convex}.
\end{proof}

Then we optimize the utility bounds with respect to the respective budget allocation strategies.
\begin{theorem}[Budget allocation - Setting No.1]\label{th:th_budget_setting_1}
Consider Algorithm \ref{alg:MP-MTL-LR} and  Algorithm \ref{alg:MP-MTL-GS}.

For \emph{convex} $f$, use Theorem \ref{th:th_bound_LR_main_convex_setting_1} and Theorem \ref{th:th_bound_GS_main_convex_setting_1}.

(1) \emph{No acceleration}: Both the bounds for low-rank and group-sparse estimators achieve their respective minimums w.r.t. $\alpha$ at $\alpha = 0$. Meanwhile, $\phi(\alpha) = 1/3$.

(2) \emph{Accelerated}: Both the bounds for low-rank and group-sparse estimators achieve their respective minimums w.r.t. $\alpha$ at $\alpha = 2/3$. Meanwhile, $\phi(\alpha) = 2/5$.

For \emph{strongly convex} $f$, use Theorem \ref{th:th_bound_LR_main_strong_convex_setting_1} and Theorem \ref{th:th_bound_GS_main_strong_convex_setting_1}.

(1) \emph{No acceleration}: Both the bounds for low-rank and group-sparse estimators achieve their respective minimums w.r.t. $Q$ at $Q = Q_0^{2/3}$. Meanwhile, $\log_{\psi(Q,Q_0^2)}Q_0 = 1/2$.

(2) \emph{Accelerated}: Both the bounds for low-rank and group-sparse estimators achieve their respective minimums w.r.t. $Q$ at $Q = (Q'_0)^{1/3}$. Meanwhile, $\log_{\psi(Q,Q'_0)}Q'_0 = 1$.
\end{theorem}
\begin{proof}
Consider the bound in \eqref{eq:bound_LR_main_basic_convex_setting_1}. First, by Assumption \ref{th:asm_m}, $\mathcal{E}$ is minimized by maximizing $\phi(\alpha)$ and $(\alpha/2 - 1)^2|\alpha+1|$, which are maximized simultaneously when $\alpha = 0$. Results under other settings can be proved similarly.
\end{proof}

\subsection{Setting No.2}\label{sec:result_utility_other_two_setting_2}
In this setting, we have
\begin{align*}
 \epsilon =\sum_{t=1}^{T}\frac{(e^{\epsilon_t}-1)\epsilon_t}{(e^{\epsilon_t}+1)}
+\sqrt{\sum_{t=1}^{T}2\epsilon_t^2\log\biggl(\frac{1}{\delta}\biggr)}.
 \end{align*}

First, we consider heterogeneous privacy budgets and set $\epsilon_t =\Theta(t^{\alpha})$ for $\alpha\in \mathbb{R}$ and $t\in[T]$ for the convex case. In this case, denote by
\begin{align*}
   M_0= \left.\sqrt{\log(1/\delta)+2\epsilon}\middle/{ \sqrt{|2\alpha+1|}\epsilon}\right..
\end{align*}

\begin{theorem}[Low rank - Convexity - Setting No.2]\label{th:th_bound_LR_main_convex_setting_2}
Consider Algorithm \ref{alg:MP-MTL-LR}. For an index $k\leq q$ that suffices the definition in Lemma \ref{th:lem_LR_iter_t} for all $t\in[T]$, $\eta =1/L$, $\lambda = \Theta(LK\sqrt{m} )$, assume $\epsilon_t\leq 4Kk^2d(\log d)/q^2$ for $t\in[T]$. Denote by
\begin{align*}
   M=M_0Kkd\log d/\sqrt{m}.
\end{align*}
\noindent\textbf{No acceleration}: If we set $\beta_t = 0$ for $t\in [m]$, then setting $T = \Theta(((\alpha/2 - 1)^2/M)^{\phi(\alpha)} )$
for $\mathcal{E} = f(\frac{1}{T}\sum_{t=1}^{T}\widehat{\W}^{(t)}) - f(\W_*)$, we have with high probability,
\begin{equation}\label{eq:bound_LR_main_basic_convex_setting_2}
\mathcal{E}= O(K^2L(M/(\alpha/2 - 1)^2)^{\phi(\alpha)}),
\end{equation}
where
\begin{equation}\label{eq:bound_LR_main_basic_convex_phi_setting_2}
\begin{split}
\phi(\alpha) =
\left\{
  \begin{array}{ll}
    2/(2\alpha+1), &  {\alpha>2;} \\
    2/5, &  {-1/2<\alpha<2;} \\
    1/(2-\alpha), &  {\alpha<-1/2.}
  \end{array}
\right.
\end{split}
\end{equation}

\noindent\textbf{Use acceleration}: If we set $\beta_t = {(t-1)}/{(t+2)}$ for $t\in [m]$, then setting $T = \Theta(((\alpha/2 - 2)^2/M)^{\phi(\alpha)/2} )$
for $\mathcal{E} = f(\widehat{\W}^{(T)}) - f(\W_*)$, we have with high probability,
\begin{equation}\label{eq:bound_LR_main_acc_convex_setting_2}
\mathcal{E}= O(K^2L(M/(\alpha/2 - 2)^2)^{\phi(\alpha)}),
\end{equation}
where
\begin{equation}\label{eq:bound_LR_main_acc_convex_phi_setting_2}
\begin{split}
\phi(\alpha) =
\left\{
  \begin{array}{ll}
    4/(2\alpha+1), &  {\alpha>4;} \\
    4/9, &  {-1/2<\alpha<4;} \\
    2/(4-\alpha), &  {\alpha<-1/2.}
  \end{array}
\right.
\end{split}
\end{equation}
\end{theorem}
\begin{proof}
It is the corollary of Theorem \ref{th:th_bound_LR_main_convex}, replacing the term $\sqrt{\log(e+\epsilon/\delta)}$ with the term $\sqrt{\log(1/\delta)}$ by Lemma \ref{th:lem_bound_sum_eps_case_2_and_3}.
\end{proof}

\begin{theorem}[Group sparse - Convexity - Setting No.2]\label{th:th_bound_GS_main_convex_setting_2}
Consider Algorithm \ref{alg:MP-MTL-GS}. For an index $k\leq d$ that suffices the definition in Lemma \ref{th:lem_GS_iter_t} for all $t\in[T]$, $\eta =1/L$, $\lambda = \Theta(LKd\sqrt{m} )$, assume $\epsilon_t\leq {k^2\log(d)}/{4Kd(d-k)^2m}$ for $t\in[T]$. Denote by
\begin{align*}
   M=M_0k\log d/{m}.
\end{align*}
\noindent\textbf{No acceleration}: If we set $\beta_t = 0$ for $t\in [m]$, then setting $T = \Theta(((\alpha/2 - 1)^2/M)^{\phi(\alpha)} )$
for $\mathcal{E} = f(\frac{1}{T}\sum_{t=1}^{T}\widehat{\W}^{(t)}) - f(\W_*)$, we have with high probability,
\begin{equation}\label{eq:bound_GS_main_basic_convex_setting_2}
\mathcal{E}= O(K^2L(M/(\alpha/2 - 1)^2)^{\phi(\alpha)}),
\end{equation}
where $\phi(\alpha)$ is defined in \eqref{eq:bound_LR_main_basic_convex_phi_setting_2}.

\noindent\textbf{Use acceleration}: If we set $\beta_t = {(t-1)}/{(t+2)}$ for $t\in [m]$, then setting $T = \Theta(((\alpha/2 - 2)^2/M)^{\phi(\alpha)/2} )$
for $\mathcal{E} = f(\widehat{\W}^{(T)}) - f(\W_*)$, we have with high probability,
\begin{equation}\label{eq:bound_GS_main_acc_convex_setting_2}
\mathcal{E}= O(K^2L(M/(\alpha/2 - 2)^2)^{\phi(\alpha)} ),
\end{equation}
where $\phi(\alpha)$ is defined in \eqref{eq:bound_LR_main_acc_convex_phi_setting_2}.
\end{theorem}
\begin{proof}
It is the corollary of Theorem \ref{th:th_bound_GS_main_convex}, replacing the term $\sqrt{\log(e+\epsilon/\delta)}$ with the term $\sqrt{\log(1/\delta)}$ by Lemma \ref{th:lem_bound_sum_eps_case_2_and_3}.
\end{proof}

Now we further assume that $mf(\W)$ is $\mu$-strongly convex and has $L$-Lipschitz-continuous gradient, where $\mu<L$.
We set $\epsilon_t =\Theta(Q^{-t})$ for $Q>0$ and $t\in[T]$ for this case. In this case, denote by
\begin{align*}
   M'_0= \left.\sqrt{\log(1/\delta)+2\epsilon}\middle/{ \sqrt{|1-Q^2|}\epsilon}\right..
\end{align*}

\begin{theorem}[Low rank - Strong convexity - Setting No.2]\label{th:th_bound_LR_main_strong_convex_setting_2}
Consider Algorithm \ref{alg:MP-MTL-LR}. For an index $k\leq q$ that suffices the definition in Lemma \ref{th:lem_LR_iter_t} for all $t\in[T]$, $\eta =1/L$, $\lambda = \Theta(LK\sqrt{m} )$, assume $\epsilon_t\leq 4Kk^2d(\log d)/q^2$ for $t\in[T]$. Denote by
\begin{align*}
   M = M'_0Kkd\log d/\sqrt{m}.
\end{align*}
\noindent\textbf{No acceleration}: If we set $\beta_t = 0$ for $t\in [m]$, then denoting $Q_0 = 1-\mu/L$ and setting $T = \Theta(\log_{1/\psi(Q,Q_0^2)}((Q_0/\sqrt{Q} - 1)^2/M))$
for $\mathcal{E} = \frac{1}{\sqrt{m}}\|\widehat{\W}^{(T)}-\W_*\|_F$, we have with high probability,
\begin{equation}\label{eq:bound_LR_main_basic_strong_convex_setting_2}
\mathcal{E}= O(K(M/(Q_0/\sqrt{Q} - 1)^2)^{\log_{\psi(Q,Q_0^2)}Q_0}),
\end{equation}
where $\psi(\cdot,\cdot)$ is defined in \eqref{eq:bound_LR_main_basic_strong_convex_psi}.

\noindent\textbf{Use acceleration}: If we set $\beta_t = {(1-\sqrt{\mu/L})}/{(1+\sqrt{\mu/L})}$ for $t\in [m]$, then denoting $Q'_0 = 1-\sqrt{\mu/L}$ and setting $T = \Theta(\log_{1/\psi(Q,Q'_0)}((\sqrt{Q'_0}/\sqrt{Q} - 1)^2/M))$
for $\mathcal{E} = f(\widehat{\W}^{(T)}) - f(\W_*)$, we have with high probability,
\begin{equation}\label{eq:bound_LR_main_acc_strong_convex_setting_2}
\mathcal{E}= O(K(M/(\sqrt{Q'_0}/\sqrt{Q} - 1)^2)^{\log_{\psi(Q,Q'_0)}Q'_0}),
\end{equation}
where $\psi(\cdot,\cdot)$ is defined in \eqref{eq:bound_LR_main_basic_strong_convex_psi}.
\end{theorem}
\begin{proof}
It is the corollary of Theorem \ref{th:th_bound_LR_main_strong_convex}, replacing the term $\sqrt{\log(e+\epsilon/\delta)}$ with the term $\sqrt{\log(1/\delta)}$ by Lemma \ref{th:lem_bound_sum_eps_case_2_and_3}.
\end{proof}

\begin{theorem}[Group sparse - Strong convexity - Setting No.2]\label{th:th_bound_GS_main_strong_convex_setting_2}
Consider Algorithm \ref{alg:MP-MTL-GS}. For an index $k\leq d $that suffices the definition in Lemma \ref{th:lem_GS_iter_t} for all $t\in[T]$, $\eta =1/L$, $\lambda = \Theta(LKd\sqrt{m} )$, assume $\epsilon_t\leq {k^2\log(d)}/{4Kd(d-k)^2m}$ for $t\in[T]$. Denote by
\begin{align*}
   M = M'_0k\log d/{m}.
\end{align*}
\noindent\textbf{No acceleration}: If we set $\beta_t = 0$ for $t\in [m]$, then denoting $Q_0 = 1-\mu/L$ and setting $T = \Theta(\log_{1/\psi(Q,Q_0^2)}((Q_0/\sqrt{Q} - 1)^2/M))$
for $\mathcal{E} = \frac{1}{\sqrt{m}}\|\widehat{\W}^{(T)}-\W_*\|_F$, we have with high probability,
\begin{equation}\label{eq:bound_GS_main_basic_strong_convex_setting_2}
\mathcal{E}= O(K(M/(Q_0/\sqrt{Q} - 1)^2)^{\log_{\psi(Q,Q_0^2)}Q_0} ),
\end{equation}
where $\psi(\cdot,\cdot)$ is defined in \eqref{eq:bound_LR_main_basic_strong_convex_psi}.

\noindent\textbf{Use acceleration}: If we set $\beta_t = {(1-\sqrt{\mu/L})}/{(1+\sqrt{\mu/L})}$ for $t\in [m]$, then denoting $Q'_0 = 1-\sqrt{\mu/L}$ and setting
$T = \Theta(\log_{1/\psi(Q,Q'_0)}((\sqrt{Q'_0}/\sqrt{Q} - 1)^2/M))$
for $\mathcal{E} = f(\widehat{\W}^{(T)}) - f(\W_*)$, we have with high probability,
\begin{equation}\label{eq:bound_GS_main_acc_strong_convex_setting_2}
\mathcal{E}= O(K(M/(\sqrt{Q'_0}/\sqrt{Q} - 1)^2)^{\log_{\psi(Q,Q'_0)}Q'_0}),
\end{equation}
where $\psi(\cdot,\cdot)$ is defined in \eqref{eq:bound_LR_main_basic_strong_convex_psi}.
\end{theorem}
\begin{proof}
It is the corollary of Theorem \ref{th:th_bound_GS_main_strong_convex}, replacing the term $\sqrt{\log(e+\epsilon/\delta)}$ with the term $\sqrt{\log(1/\delta)}$ by Lemma \ref{th:lem_bound_sum_eps_case_2_and_3}.
\end{proof}

Then we optimize the utility bounds with respect to the respective budget allocation strategies.
\begin{theorem}[Budget allocation - Setting No.2]\label{th:th_budget_setting_2}
Consider Algorithm \ref{alg:MP-MTL-LR} and  Algorithm \ref{alg:MP-MTL-GS}.

For \emph{convex} $f$, use Theorem \ref{th:th_bound_LR_main_convex_setting_2} and Theorem \ref{th:th_bound_GS_main_convex_setting_2}.

(1) \emph{No acceleration}: Both the bounds for low-rank and group-sparse estimators achieve their respective minimums w.r.t. $\alpha$ at $\alpha = 0$. Meanwhile, $\phi(\alpha) = 2/5$.

(2) \emph{Accelerated}: Both the bounds for low-rank and group-sparse estimators achieve their respective minimums w.r.t. $\alpha$ at $\alpha = 2/5$. Meanwhile, $\phi(\alpha) = 4/9$.

For \emph{strongly convex} $f$, use Theorem \ref{th:th_bound_LR_main_strong_convex_setting_2} and Theorem \ref{th:th_bound_GS_main_strong_convex_setting_2}.

(1) \emph{No acceleration}: Both the bounds for low-rank and group-sparse estimators achieve their respective minimums w.r.t. $Q$ at $Q = Q_0^{2/5}$. Meanwhile, $\log_{\psi(Q,Q_0^2)}Q_0 = 1/2$.

(2) \emph{Accelerated}: Both the bounds for low-rank and group-sparse estimators achieve their respective minimums w.r.t. $Q$ at $Q = (Q'_0)^{1/5}$. Meanwhile, $\log_{\psi(Q,Q'_0)}Q'_0 = 1$.
\end{theorem}
\begin{proof}
It is the corollaries of Theorem \ref{th:th_budget}, replacing the term $\sqrt{\log(e+\epsilon/\delta)}$ with the term $\sqrt{\log(1/\delta)}$ by Lemma \ref{th:lem_bound_sum_eps_case_2_and_3}.
\end{proof}

}

{
\section{Varying Training-Data Percentage}

Since the MTL behavior may change when the training-data percentage (the size of the training data divided by the size of the entire dataset) changes, we evaluated the methods on both real-world datasets at different training-data percentages. Here, we present the results mostly for our low-rank algorithm (denoted by MP-MTL-LR) because it always outperforms our group-sparse algorithm (MP-MTL-GS) in the above experiments. The results corresponding to School Data are shown in Fig. \ref{fig:real_school_percent}; the results corresponding to LSOA II Data are shown in Fig. \ref{fig:real_LSOA_percent}. From those plots, we observe that on both real-world datasets, our MP-MTL method behaves similarly at different training-data percentages and outperforms DP-MTRL and DP-AGGR, especially when $\epsilon$ is small.


\begin{figure}[t!]
\centering
\subfigure[30\% - Fine scale]{\includegraphics[width=1.7in]{school_2}}
\subfigure[30\% - Coarse scale]{\includegraphics[width=1.7in]{school_2_DPMTRL_DPAGGR}}
\subfigure[50\% - Fine scale]{\includegraphics[width=1.7in]{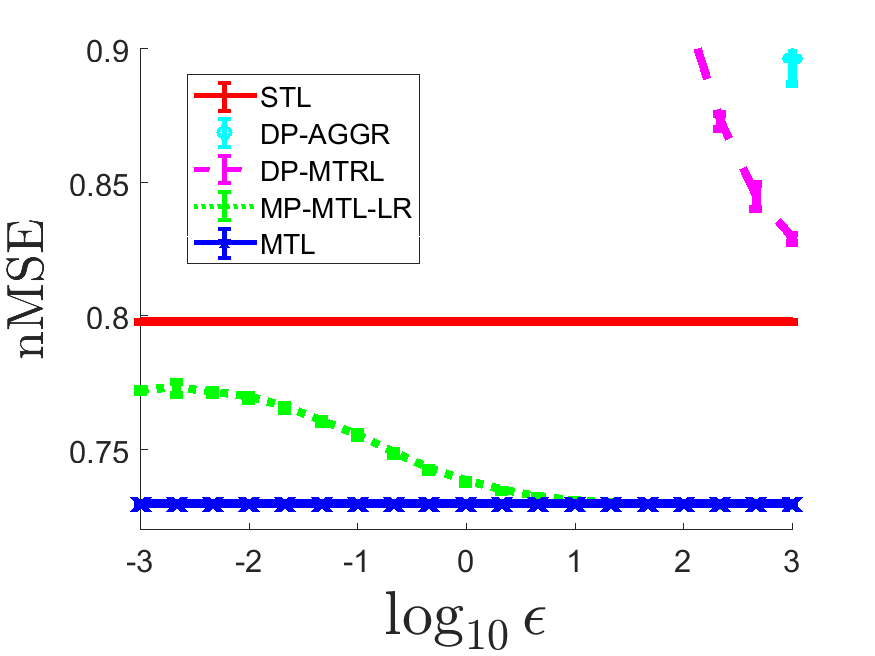}}
\subfigure[50\% - Coarse scale]{\includegraphics[width=1.7in]{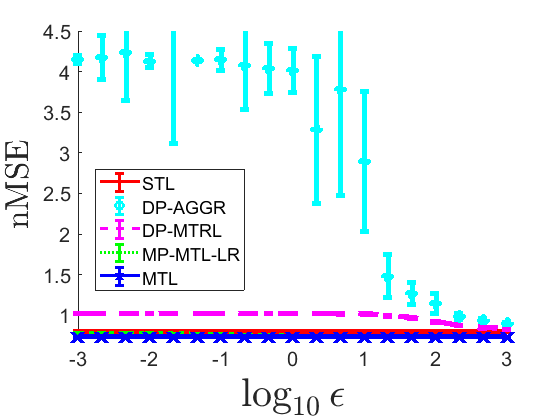}}
\subfigure[70\% - Fine scale]{\includegraphics[width=1.7in]{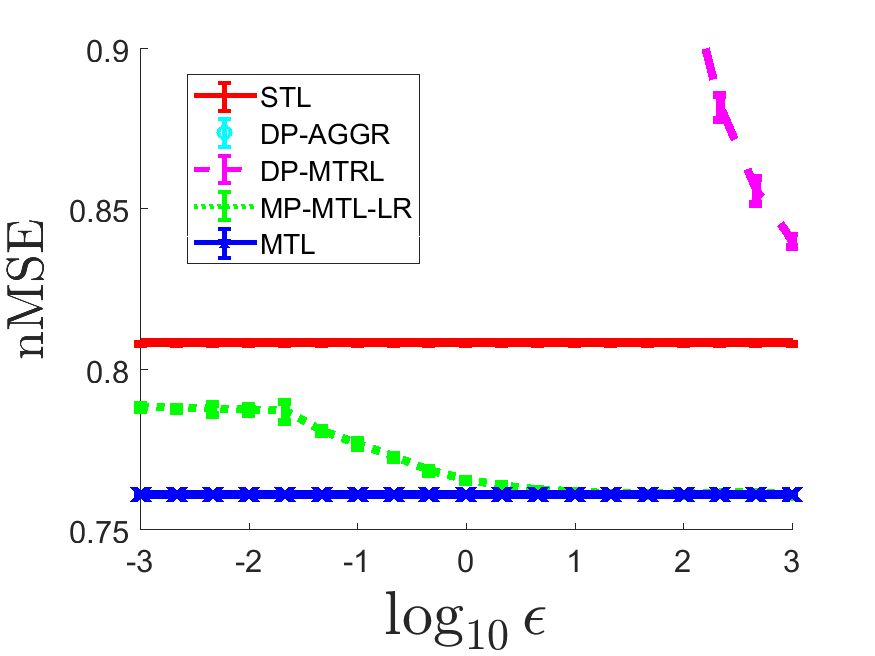}}
\subfigure[70\% - Coarse scale]{\includegraphics[width=1.7in]{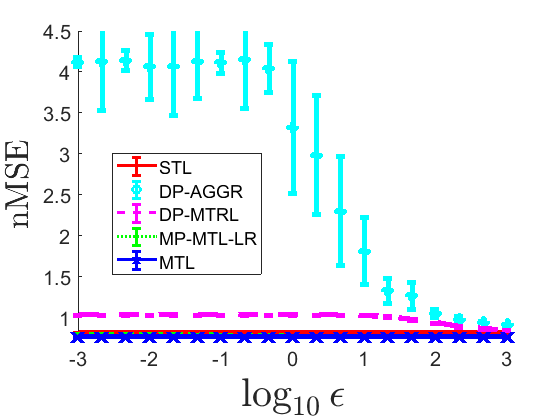}}
\subfigure[90\% - Fine scale]{\includegraphics[width=1.7in]{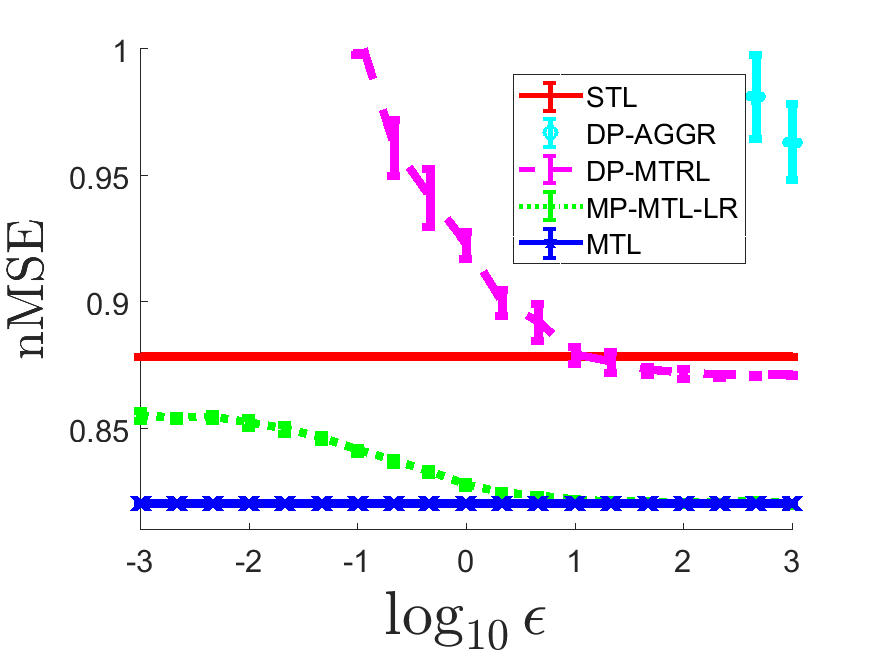}}
\subfigure[90\% - Coarse scale]{\includegraphics[width=1.7in]{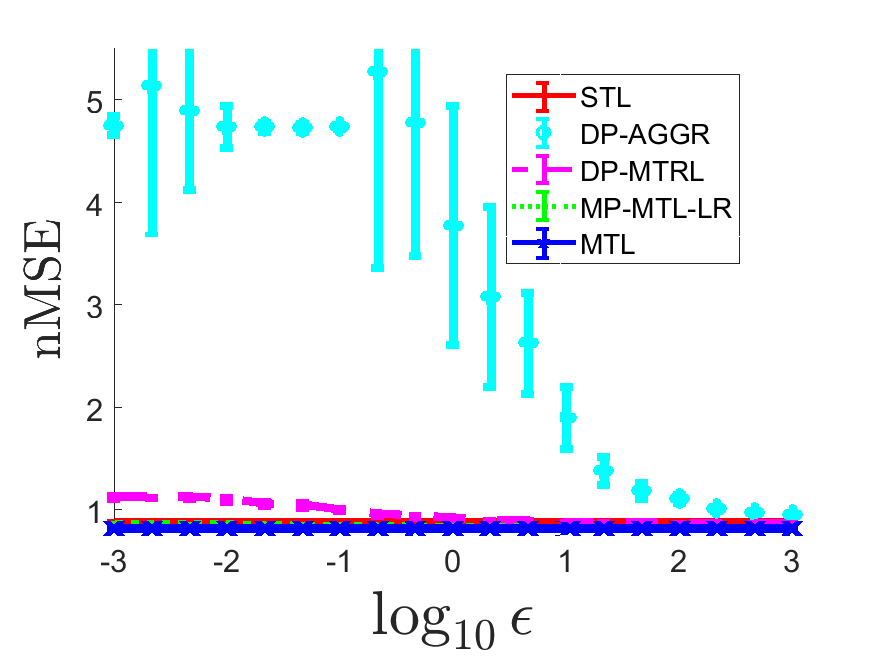}}
\caption{Privacy-accuracy tradeoff on School Data. (a) and (b) correspond to a training-data percentage of 30\%, (c) and (d) correspond to a training-data percentage of 50\%, (e) and (f) correspond to a training-data percentage of 70\%, (g) and (h) correspond to a training-data percentage of 90\%. (a), (c), (e) and (g) use fine scales of vertical axes to focus on the performances of our algorithms; (b), (d), (f) and (h) use coarse scales of vertical axes to focus on the baseline algorithms. In all the panels, MTL denotes the method with the best performance among the four non-private MTL methods proposed by~\citet{ji2009accelerated},~\citet{liu2009multi},~\citet{zhang2010convex} and DP-AGGR without perturbations; MP-MTL-LR denotes Algorithm \ref{alg:MP-MTL-LR}, whereas MP-MTL-GS denotes Algorithm \ref{alg:MP-MTL-GS}; STL denotes the method with the better performance between the $\ell_1$- and $\ell_2$-regularized methods.}\label{fig:real_school_percent}
\end{figure}

\begin{figure}[t!]
\centering
\subfigure[30\% - Fine scale]{\includegraphics[width=1.7in]{LSOA_2}}
\subfigure[30\% - Coarse scale]{\includegraphics[width=1.7in]{LSOA_2_DPMTRL_DPAGGR}}
\subfigure[50\% - Fine scale]{\includegraphics[width=1.7in]{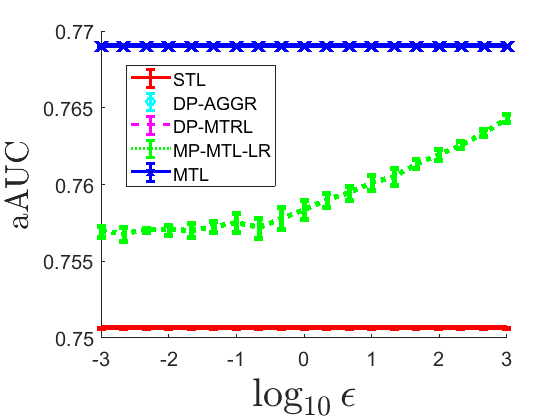}}
\subfigure[50\% - Coarse scale]{\includegraphics[width=1.7in]{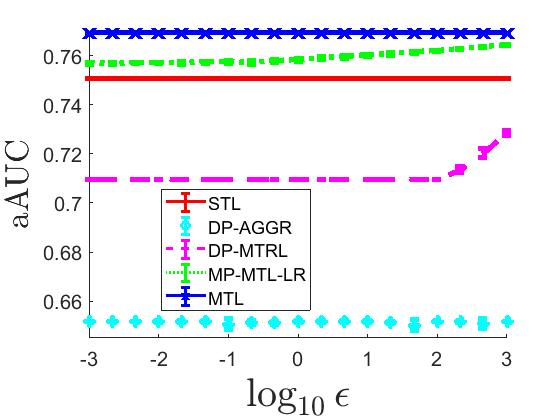}}
\subfigure[70\% - Fine scale]{\includegraphics[width=1.7in]{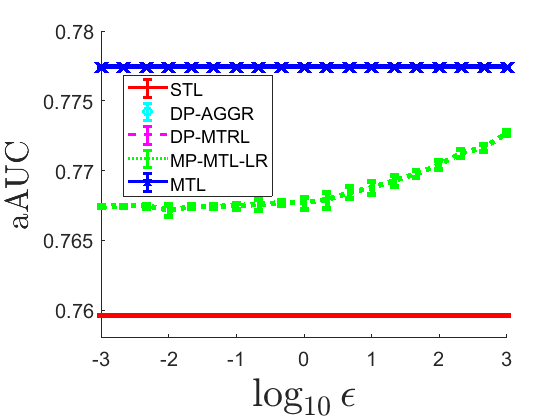}}
\subfigure[70\% - Coarse scale]{\includegraphics[width=1.7in]{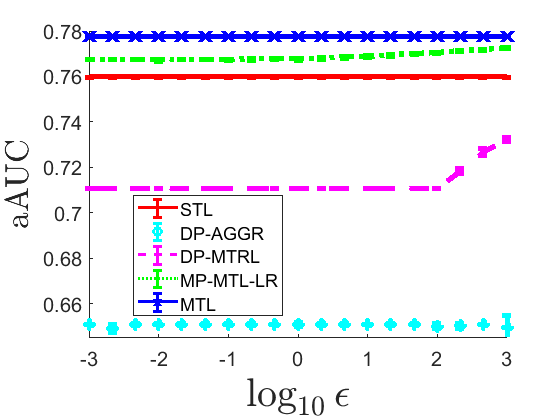}}
\subfigure[90\% - Fine scale]{\includegraphics[width=1.7in]{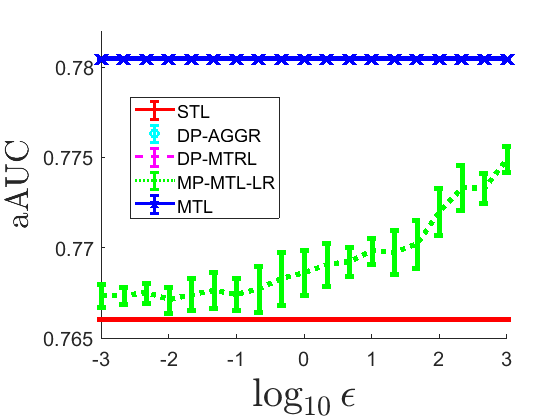}}
\subfigure[90\% - Coarse scale]{\includegraphics[width=1.7in]{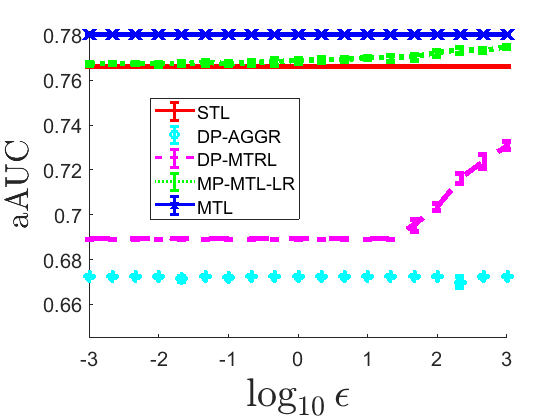}}
\caption{Privacy-accuracy tradeoff on LSOA II Data. (a) and (b) correspond to a training-data percentage of 30\%, (c) and (d) correspond to a training-data percentage of 50\%, (e) and (f) correspond to a training-data percentage of 70\%, (g) and (h) correspond to a training-data percentage of 90\%. (a), (c), (e) and (g) use fine scales of vertical axes to focus on the performances of our algorithms; (b), (d), (f) and (h) use coarse scales of vertical axes to focus on the baseline algorithms. In all the panels, MTL denotes the method with the best performance among the four non-private MTL methods proposed by~\citet{ji2009accelerated},~\citet{liu2009multi},~\citet{zhang2010convex} and DP-AGGR without perturbations; MP-MTL-LR denotes Algorithm \ref{alg:MP-MTL-LR}, whereas MP-MTL-GS denotes Algorithm \ref{alg:MP-MTL-GS}; STL denotes the method with the better performance between the $\ell_1$- and $\ell_2$-regularized methods.}\label{fig:real_LSOA_percent}
\end{figure}
}
%
%
%

\section{Properties of Differential Privacy}\label{sec:lem_df}
\begin{itemize}
\item Post-Processing immunity. This property helps us safely use the output of a differentially private algorithm without additional information leaking, as long as we do not touch the dataset $\mathcal{D}$ again.
\begin{property}[Post-Processing immunity. Proposition 2.1 in~\citet{dwork2014algorithmic}]\label{th:lem_post}

Let algorithm $\mathcal{A}_1(\calB_1): \mathcal{D} \rightarrow \theta_1 \in \mathcal{C}_1$ be an $(\epsilon,\delta)$ - differential privacy algorithm, and let $f:\mathcal{C}_1\rightarrow\mathcal{C}_2$ be an arbitrary mapping. Then, algorithm $\mathcal{A}_2(\calB_2):  \mathcal{D} \rightarrow \theta_2  \in \mathcal{C}_{2}$
is still $( \epsilon, \delta )$ - differentially private, i.e., for any set $\calS\subseteq\mathcal{C}_{2}$,
\begin{align*}
\bbP(\theta_2 \in \calS \mid \calB_2 = \mathcal{D}  )
\leq  e^{\epsilon}\bbP(\theta_2 \in \calS \mid \calB_2= \mathcal{D}' ) + \delta.
\end{align*}
\end{property}
\item Group privacy. This property guarantees the graceful increment of the privacy budget when more output variables need differentially private protection.
\begin{property}[Group privacy. Lemma 2.2 in~\citet{vadhan2016complexity}]\label{th:lem_group_dp}

Let algorithm $\mathcal{A}(\calB): \mathcal{D} \rightarrow \theta \in \mathcal{C}$ be an $(\epsilon,\delta)$ - differential privacy algorithm.
Then, considering two neighboring datasets $\mathcal{D}$ and $\mathcal{D}'$ that differ in $k$ entries,
the algorithm satisfies for any set $\calS\subseteq\mathcal{C}$
\begin{align*}
\bbP(\theta \in \calS \mid \calB = \mathcal{D}  )
\leq  e^{k\epsilon}\bbP(\theta \in \calS \mid \calB = \mathcal{D}' ) + k e^{k\epsilon}\delta.
\end{align*}
\end{property}
\item Composition. This property guarantees the linear incrementing of the privacy budget when the dataset $\mathcal{D}$ is repeatedly used.
\begin{property}[Composition. Theorem 3.16 in~\citet{dwork2014algorithmic}]\label{th:lem_comb}

 Let algorithm $\mathcal{A}_i: \mathcal{D} \rightarrow \theta_i \in \mathcal{C}_{i} $ be an $(\epsilon_i,\delta_i)$ - differential privacy algorithm for all $i\in [k]$.
 Then, $\mathcal{A}_{[k]}:  \mathcal{D} \rightarrow (\theta_1,\theta_2,\cdots,\theta_{k})  \in \bigotimes_{j=1}^{k}\mathcal{C}_{j}$
 is a $(\sum_i\epsilon_i,\sum_i\delta_i)$ - differentially private algorithm.
\end{property}
\item Adaptive composition. This property guarantees privacy when an iterative algorithm is adopted on \emph{different}
datasets that may nevertheless contain information relating to the same individual.
 \begin{property}[Adaptive composition. {Directly taken Theorem 3.5 in~\citet{kairouz2017composition}}]\label{th:lem_Ad_comp}
 Let algorithm $\mathcal{A}_1(\calB_1): \mathcal{D}_1 \rightarrow \theta_1 $ be an $(\epsilon_1,\delta_1)$ - differential privacy algorithm,
 and for $t=2,\ldots,T$,  let $\mathcal{A}_t(\calB_t): (\mathcal{D}_{t},\theta_1,\theta_2,\cdots,\theta_{t-1}) \rightarrow \theta_t \in \mathcal{C}_{t}$
 be  $(\epsilon_t,\delta_t)$ - differentially private for all given \\
 $(\theta_1,\theta_2,\cdots,\theta_{t-1})\in \bigotimes_{t'=1}^{t-1}\mathcal{C}_{t'}$ .
 Then, for all neighboring datasets $\mathcal{D}_{t}$ and $\mathcal{D}'_{t}$ that differ in a single entry relating to the same individual and for any set $\mathcal{S} \subseteq \bigotimes_{t=1}^{T}\mathcal{C}_{t}$,
\begin{equation}\label{eq:ada_comp}
\begin{split}
 \mathbb{P} ((\theta&_1,\cdots,\theta_T)\in \mathcal{S}  \mid \bigcap_{t=1}^T(\calB_t = (\mathcal{D}_t, \ttheta_{1:t-1}))  )\\
  \leq & e^{\epsilon}\mathbb{P}((\theta_1,\cdots,\theta_T)\in \mathcal{S}\mid \bigcap_{t=1}^T(\calB_t = (\mathcal{D}'_t, \ttheta_{1:t-1}) ))\\
   &+ 1-(1-\delta)\prod_{t=1}^{T}(1-\delta_t),
\end{split}
\end{equation}
 where
\begin{numcases}{\ttheta_{1:t-1} = }
\nonumber \emptyset, & $t=1$\\
\nonumber \theta_1,\theta_2,\cdots,\theta_{t-1}, & $t\geq2$,
\end{numcases}
 and
  \begin{align*}
 \epsilon = &\min\biggl\{ \sum_{t=1}^{T}\epsilon_t,\sum_{t=1}^{T}\frac{(e^{\epsilon_t}-1)\epsilon_t}{(e^{\epsilon_t}+1)}
+\sqrt{\sum_{t=1}^{T}2\epsilon_t^2\log\biggl(\frac{1}{\delta}\biggr)},\\
 &\sum_{t=1}^{T}\frac{(e^{\epsilon_t}-1)\epsilon_t}{(e^{\epsilon_t}+1)}
+\sqrt{\sum_{t=1}^{T}2\epsilon_t^2\log\biggl(e+\frac{\sqrt{\sum_{t=1}^{T}\epsilon_t^2}}{\delta}\biggr)}
\biggr\}.
 \end{align*}
\end{property}
\end{itemize}

{

\section{Lemmas for Utility Analysis}\label{sec:lem_util}

\begin{lemma}\label{th:lem_sum_poly}
For a integer $T\geq 1$, a constant $\alpha\in \mathbb{R}$, by Euler–Maclaurin formula~\citep{apostol1999elementary}, we have
\begin{align*}
\sum_{t=1}^{T}t^{\alpha} =
\left\{
  \begin{array}{ll}
   O(T^{\alpha+1}/(\alpha+1)), &  {\alpha>-1;} \\
   O(1/(-\alpha-1)), &  {\alpha<-1.}
  \end{array}
\right.
\end{align*}
\end{lemma}
\begin{proof}
This is the direct result of Euler–Maclaurin formula~\citep{apostol1999elementary}.
\end{proof}

\begin{lemma}\label{th:lem_sum_exp}
For a integer $T\geq 1$ and a constant $Q>0$, we have
\begin{align*}
\sum_{t=1}^{T}Q^{-t} \leq
\left\{
  \begin{array}{ll}
    \frac{1}{Q-1}, &  {Q>1;} \\
    \frac{Q^{-T}}{1-Q}, &  {Q<1.}
  \end{array}
\right.
\end{align*}
\end{lemma}
\begin{proof}
Because $\sum_{t=1}^{T}Q^{-t} = Q^{-1}\frac{1-Q^{-T}}{1-Q^{-1}}$, we complete the proof.
\end{proof}

\begin{lemma}\label{th:lem_bound_1_div_eps}
For constants $c_1,c_2>0$, a constant $\epsilon_0>0$, a integer $T\geq 1$, a mapping $s: t\in[T] \rightarrow s(t)>0$, a mapping $S_1: T \rightarrow S_1(T)>0$ and a mapping $S_2: T \rightarrow S_2(T)>0$,
then if $\sum_{t=1}^{T}\epsilon_0s(t)\geq c_1$ and $\sum_{t=1}^{T}s(t)\leq S_1(T)$, we have
\begin{align*}
 {1}/{\epsilon_0}\leq S_1(T)/c_1.
\end{align*}
On the other hand, if
$\sqrt{\sum_{t=1}^{T}\epsilon_0^2s^2(t)}\geq c_2$ and $\sum_{t=1}^{T}s^2(t)\leq S_2(T)$, we have
\begin{align*}
 {1}/{\epsilon_0}\leq \sqrt{S_2(T)}/c_2.
\end{align*}
\end{lemma}
\begin{proof}
If $\sum_{t=1}^{T}\epsilon_0s(t)\geq c_1$, $1/{\epsilon_0}\leq \sum_{t=1}^{T}s(t)/c_1 \leq S_1(T)/c_1 $.

On the other hand, if $\sqrt{\sum_{t=1}^{T}\epsilon_0^2s^2(t)}\geq c_2$,
$1/{\epsilon_0}\leq \sqrt{\sum_{t=1}^{T}s^2(t)}/c_2 \leq \sqrt{S_2(T)}/c_2$.
\end{proof}

\begin{lemma}\label{th:lem_bound_vareps_t_LR}
Consider Algorithm \ref{alg:MP-MTL-LR}. For an index $k\leq q$ that suffices the definition in Lemma \ref{th:lem_LR_iter_t} for all $t\in[T]$, $\eta =1/L$, $\lambda = \Theta(LK\sqrt{m} )$, set $\epsilon_t \leq 4Kk^2d(\log d)/q^2$ for $t\in[T]$.
Assume in each iteration, $\E$ is the defined Wishart random matrix. We have with probability at least $1-d^{-c}$ for some constant $c>1$ that
\begin{equation}\label{eq:varepsilon_LR}
 \begin{split}
\varepsilon_t = &\frac{1}{2\eta}\|\widehat{\W}^{(t)} - \C\|_F^2 + \lambda\|\widehat{\W}^{(t)}\|_*  \\
&  - \biggl\{\min_{\W} \frac{1}{2\eta}\|\W - \C\|_F^2 + \lambda\|\W\|_*\biggr\}\\
= &O\biggl( \frac{K^3\sqrt{m}kd\log d}{\eta\epsilon_t}\biggr).
 \end{split}
\end{equation}
\end{lemma}
\begin{proof}
First, using Lemma 1 of \citet{jiang2016wishart}, we have in the $t$-th step, with probability at least $1-d^{-c}$ for some constant $c>1$,
\begin{align*}
  \sigma_1(\E)  = O\biggl(d(\log d) \sigma_1\biggl(\frac{K^2}{2\epsilon_t}\I_d\biggr)\biggr) = O(d(\log d) K^2/\epsilon_t).
\end{align*}
We also have $\sigma_1(\C)\leq \|\C\|_F \leq \sqrt{m}\max_i\|\C_i\|_2 \leq K\sqrt{m}$, where $\C_i$ is the $i$-th column of $\C$.

As such, by Lemma \ref{th:lem_LR_iter_t}, in the $t$-th iteration, for $\epsilon_t \leq 4Kk^2d(\log d)/q^2 $, where $q = \min\{d,m\}$, we have
\begin{align*}
\varepsilon_t = &\frac{1}{2\eta}\|\widehat{\W}^{(t)} - \C\|_F^2 + \lambda\|\widehat{\W}^{(t)}\|_*  \\
&  - \biggl\{\min_{\W} \frac{1}{2\eta}\|\W - \C\|_F^2 + \lambda\|\W\|_*\biggr\}\\
\leq &\frac{1}{\eta}\biggl(\frac{\sigma_1^2(\C)}{\eta\lambda}+ \sigma_1(\C)\biggl)\biggl[ k\frac{\sigma_1(\E)}{2\eta\lambda} \\
 & + (r_c-k)I(r_c>k)\sqrt{\sigma_1(\E)}
 + \biggl(\frac{k(k-1)}{
 \eta\lambda}
+ 2k \biggr) \sigma_1(\E)\biggr]\\
\leq & \frac{1}{\eta}\biggl(\frac{K^2m}{\eta\lambda}+ K\sqrt{m}\biggl)\biggl[ k\frac{\sigma_1(\E)}{2\eta\lambda} \\
 & + q\sqrt{\sigma_1(\E)}
 + \biggl(\frac{k(k-1)}{
 \eta\lambda}
+ 2k \biggr) \sigma_1(\E)\biggr]\\
=&O\biggl(
\frac{1}{\eta}\biggl(\frac{K^2m}{\eta\lambda}+ K\sqrt{m}\biggl) \biggl(\frac{k^2}{
 \eta\lambda}
+ 2k \biggr)   \frac{d(\log d) K^2}{\epsilon_t}
\biggr),
\end{align*}
where in the second inequality, the terms with $\sigma_1(\E)$ dominate due to the condition on $\epsilon_t$.

Further assuming $\eta = 1/L$ and $\lambda = \Theta(LK\sqrt{m} )$, we complete the proof.
\end{proof}

\begin{lemma}\label{th:lem_bound_vareps_t_GS}
Consider Algorithm \ref{alg:MP-MTL-GS}. For an index $k\leq d$ that suffices the definition in Lemma \ref{th:lem_GS_iter_t} for all $t\in[T]$, $\eta =1/L$, $\lambda = \Theta(LKd\sqrt{m} )$, set $\epsilon_t \leq {k^2\log(d)}/{4Kd(d-k)^2m}$ for $t\in[T]$.
Assume in each iteration, $\E$ is the defined Wishart random matrix. We have with probability at least $1-d^{-c}$ for some constant $c>1$ that
  \begin{equation}\label{eq:varepsilon_GS}
  \begin{split}
\varepsilon_t =&\frac{1}{2\eta}\|\widehat{\W}^{(t)} - \C\|_F^2 + \lambda\|\widehat{\W}^{(t)}\|_{2,1}  \\
&  - \biggl\{\min_{\W} \frac{1}{2\eta}\|\W - \C\|_F^2 + \lambda\|\W\|_{2,1}\biggr\}\\
=&O\biggl( \frac{K^2k\log d}{\eta\epsilon_t}\biggr).
\end{split}
\end{equation}
\end{lemma}
\begin{proof}
Similarly as in proof for Lemma \ref{th:lem_bound_vareps_t_LR}, by Lemma \ref{th:lem_GS_iter_t}, in the $t$-th iteration, we have
\begin{align*}
\varepsilon_t = &\frac{1}{2\eta}\|\widehat{\W}^{(t)} - \C\|_F^2 + \lambda\|\widehat{\W}^{(t)}\|_{2,1}  \\
&  - \biggl\{\min_{\W} \frac{1}{2\eta}\|\W - \C\|_F^2 + \lambda\|\W\|_{2,1}\biggr\}\\
\leq &\frac{1}{\eta}\biggl[\frac{r_{c,s}}{ \eta\lambda}\biggl(\max_{j\in[d]}\|\C^j\|_{2}\biggr)^2+ \biggl(\max_{j\in[d]}\|\C^j\|_{2}\biggr)\biggl]\\
 &\cdot\biggl[ \frac{k}{2\eta\lambda} \max_{j:\eta^2\lambda^2\leq\SSigma_{jj,0}}|\E_{jj}|\\
& + (r_{c,s}-k)I(r_{c,s}>k) \max_{j:\eta^2\lambda^2>\SSigma_{jj,0}}\sqrt{|\E_{jj}|}\biggr]\\
\leq &\frac{1}{\eta}\biggl[\frac{r_{c,s}}{ \eta\lambda}\|\C\|_F^2+ \|\C\|_F\biggl]\\
 &\cdot\biggl[ \frac{k}{2\eta\lambda} \sigma_1(\E)
 + (r_{c,s}-k)I(r_{c,s}>k)\sqrt{\sigma_1(\E)}\biggr]\\
\leq & \frac{1}{\eta}\biggl(\frac{dK^2m}{\eta\lambda}+ K\sqrt{m}\biggl)\biggl[ k\frac{\sigma_1(\E)}{2\eta\lambda} + (d-k)\sqrt{\sigma_1(\E)} \biggr].
\end{align*}
Further setting $\eta = 1/L$ and $\lambda = \Theta(LKd\sqrt{m} )$, assuming $\epsilon_t \leq  {k^2\log(d)}/{4Kd(d-k)^2m}$, we have
  \begin{equation}\label{eq:varepsilon_GS}
  \begin{split}
\varepsilon_t =& O\biggl(
\frac{1}{\eta}\biggl(\frac{dK^2m}{\eta\lambda}+ K\sqrt{m}\biggl)  \frac{k}{
 \eta\lambda}
   \frac{d(\log d) K^2}{\epsilon_t}
\biggr)\\
=&O\biggl( \frac{K^2k\log d}{\eta\epsilon_t}\biggr).
\end{split}
\end{equation}
\end{proof}

\begin{lemma}\label{th:lem_dist_W}
For matrices $\W_1,\W_2\in \mathcal{W} \subset \mathbb{R}^{d\times m}$, we have
\begin{align*}
\|\W_1-\W_2\|_F = O(K\sqrt{m}).
\end{align*}
\end{lemma}
\begin{proof}
Because $\W_1,\W_2\in \mathcal{W}$, $\max_{i\in[m]}\|\w_{i,1}\|_2\leq K$. Therefore,
\begin{align*}
\|\W_1-\W_2\|_F \leq 2\|\W_1\|_F \leq 2\sqrt{m}\max_{i\in[m]}\|\w_{i,1}\|_2\leq 2K\sqrt{m}.
\end{align*}
\end{proof}

\begin{lemma}\label{th:lem_bound_sum_eps_case_2_and_3}
For constants $\epsilon,\delta\geq 0$, a integer $T\geq 1$, a series constants $\epsilon_t>0$ for $t\in[T]$,
then if
\begin{align*}
\epsilon = \sum_{t=1}^{T}\frac{(e^{\epsilon_t}-1)\epsilon_t}{(e^{\epsilon_t}+1)}
+\sqrt{\sum_{t=1}^{T}2\epsilon_t^2\log\biggl(\frac{1}{\delta}\biggr)},
\end{align*}
we have
\begin{align*}
\sqrt{\sum_{t=1}^{T}\epsilon_t^2} \geq \frac{\sqrt{2}\epsilon}{2\sqrt{\log(1/\delta)+2\epsilon} }.
\end{align*}
On the other hand, if
\begin{align*}
  \epsilon =\sum_{t=1}^{T}\frac{(e^{\epsilon_t}-1)\epsilon_t}{(e^{\epsilon_t}+1)}
+\sqrt{\sum_{t=1}^{T}2\epsilon_t^2\log\biggl(e+\frac{\sqrt{\sum_{t=1}^{T}\epsilon_t^2}}{\delta}\biggr)},
\end{align*}
we have
\begin{align*}
 \sqrt{\sum_{t=1}^{T}\epsilon_t^2} \geq \max\biggl\{\sqrt{\frac{\epsilon}{1+\sqrt{2}/(e\delta)}},
\frac{\sqrt{2}\epsilon}{2\sqrt{\log(e+\epsilon/\sqrt{2}\delta)+2\epsilon} }
\biggr\}.
\end{align*}
\end{lemma}
\begin{proof}
 If $\epsilon = \sum_{t=1}^{T}\frac{(e^{\epsilon_t}-1)\epsilon_t}{(e^{\epsilon_t}+1)}
+\sqrt{\sum_{t=1}^{T}2\epsilon_t^2\log\biggl(\frac{1}{\delta}\biggr)}$,

Because $(e^x-1)/(e^x+1)\leq x$ for $x\geq 0$, then
\begin{align*}
 \epsilon \leq \sum_{t=1}^{T}\epsilon_t^2
+\sqrt{\sum_{t=1}^{T}2\epsilon_t^2\log\biggl(\frac{1}{\delta}\biggr)} .
\end{align*}
Solving the inequality with respect to $\sqrt{\sum_{t=1}^{T}\epsilon_t^2}$, we get
\begin{align*}
\sqrt{\sum_{t=1}^{T}\epsilon_t^2}& \geq \frac{\sqrt{2}\epsilon}{\sqrt{\log(1/\delta)+2\epsilon}+\sqrt{\log(1/\delta)}}\\
&\geq \frac{\sqrt{2}\epsilon}{2\sqrt{\log(1/\delta)+2\epsilon} }.
\end{align*}

If we have
\begin{equation}\label{eq:eps_of_case_3}
  \epsilon =\sum_{t=1}^{T}\frac{(e^{\epsilon_t}-1)\epsilon_t}{(e^{\epsilon_t}+1)}
+\sqrt{\sum_{t=1}^{T}2\epsilon_t^2\log\biggl(e+\frac{\sqrt{\sum_{t=1}^{T}\epsilon_t^2}}{\delta}\biggr)}.
\end{equation}

Because $(e^x-1)/(e^x+1)\leq x$ for $x\geq 0$, then
\begin{align*}
 \epsilon \leq & \sum_{t=1}^{T}\epsilon_t^2
+\sqrt{\sum_{t=1}^{T}2\epsilon_t^2\log\biggl(e+\frac{\sqrt{\sum_{t=1}^{T}\epsilon_t^2}}{\delta}\biggr)}\\
\leq & \sum_{t=1}^{T}\epsilon_t^2
+\frac{\sqrt{2}\sum_{t=1}^{T}\epsilon_t^2}{e\delta},
\end{align*}
where the second inequality is because $\log(e+x)\leq x/e + 1$ for $x\geq 0$.

As such,
\begin{align*}
\sqrt{\sum_{t=1}^{T}\epsilon_t^2} \geq \sqrt{\frac{\epsilon}{1+\sqrt{2}/(e\delta)}}.
\end{align*}

On the other hand, by \eqref{eq:eps_of_case_3}, it also holds that $\sqrt{2}\sqrt{\sum_{t=1}^{T}\epsilon_t^2}\leq \epsilon$. Then we have
\begin{align*}
 \epsilon \leq & \sum_{t=1}^{T}\epsilon_t^2
+\sqrt{\sum_{t=1}^{T}2\epsilon_t^2\log\biggl(e+\frac{\sqrt{\sum_{t=1}^{T}\epsilon_t^2}}{\delta}\biggr)}\\
\leq & \sum_{t=1}^{T}\epsilon_t^2
+\sqrt{\sum_{t=1}^{T}2\epsilon_t^2\log\biggl(e+\frac{\epsilon}{\sqrt{2}\delta}\biggr)}.
\end{align*}
Solving the inequality with respect to $\sqrt{\sum_{t=1}^{T}\epsilon_t^2}$, we also get
\begin{align*}
\sqrt{\sum_{t=1}^{T}\epsilon_t^2} &\geq \frac{\sqrt{2}\epsilon}{\sqrt{\log(e+\epsilon/\sqrt{2}\delta)+2\epsilon}+\sqrt{\log(e+\epsilon/\sqrt{2}\delta)}}\\
& \geq \frac{\sqrt{2}\epsilon}{2\sqrt{\log(e+\epsilon/\sqrt{2}\delta)+2\epsilon} }.
\end{align*}
\end{proof}

\begin{lemma}\label{th:lem_bound_error_poly}
For constants $\kappa,\epsilon_0 > 0$, $c_1,c_2>0$, $\alpha\in\mathbb{R}$, a integer $T\geq 1$, assuming $\epsilon_t = \epsilon_0t^{\alpha}$, $\varepsilon_t = O( \kappa/\epsilon_t )$ for $t\in[T]$,
if $\sum_{t=1}^{T}\epsilon_t\geq c_1$, we have
\begin{align*}
\sum_{t=1}^{T}\sqrt{\varepsilon_t} =
\left\{
  \begin{array}{ll}
    O\biggl( \sqrt{\frac{\kappa T^{\alpha+1} }{c_1(\alpha/2 - 1)^2(\alpha+1)  }}\biggr), &  {\alpha>2;} \\
    O\biggl( \sqrt{\frac{\kappa T^{3} }{c_1(\alpha/2 - 1)^2(\alpha+1)  }}\biggr), &  {-1<\alpha<2;} \\
    O\biggl( \sqrt{\frac{\kappa T^{2-\alpha} }{c_1(\alpha/2 - 1)^2(-\alpha-1)  }}\biggr), &  {\alpha<-1,}
  \end{array}
\right.
\end{align*}
and
\begin{align*}
\sum_{t=1}^{T}t\sqrt{\varepsilon_t} =
\left\{
  \begin{array}{ll}
    O\biggl( \sqrt{\frac{\kappa T^{\alpha+1} }{c_1(\alpha/2 - 2)^2(\alpha+1)  }}\biggr), &  {\alpha>4;} \\
    O\biggl( \sqrt{\frac{\kappa T^{5} }{c_1(\alpha/2 - 2)^2(\alpha+1)  }}\biggr), &  {-1<\alpha<4;} \\
    O\biggl( \sqrt{\frac{\kappa T^{4-\alpha} }{c_1(\alpha/2 - 2)^2(-\alpha-1)  }}\biggr), &  {\alpha<-1.}
  \end{array}
\right.
\end{align*}

If $\sqrt{\sum_{t=1}^{T}\epsilon_t^2}\geq c_2$, we have
\begin{align*}
\sum_{t=1}^{T}\sqrt{\varepsilon_t} =
\left\{
  \begin{array}{ll}
    O\biggl( \sqrt{\frac{\kappa T^{\alpha+1/2} }{c_2(\alpha/2 - 1)^2\sqrt{2\alpha+1}  }}\biggr), &  {\alpha>2;} \\
    O\biggl( \sqrt{\frac{\kappa T^{5/2} }{c_2(\alpha/2 - 1)^2\sqrt{2\alpha+1}  }}\biggr), &  {-1/2<\alpha<2;} \\
    O\biggl( \sqrt{\frac{\kappa T^{2-\alpha} }{c_2(\alpha/2 - 1)^2\sqrt{-2\alpha-1}  }}\biggr), &  {\alpha<-1/2,}
  \end{array}
\right.
\end{align*}
and
\begin{align*}
\sum_{t=1}^{T}t\sqrt{\varepsilon_t} =
\left\{
  \begin{array}{ll}
    O\biggl( \sqrt{\frac{\kappa T^{\alpha+1/2} }{c_2(\alpha/2 - 2)^2\sqrt{2\alpha+1}  }}\biggr), &  {\alpha>4;} \\
    O\biggl( \sqrt{\frac{\kappa T^{9/2} }{c_2(\alpha/2 - 2)^2\sqrt{2\alpha+1}  }}\biggr), &  {-1/2<\alpha<4;} \\
    O\biggl( \sqrt{\frac{\kappa T^{4-\alpha} }{c_2(\alpha/2 - 2)^2\sqrt{-2\alpha-1}  }}\biggr), &  {\alpha<-1/2.}
  \end{array}
\right.
\end{align*}
\end{lemma}
\begin{proof}
If $\sum_{t=1}^{T}\epsilon_t = \sum_{t=1}^{T}\epsilon_0t^{\alpha} \geq c_1$. We have
\begin{align*}
\sum_{t=1}^{T}\sqrt{\varepsilon_t}&= O\biggl(\sum_{t=1}^{T}\sqrt{ \kappa/\epsilon_t}\biggr) = O\biggl(\sum_{t=1}^{T}\sqrt{ \frac{\kappa}{\epsilon_0t^{\alpha}}}\biggr)\\
&=O\biggl(\sum_{t=1}^{T}t^{-\alpha/2}\sqrt{ \frac{\kappa}{\epsilon_0}}\biggr).
\end{align*}

Using Lemma \ref{th:lem_bound_1_div_eps}, we have
\begin{align*}
\sum_{t=1}^{T}\sqrt{\varepsilon_t} = O\biggl(\sum_{t=1}^{T}t^{-\alpha/2}\sqrt{ \frac{\kappa}{c_1}\sum_{t=1}^{T}t^{\alpha}}\biggr).
\end{align*}

Then using Lemma \ref{th:lem_sum_poly}, if $\alpha>2$, i.e., $-\alpha/2<-1$, we have
\begin{align*}
\sum_{t=1}^{T}\sqrt{\varepsilon_t} &= O\biggl(\frac{1}{-(-\alpha/2)-1}\sqrt{ \frac{\kappa}{c_1}\frac{T^{\alpha+1}}{\alpha+1}}\biggr)\\
& = O\biggl( \sqrt{ \frac{\kappa}{c_1}\frac{T^{\alpha+1}}{(\alpha/2-1)^2(\alpha+1)}}\biggr).
\end{align*}

Results under other conditions can be proved similarly.
\end{proof}

\begin{lemma}\label{th:lem_bound_error_exp}
For constants $\kappa,\epsilon_0 > 0$, $c_1,c_2>0$, $Q_0\in(0,1)$, $Q>0$, a integer $T\geq 1$, assuming $\epsilon_t = \epsilon_0Q^{-t}$, $\varepsilon_t = O( \kappa/\epsilon_t )$ for $t\in[T]$,
if $\sum_{t=1}^{T}\epsilon_t\geq c_1$, we have
\begin{align*}
\sum_{t=1}^{T}Q_0^{-t}\sqrt{\varepsilon_t} =
\left\{
  \begin{array}{ll}
    O\biggl( \sqrt{\frac{\kappa Q^{-T} }{c_1 (Q_0/\sqrt{Q} - 1)^2(1-Q) } }\biggr), &  {0<Q<Q_0^2;} \\
    O\biggl( \sqrt{\frac{\kappa (Q_0^2)^{-T} }{c_1 (Q_0/\sqrt{Q} - 1)^2(1-Q) } }\biggr), &  {Q_0^2<Q<1;} \\
    O\biggl( \sqrt{\frac{\kappa (Q_0^2/Q)^{-T} }{c_1 (Q_0/\sqrt{Q} - 1)^2(Q-1) } }\biggr), &  {Q>1,}
  \end{array}
\right.
\end{align*}
and
\begin{align*}
\sum_{t=1}^{T}\sqrt{\varepsilon_tQ_0^{-t}} =
\left\{
  \begin{array}{ll}
    O\biggl( \sqrt{\frac{\kappa Q^{-T} }{c_1 (\sqrt{Q_0}/\sqrt{Q} - 1)^2(1-Q) } }\biggr), &  {0<Q<Q_0;} \\
    O\biggl( \sqrt{\frac{\kappa Q_0^{-T} }{c_1 (\sqrt{Q_0}/\sqrt{Q} - 1)^2(1-Q) } }\biggr), &  {Q_0<Q<1;} \\
    O\biggl( \sqrt{\frac{\kappa (Q_0/{Q})^{-T} }{c_1 (\sqrt{Q_0}/\sqrt{Q} - 1)^2(Q-1) } }\biggr), &  {Q>1.}
  \end{array}
\right.
\end{align*}

If $\sqrt{\sum_{t=1}^{T}\epsilon_t^2}\geq c_2$, we have
\begin{align*}
\sum_{t=1}^{T}Q_0^{-t}\sqrt{\varepsilon_t} =
\left\{
  \begin{array}{ll}
    O\biggl( \sqrt{\frac{\kappa Q^{-T} }{c_2 (Q_0/\sqrt{Q} - 1)^2\sqrt{1-Q^2} } }\biggr), &  {0<Q<Q_0^2;} \\
    O\biggl( \sqrt{\frac{\kappa (Q_0^2)^{-T} }{c_2 (Q_0/\sqrt{Q} - 1)^2\sqrt{1-Q^2} } }\biggr), &  {Q_0^2<Q<1;} \\
    O\biggl( \sqrt{\frac{\kappa (Q_0^2/{Q})^{-T} }{c_2 (Q_0/\sqrt{Q} - 1)^2\sqrt{ Q^2-1} } }\biggr), &  {Q>1,}
  \end{array}
\right.
\end{align*}
and
\begin{align*}
\sum_{t=1}^{T}\sqrt{\varepsilon_tQ_0^{-t}} =
\left\{
  \begin{array}{ll}
    O\biggl( \sqrt{\frac{\kappa Q^{-T} }{c_2 (\sqrt{Q_0}/\sqrt{Q} - 1)^2\sqrt{1-Q^2} } }\biggr), &  {0<Q<Q_0;} \\
    O\biggl( \sqrt{\frac{\kappa Q_0^{-T} }{c_2 (\sqrt{Q_0}/\sqrt{Q} - 1)^2\sqrt{1-Q^2} } }\biggr), &  {Q_0<Q<1;} \\
    O\biggl( \sqrt{\frac{\kappa (Q_0/{Q})^{-T} }{c_2 (\sqrt{Q_0}/\sqrt{Q} - 1)^2\sqrt{ Q^2-1} } }\biggr), &  {Q>1.}
  \end{array}
\right.
\end{align*}
\end{lemma}
\begin{proof}
If $\sum_{t=1}^{T}\epsilon_t = \sum_{t=1}^{T}\epsilon_0Q^{-t} \geq c_1$. We have
\begin{align*}
\sum_{t=1}^{T}Q_0^{-t}\sqrt{\varepsilon_t}&= O\biggl(\sum_{t=1}^{T}Q_0^{-t}\sqrt{ \kappa/\epsilon_t}\biggr) = O\biggl(\sum_{t=1}^{T}Q_0^{-t}\sqrt{ \frac{\kappa}{\epsilon_0Q^{-t}}}\biggr)\\
&=O\biggl(\sum_{t=1}^{T}(Q_0/\sqrt{Q})^{-t}\sqrt{ \frac{\kappa}{\epsilon_0}}\biggr).
\end{align*}

Using Lemma \ref{th:lem_bound_1_div_eps}, we have
\begin{align*}
\sum_{t=1}^{T}\sqrt{\varepsilon_t} = O\biggl(\sum_{t=1}^{T}(Q_0/\sqrt{Q})^{-t}\sqrt{ \frac{\kappa}{c_1}\sum_{t=1}^{T}Q^{-t} }\biggr).
\end{align*}

Then using Lemma \ref{th:lem_sum_exp}, if $Q<Q_0^2<1$, i.e., $Q_0/\sqrt{Q}>1$, we have
\begin{align*}
\sum_{t=1}^{T}\sqrt{\varepsilon_t} &= O\biggl(\frac{1}{Q_0/\sqrt{Q}-1}\sqrt{ \frac{\kappa}{c_1}\frac{Q^{-T}}{1-Q}}\biggr)\\
& = O\biggl( \sqrt{\frac{\kappa Q^{-T} }{c_1 (Q_0/\sqrt{Q} - 1)^2(1-Q) } }\biggr).
\end{align*}

Results under other conditions can be proved similarly.
\end{proof}

\begin{lemma}\label{th:lem_pseudo_bound_basic_convex}
For constants $L,c_3,c_4>0$, a integer $T\geq1$, matrices $\widetilde{\W}^{(0)},\W_*\in \mathcal{W} \subset \mathbb{R}^{d\times m}$, if it holds for a series of positive constants $\{\varepsilon_t\}$ that $\sum_{t=1}^{T}\sqrt{\varepsilon_t} = O(\sqrt{c_4T^{c_3}}) $, setting $T = \Theta((K^2Lm/c_4)^{1/c_3})$,
we have
\begin{align*}
\mathcal{E}& = \frac{L}{2mT}\biggl(\|\widetilde{\W}^{(0)}-\W_*\|_F + 2\sum_{t=1}^{T}\sqrt{\frac{2\varepsilon_t}{L}}+\sqrt{2\sum_{t=1}^{T}\frac{\varepsilon_t}{L}}\biggr)^2\\
& = O\biggl( K^2L\biggl[ \frac{c_4}{K^2Lm}\biggr]^{1/c_3}  \biggr).
\end{align*}
\end{lemma}
\begin{proof}
First, because $\varepsilon_t>0$ for $t\in[T]$, we have
\begin{align*}
 \sqrt{\sum_{t=1}^{T}\varepsilon_t}\leq \sum_{t=1}^{T}\sqrt{\varepsilon_t}.
\end{align*}
Then combining Lemma \ref{th:lem_dist_W} and Lemma \ref{th:lem_sum_poly}, it suffices that
\begin{align*}
\mathcal{E}& = O\biggl(\frac{L}{mT}\biggl[K\sqrt{m} + \frac{1}{\sqrt{L}} \sum_{t=1}^{T}\sqrt{\varepsilon_t}   \biggr]^2  \biggr)\\
&=O\biggl( \biggl[K\sqrt{\frac{L}{T}} + \frac{1}{\sqrt{mT}} \sum_{t=1}^{T}\sqrt{\varepsilon_t}   \biggr]^2  \biggr)\\
&=O\biggl( \biggl[K\sqrt{\frac{L}{T}} + \frac{1}{\sqrt{mT}} \sqrt{c_4T^{c_3}}  \biggr]^2  \biggr).
\end{align*}
Then setting $T = \Theta((K^2Lm/c_4)^{1/c_3})$, we complete the proof.
\end{proof}

\begin{lemma}\label{th:lem_pseudo_bound_acc_convex}
For constants $L,c_3,c_4>0$, a integer $T\geq1$, matrices $\widetilde{\W}^{(0)},\W_*\in \mathcal{W} \subset \mathbb{R}^{d\times m}$, if it holds for a series of positive constants $\{\varepsilon_t\}$ that $\sum_{t=1}^{T}\sqrt{\varepsilon_t} = O(\sqrt{c_4T^{c_3}}) $, setting $T = \Theta((K^2Lm/c_4)^{1/c_3})$,
we have
\begin{align*}
\mathcal{E} =& \frac{2L}{m(T+1)^2}\biggl(\|\widetilde{\W}^{(0)}-\W_*\|_F \\
&+ 2\sum_{t=1}^{T}t\sqrt{\frac{2\varepsilon_t}{L}}+\sqrt{2\sum_{t=1}^{T}t^2\frac{\varepsilon_t}{L}}\biggr)^2\\
 =& O\biggl( K^2L\biggl[ \frac{c_4}{K^2Lm}\biggr]^{2/c_3}  \biggr).
\end{align*}
\end{lemma}
\begin{proof}
First, because $\varepsilon_t>0$ for $t\in[T]$, we have
\begin{align*}
 \sqrt{\sum_{t=1}^{T}t^2\varepsilon_t}\leq \sum_{t=1}^{T}\sqrt{t^2\varepsilon_t} =\sum_{t=1}^{T}t\sqrt{\varepsilon_t}  .
\end{align*}
Then combining Lemma \ref{th:lem_dist_W} and Lemma \ref{th:lem_sum_poly}, it suffices that
\begin{align*}
\mathcal{E}& = O\biggl(\frac{L}{mT^2}\biggl[K\sqrt{m} + \frac{1}{\sqrt{L}} \sum_{t=1}^{T}t\sqrt{\varepsilon_t}   \biggr]^2  \biggr)\\
&=O\biggl( \biggl[K\frac{\sqrt{L}}{T}+ \frac{1}{\sqrt{m}T} \sum_{t=1}^{T}t\sqrt{\varepsilon_t}   \biggr]^2  \biggr)\\
&=O\biggl( \biggl[K\frac{\sqrt{L}}{T}+ \frac{1}{\sqrt{m}T} \sqrt{c_4T^{c_3}}  \biggr]^2  \biggr).
\end{align*}
Then setting $T = \Theta((K^2Lm/c_4)^{1/c_3})$, we complete the proof.
\end{proof}

\begin{lemma}\label{th:lem_pseudo_bound_basic_strong_convex}
For constants $L,c_6>0$, a constant $c_5\in(0,1)$, a constant $Q_0\in (0,1)$, a integer $T\geq1$, matrices $\widetilde{\W}^{(0)},\W_*\in \mathcal{W} \subset \mathbb{R}^{d\times m}$, if it holds for a series of positive constants $\{\varepsilon_t\}$ that $\sum_{t=1}^{T}Q_0^{-t}\sqrt{\varepsilon_t} = O(\sqrt{c_6 {c_5}^{-T}}) $, setting $T = \Theta(\log_{1/c_5}(K^2Lm/c_6))$,
we have
\begin{align*}
\mathcal{E}& =\frac{Q_0^T}{\sqrt{m}}\biggl(\|\widetilde{\W}^{(0)}-\W_*\|_F + 2\sum_{t=1}^{T}Q_0^{-t}\sqrt{\frac{2\varepsilon_t}{L}}\biggr) \\
& = O\biggl( K\biggl[ \frac{c_6}{K^2Lm}\biggr]^{\log_{c_5}Q_0}  \biggr).
\end{align*}
\end{lemma}
\begin{proof}
Using Lemma \ref{th:lem_dist_W} and Lemma \ref{th:lem_sum_exp}, it suffices that
\begin{align*}
\mathcal{E}& = O\biggl(\frac{Q_0^T}{\sqrt{m}}\biggl[K\sqrt{m} + \frac{1}{\sqrt{L}} \sum_{t=1}^{T}Q_0^{-t}\sqrt{\varepsilon_t}   \biggr]   \biggr)\\
&=O\biggl( Q_0^T\biggl[K  + \frac{1}{\sqrt{mL}} \sum_{t=1}^{T}Q_0^{-t}\sqrt{\varepsilon_t}   \biggr]   \biggr)\\
&=O\biggl( Q_0^T\biggl[K  + \frac{1}{\sqrt{mL}} \sqrt{c_6 {c_5}^{-T}} \biggr] \biggr).
\end{align*}
Then setting $T = \Theta(\log_{1/c_5}(K^2Lm/c_6))$, we complete the proof.
\end{proof}

\begin{lemma}\label{th:lem_pseudo_bound_acc_strong_convex}
For constants $L,\mu,c_6>0$, a constant $c_5\in(0,1)$, a constant $Q_0\in (0,1)$, a integer $T\geq1$, matrices $\widetilde{\W}^{(0)},\W_*\in \mathcal{W} \subset \mathbb{R}^{d\times m}$, if it holds for a series of positive constants $\{\varepsilon_t\}$ that $\sum_{t=1}^{T}\sqrt{\varepsilon_tQ_0^{-t}} = O(\sqrt{c_6 {c_5}^{-T}}) $, setting $T = \Theta((K^2Lm/c_4)^{1/c_3})$,
we have
\begin{align*}
\mathcal{E} =&\frac{(Q_0)^T}{m}\biggl(K\sqrt{Lm} + 2\sqrt{\frac{L}{\mu}}\sum_{t=1}^{T}\sqrt{\varepsilon_t(Q_0)^{-t}} \\
&+ \sqrt{\sum_{t=1}^{T}\varepsilon_t(Q_0)^{-t}}\biggr)^2 \\
 = &O\biggl( K^2L\biggl[ \frac{c_6}{K^2\mu m}\biggr]^{\log_{c_5}Q_0}  \biggr).
\end{align*}
\end{lemma}
\begin{proof}
First, because $\varepsilon_t>0$ for $t\in[T]$, we have
\begin{align*}
 \sqrt{\sum_{t=1}^{T} \varepsilon_tQ_0^{-t}}\leq \sum_{t=1}^{T}\sqrt{ \varepsilon_tQ_0^{-t}}   .
\end{align*}
Then using Lemma \ref{th:lem_dist_W} and Lemma \ref{th:lem_sum_exp}, it suffices that
\begin{align*}
\mathcal{E}& = O\biggl(\frac{Q_0^T}{ m }\biggl[K\sqrt{Lm} + \sqrt{\frac{L}{\mu}} \sum_{t=1}^{T}\sqrt{\varepsilon_tQ_0^{-t}}   \biggr]^2   \biggr)\\
&=O\biggl( Q_0^T\biggl[K\sqrt{L}  + \sqrt{\frac{L}{m\mu}} \sum_{t=1}^{T}\sqrt{\varepsilon_tQ_0^{-t}}   \biggr]^2   \biggr)\\
&=O\biggl( Q_0^T\biggl[K\sqrt{L}  + \sqrt{\frac{L}{m\mu}} \sqrt{c_6 {c_5}^{-T}} \biggr]^2 \biggr).
\end{align*}
Then setting $T = \Theta(\log_{1/c_5}(K^2\mu m/c_6))$, we complete the proof.
\end{proof}

}

\section{Proof of Results in the Main Text}

\subsection{Proof of Claim \ref{th:lem_STL}}

\begin{proof}
Under the setting of single-task learning, each task is learned independently, and thus, using the notations in Definition \ref{th:df_MP-MTL-iter}, we have for $i=1,\ldots,m$, for any set $\calS$,
\begin{align*}
\bbP( &  \hat{\w}_{[-i]}^{(1:T)}  \in \calS  \mid \bigcap_{t=1}^T \calB_t = (\W^{(t-1)}, \calD^m, \ttheta_{1:t-1}) ) \\
&= \bbP(    \hat{\w}_{[-i]}^{(1:T)}  \in \calS  \mid \bigcap_{t=1}^T \calB_t = (\W_{[-i]}^{(t-1)}, \calD_{[-i]}, \ttheta_{1:t-1}) ).
\end{align*}

As such, we have for $i=1,\ldots,m$,
\begin{align*}
 &\frac{\bbP(  \hat{\w}_{[-i]}^{(1:T)}  \in \calS  \mid \bigcap_{t=1}^T \calB_t = (\W^{(t-1)}, \calD^m, \ttheta_{1:t-1}) )}{\bbP(  \hat{\w}_{[-i]}^{(1:T)} \in \calS \mid \bigcap_{t=1}^T \calB_t = ((\W')^{(t-1)}, (\calD')^m, \ttheta_{1:t-1}))} \\
 = & \frac{\bbP(    \hat{\w}_{[-i]}^{(1:T)}  \in \calS  \mid \bigcap_{t=1}^T \calB_t = (\W_{[-i]}^{(t-1)}, \calD_{[-i]}, \ttheta_{1:t-1}) )}{\bbP(    \hat{\w}_{[-i]}^{(1:T)}  \in \calS  \mid \bigcap_{t=1}^T \calB_t = (\W_{[-i]}^{(t-1)}, \calD_{[-i]}, \ttheta_{1:t-1}) )} = 1  \leq e^{0} .
\end{align*}
\end{proof}

{
\subsection{Proof of Proposition \ref{th:prop_better_than_STL}}\label{sec:proof_better_than_STL}
\begin{proof}
First, for Algorithm \ref{alg:MP-MTL-LR}, denoting $\SSigma_0 = \widetilde{\SSigma}^{(t)}$, the $j$-th diagonal element of $\S_{\eta\lambda}$ is
\begin{align*}
 &\max\biggl(0,1-\frac{\eta\lambda}{\sqrt{\sigma_j(\SSigma_0+\E)}}\biggr)\\
 \geq  &\max\biggl(0,1-\frac{\eta\lambda}{\sqrt{\sigma_j(\SSigma_0)+\sigma_d(\E)}}\biggr),
\end{align*}
where $\sigma_d(\E)$ is the $d$-th largest singular value, i.e., the smallest singular value, of $\E$. As such, when $\sigma_d(\E)= C \lambda^2$ for sufficiently large $C>0$, $\max\biggl(0,1-\frac{\eta\lambda}{\sqrt{\sigma_j(\SSigma_0)+\sigma_d(\E)}}\biggr) \rightarrow 1$.

Then $\hat{\w}_i^{(t-1)} = \U\S_{\eta\lambda}\U\trans\w_i^{(t-1)} = \U\U\trans\w_i^{(t-1)} = \w_i^{(t-1)} $. Therefore, all the procedures can be decoupled to independently run for each task, thus Algorithm \ref{alg:MP-MTL-LR} degrades to an STL algorithm with no random perturbation.

Similarly, for Algorithm \ref{alg:MP-MTL-GS}, for all $j\in[m]$, the $j$-th diagonal element of $\S_{\eta\lambda}$ is
\begin{align*}
 & \max\biggl(0,1-\frac{\eta\lambda}{\sqrt{|\SSigma_{jj,0}+\E_{jj}|}}\biggr)\\
 = & \max\biggl(0,1-\frac{\eta\lambda}{\sqrt{\SSigma_{jj,0}+\E_{jj}}}\biggr),
\end{align*}
where the equality is because $\SSigma_0$ is semi-positive definite and $\E$ is positive definite.

As such, when $\min_j\E_{jj}= C \lambda^2$ for sufficiently large $C>0$, $\min_j\biggl[\max\biggl(0,1-\frac{\eta\lambda}{\sqrt{\SSigma_{jj,0}+\E_{jj}}}\biggr)\biggr]\rightarrow 1$.

Then $\hat{\w}_i^{(t-1)} =  \S_{\eta\lambda} \w_i^{(t-1)} = \w_i^{(t-1)} $. Therefore, all the procedures can be decoupled to independently run for each task, thus Algorithm \ref{alg:MP-MTL-GS} degrades to an STL algorithm with no random perturbation.

\end{proof}

}

\subsection{Proof of Theorem \ref{th:th_MP-MTL}}\label{sec:proof_th_MP-MTL}
\begin{proof}
For simplicity, we omit the symbol $\calB$ used to denote the input in the conditional events in some equations.

First, we show that for all $t\in[T]$, the mapping $\W^{(t-1)}\rightarrow \SSigma^{(t)}$ is an $(\epsilon_t,0)$-differentially private algorithm.

\noindent\emph{\textbf{Case 1.}} For $\SSigma^{(t)} = \widetilde{\SSigma}^{(t)} + \E =  \widetilde{\W}^{(t-1)}(\widetilde{\W}^{(t-1)})\trans + \E$, we follow the proof of Theorem 4 of~\citet{jiang2016wishart}.

For all $i\in[m]$, consider two adjacent parameter matrices $\widetilde{\W}^{(t-1)}$ and $(\widetilde{\W}')^{(t-1)}$ that differ only in the $i$-th column such that $\widetilde{\W}^{(t-1)} = [\tilde{\w}_1^{(t-1)} \cdots \tilde{\w}_i^{(t-1)} \cdots \tilde{\w}_m^{(t-1)}]$ and $(\widetilde{\W}')^{(t-1)} = [\tilde{\w}_1^{(t-1)} \cdots (\tilde{\w}'_i)^{(t-1)} \cdots \tilde{\w}_m^{(t-1)}]$. Now, let
\begin{align*}
  \widetilde{\SSigma}^{(t)} &= \widetilde{\W}^{(t-1)}(\widetilde{\W}^{(t-1)})\trans = \sum_{j=1}^{m}\tilde{\w}_j^{(t-1)}(\tilde{\w}_j^{(t-1)})\trans\\
  (\widetilde{\SSigma}')^{(t)}& = (\widetilde{\W}')^{(t-1)}((\widetilde{\W}')^{(t-1)})\trans \\
  &= \sum_{j\in[m],j\neq i} \tilde{\w}_j^{(t-1)}(\tilde{\w}_j^{(t-1)})\trans + (\tilde{\w}'_i)^{(t-1)}((\tilde{\w}'_i)^{(t-1)})\trans\\
     \Delta &= \widetilde{\SSigma}^{(t)} - (\widetilde{\SSigma}')^{(t)} \\
    &= \tilde{\w}_i^{(t-1)}(\tilde{\w}_i^{(t-1)})\trans - (\tilde{\w}'_i)^{(t-1)}((\tilde{\w}'_i)^{(t-1)})\trans.
\end{align*}

Then, we have for the conditional densities
\begin{align*}
   \frac{p( \SSigma^{(t)} \mid \widetilde{\W}^{(t-1)})}{p( \SSigma^{(t)} \mid (\widetilde{\W}')^{(t-1)})}　
   = \frac{p( \SSigma^{(t)} = \widetilde{\W}^{(t-1)}(\widetilde{\W}^{(t-1)})\trans + \E_1)}{p( \SSigma^{(t)} = (\widetilde{\W}')^{(t-1)}((\widetilde{\W}')^{(t-1)})\trans+\E_2)}.
\end{align*}

Because $\E_1,\E_2\sim W_D(D+1,\frac{K^2}{2\epsilon_t}\I_D)$, letting $\V = \frac{K^2}{2\epsilon_t}\I_D$, $\alpha = \frac{K^2}{2\epsilon_t}$,
\begin{align*}
 &\frac{p( \SSigma^{(t)} = \widetilde{\W}^{(t-1)}(\widetilde{\W}^{(t-1)})\trans + \E_1)}{p( \SSigma^{(t)} = (\widetilde{\W}')^{(t-1)}((\widetilde{\W}')^{(t-1)})\trans+\E_2)}\\
 & = \frac{\exp[-\mbox{tr}(\V^{-1}( \SSigma^{(t)}- \widetilde{\W}^{(t-1)}(\widetilde{\W}^{(t-1)})\trans))/2]}{\exp[-\mbox{tr}(\V^{-1}(\SSigma^{(t)} -(\widetilde{\W}')^{(t-1)}((\widetilde{\W}')^{(t-1)})\trans))/2]}\\
 & = \exp[\mbox{tr}(\V^{-1}(\SSigma^{(t)} -(\widetilde{\W}')^{(t-1)}((\widetilde{\W}')^{(t-1)})\trans))/2\\
 &-\mbox{tr}(\V^{-1}( \SSigma^{(t)}- \widetilde{\W}^{(t-1)}(\widetilde{\W}^{(t-1)})\trans))/2]\\
 &=\exp[\mbox{tr}(\V^{-1}\Delta)/2]\\
 &=\exp[\mbox{tr}(  \tilde{\w}_i^{(t-1)}(\tilde{\w}_i^{(t-1)})\trans - (\tilde{\w}'_i)^{(t-1)}((\tilde{\w}'_i)^{(t-1)})\trans)/(2\alpha)]\\
 &=\exp[(\mbox{tr}(  \tilde{\w}_i^{(t-1)}(\tilde{\w}_i^{(t-1)})\trans) - \mbox{tr}((\tilde{\w}'_i)^{(t-1)}((\tilde{\w}'_i)^{(t-1)})\trans))/(2\alpha)]\\
 &=\exp[(\mbox{tr}(  (\tilde{\w}_i^{(t-1)})\trans\tilde{\w}_i^{(t-1)}) - \mbox{tr}(((\tilde{\w}'_i)^{(t-1)})\trans(\tilde{\w}'_i)^{(t-1)}))/(2\alpha)]\\
 &=\exp[(\|\tilde{\w}_i^{(t-1)}\|_2^2 - \|(\tilde{\w}'_i)^{(t-1)}\|_2^2)/(2\alpha)]\\
 &\leq \exp[\|\tilde{\w}_i^{(t-1)}\|_2^2/(2\alpha)]
 \leq \exp[K^2/(2\alpha)]=\exp(\epsilon_t).
\end{align*}
As such, we have
\begin{align*}
\frac{p( \SSigma^{(t)} \mid \widetilde{\W}^{(t-1)})}{p( \SSigma^{(t)} \mid (\widetilde{\W}')^{(t-1)})}　\leq \exp(\epsilon_t).
\end{align*}

\noindent\emph{\textbf{Case 2.}} Consider $\SSigma^{(t)} = \widetilde{\SSigma}^{(t)} + \E =  (\widetilde{\W}^{(t-1)})\trans\widetilde{\W}^{(t-1)} + \E$.

For all $i\in[m]$, consider two adjacent parameter matrices $\widetilde{\W}^{(t-1)}$ and $(\widetilde{\W}')^{(t-1)}$ that differ only in the $i$-th column such that $\widetilde{\W}^{(t-1)} = [\tilde{\w}_1^{(t-1)} \cdots \tilde{\w}_i^{(t-1)} \cdots \tilde{\w}_m^{(t-1)}]$ and $(\widetilde{\W}')^{(t-1)} = [\tilde{\w}_1^{(t-1)} \cdots (\tilde{\w}'_i)^{(t-1)} \cdots \tilde{\w}_m^{(t-1)}]$.
Let
\begin{align*}
  \widetilde{\SSigma}^{(t)} &= (\widetilde{\W}^{(t-1)})\trans\widetilde{\W}^{(t-1)}  \\
  (\widetilde{\SSigma}')^{(t)}& = ((\widetilde{\W}')^{(t-1)})\trans(\widetilde{\W}')^{(t-1)} \\
     \Delta &= \widetilde{\SSigma}^{(t)} - (\widetilde{\SSigma}')^{(t)},
\end{align*}
where the $i$-th diagonal element of $\Delta$ is $\|\tilde{\w}_i^{(t-1)}\|_2^2 - \|(\tilde{\w}'_i)^{(t-1)}\|_2^2$  and the other diagonal elements of $\Delta$ are zeros.

Then, we have
\begin{align*}
   \frac{p( \SSigma^{(t)} \mid \widetilde{\W}^{(t-1)})}{p( \SSigma^{(t)} \mid (\widetilde{\W}')^{(t-1)})}　
   = \frac{p( \SSigma^{(t)} = (\widetilde{\W}^{(t-1)})\trans\widetilde{\W}^{(t-1)} + \E_1)}{p( \SSigma^{(t)} = ((\widetilde{\W}')^{(t-1)})\trans(\widetilde{\W}')^{(t-1)}+\E_2)}.
\end{align*}

Because $\E_1,\E_2\sim W_m(m+1,\frac{K^2}{2\epsilon_t}\I_m)$, letting $\V = \frac{K^2}{2\epsilon_t}\I_m$,
\begin{align*}
 &\frac{p( \SSigma^{(t)} = (\widetilde{\W}^{(t-1)})\trans\widetilde{\W}^{(t-1)} + \E_1)}{p( \SSigma^{(t)} = ((\widetilde{\W}')^{(t-1)})\trans(\widetilde{\W}')^{(t-1)}+\E_2)}\\
    = &\frac{\exp[-\mbox{tr}(\V^{-1}( \SSigma^{(t)}- (\widetilde{\W}^{(t-1)})\trans\widetilde{\W}^{(t-1)}))/2]}{\exp[-\mbox{tr}(\V^{-1}(\SSigma^{(t)} -((\widetilde{\W}')^{(t-1)})\trans(\widetilde{\W}')^{(t-1)}))/2]}\\
     = &\exp[\mbox{tr}(\V^{-1}(\SSigma^{(t)} -((\widetilde{\W}')^{(t-1)})\trans(\widetilde{\W}')^{(t-1)}))/2\\
    &-\mbox{tr}(\V^{-1}( \SSigma^{(t)}- (\widetilde{\W}^{(t-1)})\trans\widetilde{\W}^{(t-1)}))/2]\\
    =&\exp[\mbox{tr}(\V^{-1}\Delta)/2] \\
    =&\exp[(\|\tilde{\w}_i^{(t-1)}\|_2^2 - \|(\tilde{\w}'_i)^{(t-1)}\|_2^2)/(2v_{ii})] \\
    \leq & \exp[\|\tilde{\w}_i^{(t-1)}\|_2^2/(2v_{ii})]
    \leq  \exp[K^2/(2v_{ii})]=  \exp(\epsilon_t).
\end{align*}
As such, we also have
\begin{align*}
\frac{p( \SSigma^{(t)} \mid \widetilde{\W}^{(t-1)})}{p( \SSigma^{(t)} \mid (\widetilde{\W}')^{(t-1)})}　\leq \exp(\epsilon_t).
\end{align*}

Because the norm clipping is a deterministic STL algorithm and because the mapping $\widetilde{\W}^{(t-1)} \rightarrow \SSigma^{(t)}$ is an $(\epsilon_t,0)$ - differentially private algorithm, we have for any set $\calS\subseteq \mathbb{R}^{d\times d}$ that
\begin{align*}
  \bbP( \SSigma&^{(t)}  \in \calS  \mid  \w_{[-i]}^{(t-1)}, \w_{i}^{(t-1)} ) \\
  = & \bbP( \SSigma^{(t)}  \in \calS  \mid  \tilde{\w}_{[-i]}^{(t-1)}, \tilde{\w}_{i}^{(t-1)} )\\
\leq   &e^{\epsilon_t} \bbP( \SSigma^{(t)}  \in \calS  \mid  \tilde{\w}_{[-i]}^{(t-1)}, (\tilde{\w}'_{i})^{(t-1)} )\\
=& e^{\epsilon_t} \bbP( \SSigma^{(t)}  \in \calS  \mid  \w_{[-i]}^{(t-1)},  ({\w}'_{i})^{(t-1)} ),
\end{align*}
which renders the mapping ${\W}^{(t-1)} \rightarrow \SSigma^{(t)}$ as an $(\epsilon_t,0)$ - differentially private algorithm as well.

Next, given $t\in[T]$, $\SSigma^{(1:t-1)}$ (when $t=1$, $\SSigma^{(1:t-1)}=\emptyset$) and the mapping $f:\SSigma^{(1:t)}\rightarrow \M^{(t)}$, which does not touch any unperturbed sensitive information, using the \emph{Post-Processing immunity} property (Property \ref{th:lem_post}) for the mapping $f': \SSigma^{(1:t)}\rightarrow (\M^{(t)},\SSigma^{(t)})$, the algorithm $(\W^{(t-1)},\SSigma^{(1:t-1)})\rightarrow (\M^{(t)},\SSigma^{(t)})$ is still an $(\epsilon_t,0)$-differentially private algorithm.

Then, because $\hat{\w}_i^{(t)}= \calA_{\mbox{st},i}(\M^{(t)}, \tilde{\w}_i^{(0:t-1)},\X_i,\y_i)$ is an STL algorithm for the $i$-th task, for $i = 1,\ldots,m$, the mapping $( \M^{(t)}, \tilde{\w}_{[-i]}^{(0:t-1)}, \X_{[-i]},\y_{[-i]} ) \rightarrow (\hat{\w}_{[-i]}^{(t)}  )$ thus does not touch any unperturbed sensitive information for the $i$-th task. As such,
applying the \emph{Post-Processing immunity} Lemma again for the mapping $f'': ( \M^{(t)}, \tilde{\w}_{[-i]}^{(0:t-1)}, \X_{[-i]},\y_{[-i]},\SSigma^{(1:t-1)}) \rightarrow (\hat{\w}_{[-i]}^{(t)}, \M^{(t)},\SSigma^{(t)})$,  for the algorithm $(\W^{(t-1)},\SSigma^{(1:t-1)},\w_{[-i]}^{(0:t-2)}, \X_{[-i]},\y_{[-i]})\rightarrow (\hat{\w}_{[-i]}^{(t)}, \M^{(t)},\SSigma^{(t)})$ (when $t=1$, $\w_{[-i]}^{(0:t-2)}=\emptyset$), denoting $\vartheta_{t,i} = (\hat{\w}_{[-i]}^{(t)}, \M^{(t)},\SSigma^{(t)})\in \calC_{t,i}$ , we have for any set $\calS_{t,i}\subseteq\calC_{t,i} $
\begin{align*}
 \bbP( \vartheta&_{t,i} \in \calS_{t,i}\mid \W^{(t-1)},\SSigma^{(1:t-1)}, \tilde{\w}_{[-i]}^{(0:t-2)}, \calD^m ) \\
 \leq &e^{\epsilon_t}\bbP( \vartheta_{t,i} \in \calS_{t,i}\mid (\W')^{(t-1)},\SSigma^{(1:t-1)},  \tilde{\w}_{[-i]}^{(0:t-2)}, (\calD')^m ),
\end{align*}
where $\W^{(t-1)}$ and $(\W')^{(t-1)}$ differ only in the $i$-th column and $\calD^m$ and $(\calD')^m$ differ only in the $i$-th task.

Now, again,  for $t=1,\ldots,T$, we take the $t$-th dataset $\widetilde{\calD}_t = \{(\w_1^{(t-1)}, \calD_1),\ldots,(\w_m^{(t-1)}, \calD_m)\}$. Given that $\W^{(t)} = \widehat{\W}^{(t)}$ for all $t\in[T]$, we have for any set $\calS_{t,i}\subseteq\calC_{t,i} $ that
\begin{align*}
 \bbP( \vartheta&_{t,i} \in \calS_{t,i}\mid \widetilde{\calD}_t, \vvartheta_{1:t-1} ) \\
 \leq &e^{\epsilon_t}\bbP( \vartheta_{t,i} \in \calS_{t,i}\mid \widetilde{\calD}'_t, \vvartheta_{1:t-1} ),
\end{align*}
where $\widetilde{\calD}_t$ and $\widetilde{\calD}'_t$ are two adjacent datasets that differ in a single entry, the $i$-th data instance $(\w_i^{(t-1)}, \calD_i = (\X_i,\y_i))$, and
\begin{numcases}{\vvartheta_{1:t-1} = }
\nonumber \emptyset, & $t=1$\\
\nonumber (\vartheta_{1,1},\dots,\vartheta_{1,m}) \ldots,(\vartheta_{t-1,1},\dots,\vartheta_{t-1,m}), & $t\geq2$.
\end{numcases}

This renders the algorithm in the $t$-th iteration an $(\epsilon_t,0)$-differentially private algorithm.

Now, again by the \emph{Adaptive composition} property (Property \ref{th:lem_Ad_comp}), for all $i\in[m]$ and for any set $\calS'\subseteq\bigotimes_{t=1}^{T}\mathcal{C}_{t_i}$, we have
\begin{align*}
  \mathbb{P} ((\vartheta&_{1,i},\cdots,\vartheta_{T,i})\in \mathcal{S}'  \mid \bigcap_{t=1}^T(\calB_t = (\widetilde{\calD}_t, \vvartheta_{1:t-1}))  )\\
  \leq & e^{\tilde{\epsilon}}\mathbb{P}((\vartheta_{1,i},\cdots,\vartheta_{T,i})\in \mathcal{S}'\mid \bigcap_{t=1}^T(\calB_t = (\widetilde{\calD}'_t, \vvartheta_{1:t-1}) ))\\
   &+ \delta,
\end{align*}
where for all $t\in[T]$, $\calB_t$ denotes the input for the $t$-th iteration.

Finally, taking $\theta_t = (\vartheta_{t,1},\dots,\vartheta_{t,m})$ for all $t\in[T]$, we have for any set $\calS\subseteq \mathbb{R}^{d\times(m-1)\times T}$
\begin{align*}
 \bbP( &  \hat{\w}_{[-i]}^{(1:T)}  \in \calS  \mid \bigcap_{t=1}^T \calB_t = (\W^{(t-1)}, \calD^m, \ttheta_{1:t-1}) ) \\
\leq   &e^{\epsilon} \bbP(  \hat{\w}_{[-i]}^{(1:T)} \in \calS \mid \bigcap_{t=1}^T \calB_t = ((\W')^{(t-1)}, (\calD')^m, \ttheta_{1:t-1})) \\
&+ \delta,
\end{align*}
\end{proof}

\subsection{Proof of Corollary \ref{th:th_MP-MTL-LR}}\label{sec:proof_MP-MTL-LR}
\begin{proof}
For simplicity, we omit the symbol $\calB$ used to denote the input in the conditional events in some equations.

Using Theorem \ref{th:th_MP-MTL}, we only need to show that Algorithm \ref{alg:MP-MTL-LR} complies with our MP-MTL framework in Algorithm \ref{alg:MP-MTL}.



Let $\M^{(t)} = \U\S_{\eta\lambda}\U\trans$. As such, the \ref{eq:trans_LR}-th step to the \ref{eq:feature_tran-LR}-th step can be treated as the process of first performing a mapping $f:\SSigma^{(1:t)}\rightarrow \M^{(t)}$ and then applying an STL algorithm:
\begin{equation}\label{eq:STL1-LR}
  \hat{\w}_i^{(t)} = \U\S_{\eta\lambda}\U\trans\tilde{\w}_i^{(t-1)}, \ \mbox{for all} \ i\in[m].
\end{equation}

Now, because \eqref{eq:STL1-LR}, the \ref{eq:acc-LR}-th step and the \ref{eq:gd-LR}-th step are all STL algorithms, they can be treated as a entire STL algorithm performing the mapping: $(\M^{(t)}, \tilde{\w}_i^{(0:t-1)},\X_i,\y_i)\rightarrow  (\hat{\w}_i^{(t)},\tilde{\w}_i^{(t)})$.

As such, in all the iterations, Algorithm \ref{alg:MP-MTL-LR} complies with Algorithm \ref{alg:MP-MTL}. Thus, the result of Theorem \ref{th:th_MP-MTL} can be applied to Algorithm \ref{alg:MP-MTL-LR}.

Similarly, using Theorem \ref{th:th_MP-MTL}, we only need to show that Algorithm \ref{alg:MP-MTL-GS} complies with our MP-MTL framework in Algorithm \ref{alg:MP-MTL}.

The proof for the sensitivity is the same.

Now, let $\M^{(t)} = \S_{\eta\lambda}$. As such, the \ref{eq:construct_Mt}-th step can be treated as a mapping $f:\SSigma^{(1:t)}\rightarrow \M^{(t)}$.

Then, because the \ref{eq:feature_select}-th step, the \ref{eq:acc_select}-th step and the \ref{eq:gd}-th step are all STL algorithms, they can be treated as a entire STL algorithm performing the mapping: $(\M^{(t)}, \tilde{\w}_i^{(0:t-1)},\X_i,\y_i)\rightarrow (\hat{\w}_i^{(t)},\tilde{\w}_i^{(t)})$.

Therefore, in all the iterations, Algorithm \ref{alg:MP-MTL-GS} complies with Algorithm \ref{alg:MP-MTL}, and thus, the result of Theorem \ref{th:th_MP-MTL} can be applied to Algorithm \ref{alg:MP-MTL-GS}.

\end{proof}

{
\subsection{Proof of Lemma \ref{th:lem_LR_iter_t}}
\begin{proof}
We invoke the results of \citet{schmidt2011convergence} to bound the empirical optimization error.

In the $t$-th step, a standard proximal operator (see \citet{ji2009accelerated}) optimizes the following problem:
\begin{align*}
  \min_{\W} \frac{1}{2\eta}\|\W - \C\|_F^2 + \lambda\|\W\|_*,
\end{align*}
where $\C = \widetilde{\W}^{(t-1)}$. By Theorem 3.1 of \citet{ji2009accelerated}, denote the solution of the problem by $\widehat{\W}_0^{(t)} = \U_0\S_{\eta\lambda,0}\U_0\trans\C$. Let $\U_0\LLambda_0\U_0\trans = \C\C\trans$ be the SVD decomposition of $\C\C\trans$. $\S_{\eta\lambda,0}$ is also a diagonal matrix and $\S_{\eta\lambda,ii,0} = \max\{0, 1 - \eta\lambda/\sqrt{\LLambda_{ii,0}}\}$ for $i=1,\ldots,\min\{d,m\}$.

By Algorithm \ref{alg:MP-MTL-LR}, $\widehat{\W}^{(t)} = \U\S_{\eta\lambda}\U\trans\C$.

Then we analyse the bound of $\frac{1}{2\eta}\|\widehat{\W}^{(t)} - \C\|_F^2 + \lambda\|\widehat{\W}^{(t)}\|_*  \\
  - \{\frac{1}{2\eta}\|\widehat{\W}_0^{(t)} - \C\|_F^2 + \lambda\|\widehat{\W}_0^{(t)}\|_*\}.$

First, we have
\begin{equation}\label{eq:part_1}
\begin{split}
 \|&\widehat{\W}^{(t)} - \C\|_F^2 - \|\widehat{\W}_0^{(t)} - \C\|_F^2 \\
= &\mbox{tr}((\widehat{\W}^{(t)} - \C)\trans(\widehat{\W}^{(t)} - \C)) -  \mbox{tr}((\widehat{\W}_0^{(t)} - \C)\trans(\widehat{\W}_0^{(t)} - \C))\\
  =& \mbox{tr}((\widehat{\W}^{(t)})\trans\widehat{\W}^{(t)}) -  \mbox{tr}((\widehat{\W}_0^{(t)})\trans\widehat{\W}_0^{(t)})
 - 2\mbox{tr}((\widehat{\W}^{(t)}  - \widehat{\W}^{(t)}_0)\trans\C)\\
 =& \mbox{tr}((\widehat{\W}^{(t)}-\widehat{\W}_0^{(t)})\trans(\widehat{\W}^{(t)}+\widehat{\W}_0^{(t)}))
 - 2\mbox{tr}((\widehat{\W}^{(t)}  - \widehat{\W}^{(t)}_0)\trans\C)\\
 =& \mbox{tr}((\widehat{\W}^{(t)}-\C)\trans(\widehat{\W}^{(t)}-\widehat{\W}_0^{(t)}))\\
 &+ \mbox{tr}((\widehat{\W}_0^{(t)}-\C)\trans(\widehat{\W}^{(t)}-\widehat{\W}_0^{(t)}))\\
 \leq& \sigma_1(\widehat{\W}^{(t)}-\C)\|\widehat{\W}^{(t)}-\widehat{\W}_0^{(t)}\|_*\\
 &+\sigma_1(\widehat{\W}_0^{(t)}-\C)\|\widehat{\W}^{(t)}-\widehat{\W}_0^{(t)}\|_*,
\end{split}
\end{equation}
%
where $\sigma_1(\cdot)$ denotes the largest singular value of the enclosed matrix.

Denote $\T = \U\S_{\eta\lambda}\U\trans, \T_0 = \U_0\S_{\eta\lambda,0}\U_0\trans$. Since $\U$ is decomposed from a symmetric matrix, we have
\begin{align*}
\sigma_1(\widehat{\W}^{(t)}-\C) & = \sigma_1(\T\C-\C)\leq \sigma_1(\C)\sigma_1(\T - \I) \\
&= \sigma_1(\C)\sigma_1(\U\S_{\eta\lambda}\U\trans - \U\U\trans) \\
&= \sigma_1(\C)\sigma_1(\U(\S_{\eta\lambda}-\I)\U\trans).
\end{align*}
Since $\S_{\eta\lambda}-\I$ is a diagonal matrix, whose $i$-th diagonal element is $\max\{0,1 - {\eta\lambda}/{\sqrt{\LLambda_{ii}}}\}-1\in[-1,0)$, so $\sigma_1(\U(\S_{\eta\lambda}-\I)\U\trans)\leq 1$ and
\begin{equation}\label{eq:sigma_1_W_C}
  \sigma_1(\widehat{\W}^{(t)}-\C) \leq \sigma_1(\C).
\end{equation}
Similarly, 
\begin{equation}\label{eq:sigma_1_W_0_C}
\sigma_1(\widehat{\W}_0^{(t)}-\C) \leq \sigma_1(\C).
\end{equation}

On the other hand,
\begin{equation}\label{eq:W_W_0}
\begin{split}
 &\|\widehat{\W}^{(t)}-\widehat{\W}_0^{(t)}\|_*=\|\T\C - \T_0\C\|_* \\
=& \biggl\|\sum_{j=1}^{d}\sigma_j(\T)\u_j\u_j\trans\C - \sum_{j=1}^{d}\sigma_j(\T_0)\u_{j,0}\u_{j,0}\trans\C \biggr\|_*\\
=& \biggl\|\sum_{j=1}^{d}(\sigma_j(\T_0)+\sigma_j(\T)-\sigma_j(\T_0))\u_j\u_j\trans\C \\
&- \sum_{j=1}^{d}\sigma_j(\T_0)\u_{j,0}\u_{j,0}\trans\C \biggr\|_*\\
=& \biggl\|\sum_{j=1}^{d}\sigma_j(\T_0)(\u_j\u_j\trans-\u_{j,0}\u_{j,0}\trans )\C\\
&+\sum_{j=1}^{d}(\sigma_j(\T)-\sigma_j(\T_0))\u_j\u_j\trans\C \biggr\|_*\\
\leq& \biggl\|\sum_{j=1}^{d}\sigma_j(\T_0)(\u_j\u_j\trans-\u_{j,0}\u_{j,0}\trans )\C \biggr\|_*\\
&+\biggl\|\sum_{j=1}^{d}(\sigma_j(\T)-\sigma_j(\T_0))\u_j\u_j\trans\C \biggr\|_*,
\end{split}
\end{equation}
where $\u_j$ and $\u_{j,0}$ are the $j$-th column of $\U$ and $\U_0$, respectively.

Let $r_c = \mbox{rank}(\C)\leq\min\{d,m\}$ be the rank of $\C$.   Then we have
\begin{equation}\label{eq:part_1_1_head}
\begin{split}
&\biggl\|\sum_{j=1}^{d}(\sigma_j(\T)-\sigma_j(\T_0))\u_j\u_j\trans\C \biggr\|_*\\
  \leq& \sum_{j=1}^{r_c}|\sigma_j(\T)-\sigma_j(\T_0)|\sigma_j(\C)\leq \sigma_1(\C)\sum_{j=1}^{r_c}|\sigma_j(\T)-\sigma_j(\T_0)|.
\end{split}
\end{equation}
%

Denote $\SSigma_0 = \widetilde{\SSigma}^{(t)} = \C\C\trans$. Then we have for $j\in[r_c]$,
\begin{align*}
&|\sigma_j(\T)-\sigma_j(\T_0)|\\
 =&\biggl|\max\biggl(0,1-\frac{\eta\lambda}{\sqrt{\sigma_j(\SSigma_0+\E)}}\biggr)
  - \max\biggl(0,1-\frac{\eta\lambda}{\sqrt{\sigma_j(\SSigma_0)}}\biggr)  \biggr|\\
 \leq& \biggl|\max\biggl(0,1-\frac{\eta\lambda}{\sqrt{\sigma_j(\SSigma_0)+\sigma_1(\E)}}\biggr)
  - \max\biggl(0,1-\frac{\eta\lambda}{\sqrt{\sigma_j(\SSigma_0)}}\biggr)  \biggr|.
\end{align*}

\noindent\textbf{Case 1:} $\eta\lambda>\sqrt{\sigma_j(\SSigma_0)}$. Then
\begin{align*}
&|\sigma_j(\T)-\sigma_j(\T_0)|\\
=&\max\biggl(0,1-\frac{\eta\lambda}{\sqrt{\sigma_j(\SSigma_0)+\sigma_1(\E)}}\biggr)
\leq1-\frac{\eta\lambda}{ \sqrt{\eta^2\lambda^2+\sigma_1(\E)}}\\
\leq&1-\frac{\eta\lambda}{\eta\lambda+\sqrt{\sigma_1(\E)}} = \frac{\sqrt{\sigma_1(\E)}}{\eta\lambda+ \sqrt{\sigma_1(\E)}}
\leq  \frac{ \sqrt{\sigma_1(\E)}}{\eta\lambda}
\end{align*}
\noindent\textbf{Case 2:} $\eta\lambda\leq\sqrt{\sigma_j(\SSigma_0)}$. Then
\begin{align*}
&|\sigma_j(\T)-\sigma_j(\T_0)|\\
=&1-\frac{\eta\lambda}{\sqrt{\sigma_j(\SSigma_0)+\sigma_1(\E)}} - 1+\frac{\eta\lambda}{\sqrt{\sigma_j(\SSigma_0)}}\\
=&\eta\lambda\cdot\frac{\sqrt{\sigma_j(\SSigma_0)+\sigma_1(\E)} - \sqrt{\sigma_j(\SSigma_0)}}{\sqrt{\sigma_j^2(\SSigma_0)+\sigma_j(\SSigma_0)\sigma_1(\E)}}\\
=& \frac{\eta\lambda\sigma_1(\E)}{[\sqrt{\sigma_j(\SSigma_0)+\sigma_1(\E)} + \sqrt{\sigma_j(\SSigma_0)}]\sqrt{\sigma_j^2(\SSigma_0)+\sigma_j(\SSigma_0)\sigma_1(\E)}}\\
\leq&\frac{\eta\lambda\sigma_1(\E)}{[\sqrt{\eta^2\lambda^2+0} + \sqrt{\eta^2\lambda^2}]\sqrt{\eta^4\lambda^4+0}}
=\frac{\sigma_1(\E)}{2\eta^2\lambda^2}.
\end{align*}

Suppose that there exists an index $k\leq d$ such that
\begin{align*}
\sigma_{k}^2(\C) = \sigma_{k}(\SSigma_0) > \eta^2\lambda^2, \sigma_{k+1}^2(\C) =\sigma_{k+1}(\SSigma_0) \leq \eta^2\lambda^2,
\end{align*}
then $\sigma_j(\T_0)>0$ for $j\leq k$, $k\leq r_c$, and
\begin{equation}\label{eq:part_1_1_tail}
\begin{split}
&\sum_{j=1}^{r_c}|\sigma_j(\T)-\sigma_j(\T_0)| \\
\leq &k\frac{\sigma_1(\E)}{2\eta^2\lambda^2}  + (r_c-k)I(r_c>k)\frac{ \sqrt{\sigma_1(\E)}}{\eta\lambda}.
\end{split}
\end{equation}
%

For another part of \eqref{eq:W_W_0},
\begin{equation}\label{eq:part_1_2_head}
\begin{split}
&\biggl\|\sum_{j=1}^{d}\sigma_j(\T_0)(\u_j\u_j\trans-\u_{j,0}\u_{j,0}\trans )\C \biggr\|_* \\
= &\biggl\|\sum_{j=1}^{k}\sigma_j(\T_0)(\u_j\u_j\trans-\u_{j,0}\u_{j,0}\trans )\C \biggr\|_* \\
\leq & \sigma_1(\C)\biggl\|\sum_{j=1}^{k}\sigma_j(\T_0)(\u_j\u_j\trans-\u_{j,0}\u_{j,0}\trans )  \biggr\|_*.
\end{split}
\end{equation}
%

Denote $\U_j = \sum_{j'=1}^{j}\u_{j'}\u_{j'},\U_{j,0} = \sum_{j'=1}^{j}\u_{j',0}\u_{j',0}$ for $j \in [d]$.
Let $\U_0 = \U_{0,0} = \textbf{0}$. Then $\u_j\u_j = \U_j - \U_{j-1}, \u_{j,0}\u_{j,0} = \U_{j,0} - \U_{j-1,0}$ for $j\in[d]$.

Then we have
\begin{align*}
&\biggl\|\sum_{j=1}^{k}\sigma_j(\T_0)(\u_j\u_j\trans-\u_{j,0}\u_{j,0}\trans )  \biggr\|_*\\
=&\biggl\|\sum_{j=1}^{k}\sigma_j(\T_0)(\U_j - \U_{j,0} -(\U_{j-1} - \U_{j-1,0}))  \biggr\|_*\\
=&\biggl\|\sum_{j=1}^{k-1}(\sigma_j(\T_0)-\sigma_{j+1}(\T_0))(\U_j - \U_{j,0})\\
& + \sigma_k(\T_0)(\U_k - \U_{k,0})    \biggr\|_*\\
\leq &\sum_{j=1}^{k-1}(\sigma_j(\T_0)-\sigma_{j+1}(\T_0))\|\U_j - \U_{j,0}\|_*\\
& + \sigma_k(\T_0)\|\U_k - \U_{k,0}\|_*.
\end{align*}

We assume $2\sigma_1(\E)\leq \sigma_j(\SSigma_0)-\sigma_{j+1}(\SSigma_0)$ for all $j \in [k]$, and apply the Theorem 6 of \citet{jiang2016wishart}. Then for $j \in [k]$,
\begin{align*}
\|\U_j - \U_{j,0}\|_* \leq& \min\{2j,k\}\|\U_j - \U_{j,0}\|_2 \\
\leq& \min\{2j,k\}\frac{2\sigma_1(\E)}{\sigma_j(\SSigma_0)-\sigma_{j+1}(\SSigma_0)}.
\end{align*}

Since $j \in [k-1]$,
\begin{align*}
\sigma_j(\T_0)-\sigma_{j+1}(\T_0) =&  1-\frac{\eta\lambda}{\sqrt{\sigma_j(\SSigma_0) }}
  -  \biggl(1-\frac{\eta\lambda}{\sqrt{\sigma_{j+1}(\SSigma_0)}}  \biggr) \\
  = &  \eta\lambda \frac{\sqrt{\sigma_{j}(\SSigma_0)} -\sqrt{\sigma_{j+1}(\SSigma_0)} }{\sqrt{\sigma_j(\SSigma_0)\sigma_{j+1}(\SSigma_0) }},
\end{align*}
and
\begin{align*}
\sigma_k(\T_0) =&  1-\frac{\eta\lambda}{\sqrt{\sigma_k(\SSigma_0) }}
  \leq   1- \frac{\sqrt{\sigma_{k+1}(\SSigma_0) }}{\sqrt{\sigma_k(\SSigma_0) }},
\end{align*}
therefore,
\begin{equation}\label{eq:part_1_2_tail}
\begin{split}
&\biggl\|\sum_{j=1}^{k}\sigma_j(\T_0)(\u_j\u_j\trans-\u_{j,0}\u_{j,0}\trans )  \biggr\|_*\\
\leq &\sum_{j=1}^{k-1}\frac{2\eta\lambda\min\{2j,k\}\sigma_1(\E)}{
(\sqrt{\sigma_j(\SSigma_0)}+\sqrt{\sigma_{j+1}(\SSigma_0)})\sqrt{\sigma_j(\SSigma_0)\sigma_{j+1}(\SSigma_0) }}\\
& + \frac{2\eta\lambda k\sigma_1(\E)}{
(\sqrt{\sigma_k(\SSigma_0)}+\sqrt{\sigma_{k+1}(\SSigma_0)})\sqrt{\sigma_k(\SSigma_0)}}\\
\leq &\sum_{j=1}^{k-1}\frac{2\eta\lambda\min\{2j,k\}\sigma_1(\E)}{
(\eta\lambda+\eta\lambda)\sqrt{\eta^2\lambda^2\eta^2\lambda^2}}
+ \frac{2\eta\lambda k\sigma_1(\E)}{
(\eta\lambda+0)\eta\lambda}\\
\leq & \biggl(\frac{k(k-1)}{
 \eta^2\lambda^2}
+ \frac{2  k }{
 \eta\lambda}\biggr) \sigma_1(\E).
\end{split}
\end{equation}

Combining \eqref{eq:part_1}, \eqref{eq:sigma_1_W_C}, \eqref{eq:sigma_1_W_0_C}, \eqref{eq:W_W_0}, \eqref{eq:part_1_1_head}, \eqref{eq:part_1_1_tail}, \eqref{eq:part_1_2_head} and \eqref{eq:part_1_2_tail}, it follows that
\begin{equation}\label{eq:part_1_final}
\begin{split}
 \|&\widehat{\W}^{(t)} - \C\|_F^2 - \|\widehat{\W}_0^{(t)} - \C\|_F^2 \\
 \leq &2\biggl[ k\frac{\sigma_1(\E)}{2\eta^2\lambda^2}  + (r_c-k)I(r_c>k)\frac{ \sqrt{\sigma_1(\E)}}{\eta\lambda} \\
 &+ \biggl(\frac{k(k-1)}{
 \eta^2\lambda^2}
+ \frac{2  k }{
 \eta\lambda}\biggr) \sigma_1(\E)\biggr]\sigma_1^2(\C).
\end{split}
\end{equation}

On the other hand,
\begin{align*}
  \|\widehat{\W}^{(t)}\|_* -\|\widehat{\W}_0^{(t)}\|_*\leq \|\widehat{\W}^{(t)}-\widehat{\W}_0^{(t)}\|_*.
\end{align*}

As such, we have
\begin{equation}\label{eq:mid_bound_final}
\begin{split}
&\frac{1}{2\eta}\|\widehat{\W}^{(t)} - \C\|_F^2 + \lambda\|\widehat{\W}^{(t)}\|_*  \\
&  - \biggl\{\min_{\W} \frac{1}{2\eta}\|\W - \C\|_F^2 + \lambda\|\W\|_*\biggr\}\\
=&\frac{1}{2\eta}( \|\widehat{\W}^{(t)} - \C\|_F^2 - \|\widehat{\W}_0^{(t)} - \C\|_F^2 )\\
 &+ \lambda(\|\widehat{\W}^{(t)}\|_* -\|\widehat{\W}_0^{(t)}\|_* )\\
 \leq &  \biggl(\frac{\sigma_1^2(\C)}{\eta}+\lambda\sigma_1(\C)\biggl)\biggl[ k\frac{\sigma_1(\E)}{2\eta^2\lambda^2} \\
 & + (r_c-k)I(r_c>k)\frac{ \sqrt{\sigma_1(\E)}}{\eta\lambda}
 + \biggl(\frac{k(k-1)}{
 \eta^2\lambda^2}
+ \frac{2  k }{
 \eta\lambda}\biggr) \sigma_1(\E)\biggr]\\
 =&\frac{1}{\eta}\biggl(\frac{\sigma_1^2(\C)}{\eta\lambda}+ \sigma_1(\C)\biggl)\biggl[ k\frac{\sigma_1(\E)}{2\eta\lambda} \\
 & + \max(0,r_c-k)\sqrt{\sigma_1(\E)}
 + \biggl(\frac{k(k-1)}{
 \eta\lambda}
+ 2k \biggr) \sigma_1(\E)\biggr].
\end{split}
\end{equation}

\end{proof}

\subsection{Proof of Lemma \ref{th:lem_GS_iter_t}}
\begin{proof}

In the $t$-th step, a standard proximal operator (see \citet{liu2009multi}) optimizes the following problem:
\begin{align*}
  \min_{\W} \frac{1}{2\eta}\|\W - \C\|_F^2 + \lambda\|\W\|_{2,1},
\end{align*}
where $\C = \widetilde{\W}_i^{(t-1)}$. By Theorem 5 of \citet{liu2009multi}, denote the solution of the problem by $\widehat{\W}_0^{(t)} = \S_{\eta\lambda,0}\C$. Let $\LLambda_0 $ be a diagonal matrix containing the diagonal elements of $\C\C\trans$, and let $\S_0$ be a diagonal matrix and suffices $\S_{ii,0} = \sqrt{\LLambda_{ii,0}}$ for $i=1,\ldots,\min\{d,m\}$. $\S_{\eta\lambda,0}$ is also a diagonal matrix and $\S_{\eta\lambda,ii,0} = \max\{0, 1- \eta\lambda/{\S}_{ii,0} \}$ for $i=1,\ldots,\min\{d,m\}$.

By Algorithm \ref{alg:MP-MTL-LR}, $\widehat{\W}^{(t)} = \U\S_{\eta\lambda}\U\trans\C$.

Then we analyse the bound of $\frac{1}{2\eta}\|\widehat{\W}^{(t)} - \C\|_F^2 + \lambda\|\widehat{\W}^{(t)}\|_{2,1}  \\
  - \{\frac{1}{2\eta}\|\widehat{\W}_0^{(t)} - \C\|_F^2 + \lambda\|\widehat{\W}_0^{(t)}\|_{2,1}\}.$

First, similarly as in \eqref{eq:part_1}, we have
\begin{equation}\label{eq:part_1_GS}
\begin{split}
 &\|\widehat{\W}^{(t)} - \C\|_F^2 - \|\widehat{\W}_0^{(t)} - \C\|_F^2 \\
 =& \mbox{tr}((\widehat{\W}^{(t)}-\widehat{\W}_0^{(t)})(\widehat{\W}^{(t)}-\C)\trans)\\
 &+ \mbox{tr}((\widehat{\W}^{(t)}-\widehat{\W}_0^{(t)})(\widehat{\W}_0^{(t)}-\C)\trans)\\
 =& \sum_{j=1}^{d}(\widehat{\W}^{(t)}-\widehat{\W}_0^{(t)})^j((\widehat{\W}^{(t)}-\C)^j)\trans\\
 &+\sum_{j=1}^{d}(\widehat{\W}^{(t)}-\widehat{\W}_0^{(t)})^j((\widehat{\W}_0^{(t)}-\C)^j)\trans\\
 \leq& \|\widehat{\W}^{(t)}-\C\|_{2,1}\|\widehat{\W}^{(t)}-\widehat{\W}_0^{(t)}\|_{2,1}\\
 &+\|\widehat{\W}_0^{(t)}-\C\|_{2,1}\|\widehat{\W}^{(t)}-\widehat{\W}_0^{(t)}\|_{2,1},
\end{split}
\end{equation}
where $(\cdot)^j$ denotes the $j$-th row vector of the enclosed matrix.

Denote $\T = \S_{\eta\lambda}, \T_0 = \S_{\eta\lambda,0}$.
Denote the indices of non-zero rows of $\C$ by $\mathcal{I}_c = \{j:\C^j\neq \mathbf{0}\}$ and let $r_{c,s} = |\mathcal{I}_c| \leq d$.

We have
\begin{align*}
& \|\widehat{\W}^{(t)}-\C\|_{2,1} = \|(\T-\I)\C\|_{2,1}\\
&=\sum_{j=1}^{d}\sqrt{\sum_{i=1}^{m}|(\T-\I)^j\C_{i}|^2}
= \sum_{j\in \mathcal{I}_c }\sqrt{\sum_{i=1}^{m}|(\T-\I)_{jj}\C_{ij}|^2}\\
&= \sum_{j\in \mathcal{I}_c}\sqrt{\sum_{i=1}^{m}|(\T-\I)_{jj}|^2|\C_{ij}|^2}= \sum_{j\in \mathcal{I}_c}|(\T-\I)_{jj}|\|\C^j\|_{2}.
\end{align*}

Since $\S_{\eta\lambda}-\I$ is a diagonal matrix, whose $i$-th diagonal element is $\max\{0,1 - {\eta\lambda}/{\S_{ii}}\}-1\in[-1,0)$, so
\begin{equation}\label{eq:sigma_1_W_C_GS}
  \|\widehat{\W}^{(t)}-\C\|_{2,1}\leq  \sum_{j\in \mathcal{I}_c}\|\C^j\|_{2}\leq r_{c,s}\max_{j\in[d]}\|\C^j\|_{2}.
\end{equation}
Similarly, 
\begin{equation}\label{eq:sigma_1_W_0_C_GS}
\|\widehat{\W}_0^{(t)}-\C\|_{2,1}\leq   \sum_{j\in \mathcal{I}_c}\|\C^j\|_{2}\leq r_{c,s}\max_{j\in[d]}\|\C^j\|_{2}.
\end{equation}

On the other hand,
\begin{equation}\label{eq:W_W_0_GS}
\begin{split}
 &\|\widehat{\W}^{(t)}-\widehat{\W}_0^{(t)}\|_{2,1}  = \|\S_{\eta\lambda}\C - \S_{\eta\lambda,0}\C\|_{2,1}  \\
=& \sum_{j\in \mathcal{I}_c}|\S_{\eta\lambda,jj}-\S_{\eta\lambda,jj,0}|\|\C^j\|_{2}\\
\leq& \max_{j'\in[d]}\|\C^{j'}\|_{2}\sum_{j\in \mathcal{I}_c}|\S_{\eta\lambda,jj}-\S_{\eta\lambda,jj,0}|.
\end{split}
\end{equation}

Denote $\SSigma_0 = \widetilde{\SSigma}^{(t)} = \C\C\trans$. Then we have for $j\in \mathcal{I}_c$,
\begin{align*}
&|\S_{\eta\lambda,jj}-\S_{\eta\lambda,jj,0}|\\
 =&\biggl|\max\biggl(0,1-\frac{\eta\lambda}{\sqrt{|\SSigma_{jj,0}+\E_{jj}|}}\biggr)
  - \max\biggl(0,1-\frac{\eta\lambda}{\sqrt{|\SSigma_{jj,0}|}}\biggr)  \biggr|.
\end{align*}

\noindent\textbf{Case 1:} $\eta\lambda>\sqrt{\SSigma_{jj,0}}$. Then
\begin{align*}
&|\S_{\eta\lambda,jj}-\S_{\eta\lambda,jj,0}|\\
=&\max\biggl(0,1-\frac{\eta\lambda}{\sqrt{|\SSigma_{jj,0}+\E_{jj}|}}\biggr)
\leq1-\frac{\eta\lambda}{ \sqrt{\eta^2\lambda^2+|\E_{jj}|}}\\
\leq&1-\frac{\eta\lambda}{\eta\lambda+\sqrt{|\E_{jj}|}} = \frac{\sqrt{|\E_{jj}|}}{\eta\lambda+ \sqrt{|\E_{jj}|}}
\leq  \frac{ \sqrt{|\E_{jj}|}}{\eta\lambda}
\end{align*}
\noindent\textbf{Case 2:} $\eta\lambda\leq\sqrt{\SSigma_{jj,0}}$. Then
\begin{align*}
&|\S_{\eta\lambda,jj}-\S_{\eta\lambda,jj,0}|\\
\leq&1-\frac{\eta\lambda}{\sqrt{|\SSigma_{jj,0}|+|\E_{jj}|}} - 1+\frac{\eta\lambda}{\sqrt{|\SSigma_{jj,0}|}}\\
=&\eta\lambda\cdot\frac{\sqrt{|\SSigma_{jj,0}|+|\E_{jj}|} - \sqrt{|\SSigma_{jj,0}|}}{\sqrt{|\SSigma_{jj,0}|(|\SSigma_{jj,0}|+|\E_{jj}|)}}\\
=& \frac{\eta\lambda|\E_{jj}|}{[\sqrt{|\SSigma_{jj,0}|+|\E_{jj}|} + \sqrt{|\SSigma_{jj,0}|}]\sqrt{|\SSigma_{jj,0}|^2+|\E_{jj}||\SSigma_{jj,0}|}}\\
\leq&\frac{\eta\lambda|\E_{jj}|}{[\sqrt{\eta^2\lambda^2+0} + \sqrt{\eta^2\lambda^2}]\sqrt{\eta^4\lambda^4+0}}
=\frac{|\E_{jj}|}{2\eta^2\lambda^2}.
\end{align*}

Suppose that there exists an integer $k\leq d$ such that
\begin{align*}
\sum_{j=1}^{d}I(\sqrt{\SSigma_{jj,0}}\geq \eta\lambda) = k
\end{align*}
then $k\leq r_{c,s}$ and
\begin{equation}\label{eq:part_1_1_tail_GS}
\begin{split}
&\sum_{j\in \mathcal{I}_c}|\S_{\eta\lambda,jj}-\S_{\eta\lambda,jj,0}| \\
\leq &\frac{k}{2\eta^2\lambda^2} \max_{j:\eta^2\lambda^2\leq\SSigma_{jj,0}}\E_{jj}\\
& + \frac{ (r_{c,s}-k)I(r_{c,s}>k)}{\eta\lambda} \max_{j:\eta^2\lambda^2>\SSigma_{jj,0}}\sqrt{\E_{jj}}.
\end{split}
\end{equation}

Combining \eqref{eq:part_1_GS}, \eqref{eq:sigma_1_W_C_GS}, \eqref{eq:sigma_1_W_0_C_GS}, \eqref{eq:W_W_0_GS} and \eqref{eq:part_1_1_tail_GS}, it follows that
\begin{equation}\label{eq:part_1_final_GS}
\begin{split}
 \|&\widehat{\W}^{(t)} - \C\|_F^2 - \|\widehat{\W}_0^{(t)} - \C\|_F^2 \\
 \leq &2r_{c,s}\biggl(\max_{j\in[d]}\|\C^j\|_{2}\biggr)^2 \biggl(\frac{k}{2\eta^2\lambda^2} \max_{j:\eta^2\lambda^2\leq\SSigma_{jj,0}}|\E_{jj}|\\
& + \frac{ (r_{c,s}-k)I(r_{c,s}>k)}{\eta\lambda} \max_{j:\eta^2\lambda^2>\SSigma_{jj,0}}\sqrt{|\E_{jj}|}\biggr).
\end{split}
\end{equation}

On the other hand,
\begin{align*}
  \|\widehat{\W}^{(t)}\|_{2,1} -\|\widehat{\W}_0^{(t)}\|_{2,1}\leq \|\widehat{\W}^{(t)}-\widehat{\W}_0^{(t)}\|_{2,1}.
\end{align*}

As such, we have
\begin{equation}\label{eq:mid_bound_final_GS}
\begin{split}
&\frac{1}{2\eta}\|\widehat{\W}^{(t)} - \C\|_F^2 + \lambda\|\widehat{\W}^{(t)}\|_{2,1}  \\
&  - \biggl\{\min_{\W} \frac{1}{2\eta}\|\W - \C\|_F^2 + \lambda\|\W\|_{2,1}\biggr\}\\
=&\frac{1}{2\eta}( \|\widehat{\W}^{(t)} - \C\|_F^2 - \|\widehat{\W}_0^{(t)} - \C\|_F^2 )\\
 &+ \lambda(\|\widehat{\W}^{(t)}\|_{2,1} -\|\widehat{\W}_0^{(t)}\|_{2,1} )\\
 \leq &   \biggl[\frac{r_{c,s}}{ \eta}\biggl(\max_{j\in[d]}\|\C^j\|_{2}\biggr)^2+\lambda\biggl(\max_{j\in[d]}\|\C^j\|_{2}\biggr)\biggl]\\
 &\cdot\biggl[ \frac{k}{2\eta^2\lambda^2} \max_{j:\eta^2\lambda^2\leq\SSigma_{jj,0}}|\E_{jj}|\\
& + \frac{ (r_{c,s}-k)I(r_{c,s}>k)}{\eta\lambda} \max_{j:\eta^2\lambda^2>\SSigma_{jj,0}}\sqrt{|\E_{jj}|}\biggr]\\
=&\frac{1}{\eta}\biggl[\frac{r_{c,s}}{ \eta\lambda}\biggl(\max_{j\in[d]}\|\C^j\|_{2}\biggr)^2+ \biggl(\max_{j\in[d]}\|\C^j\|_{2}\biggr)\biggl]\\
 &\cdot\biggl[ \frac{k}{2\eta\lambda} \max_{j:\eta^2\lambda^2\leq\SSigma_{jj,0}}|\E_{jj}|\\
& + \max(0,r_{c,s}-k)  \max_{j:\eta^2\lambda^2>\SSigma_{jj,0}}\sqrt{|\E_{jj}|}\biggr].
\end{split}
\end{equation}

\end{proof}

\subsection{Proof of Theorem \ref{th:th_bound_LR_main_convex}}\label{sec:proof_bound_LR_main_convex}
\begin{proof}
First, consider the case with no acceleration. We first use Proposition 1 of \citet{schmidt2011convergence} by regarding procedures from Step \ref{eq:prox_start_LR} to Step \ref{eq:feature_tran-LR} as approximation for the proximal operator in \eqref{eq:proximal_op_LR}. Note that the norm clipping only bounds the parameter space and does not affect the results of \citet{schmidt2011convergence}. Then for $\varepsilon_t$ defined in Lemma \ref{th:lem_bound_vareps_t_LR} for $t\in[T]$, we have
\begin{align*}
\mathcal{E} =& \frac{2L}{m(T+1)^2}\biggl(\|\widetilde{\W}^{(0)}-\W_*\|_F \\
&+ 2\sum_{t=1}^{T}t\sqrt{\frac{2\varepsilon_t}{L}}+\sqrt{2\sum_{t=1}^{T}t^2\frac{\varepsilon_t}{L}}\biggr)^2.
\end{align*}

Meanwhile, by Lemma \ref{th:lem_bound_vareps_t_LR}, we have
\begin{align*}
\varepsilon_t = O\biggl( \frac{\kappa}{\epsilon_t}\biggr),
\end{align*}
where $\kappa = \frac{K^3\sqrt{m}kd\log d}{\eta}$.

On the other hand, because
\begin{align*}
 \epsilon = \sum_{t=1}^{T}\frac{(e^{\epsilon_t}-1)\epsilon_t}{(e^{\epsilon_t}+1)}
+\sqrt{\sum_{t=1}^{T}2\epsilon_t^2\log\biggl(e+\frac{\sqrt{\sum_{t=1}^{T}\epsilon_t^2}}{\delta}\biggr)},
\end{align*}
then by Lemma \ref{th:lem_bound_sum_eps_case_2_and_3}, we have
\begin{align*}
 \sqrt{\sum_{t=1}^{T}\epsilon_t^2}  \geq
\frac{\sqrt{2}\epsilon}{2\sqrt{\log(e+\epsilon/\sqrt{2}\delta)+2\epsilon} } = c_2.
\end{align*}

Then by Lemma \ref{th:lem_bound_error_poly}, we have
\begin{align*}
\sum_{t=1}^{T}\sqrt{\varepsilon_t} =
\left\{
  \begin{array}{ll}
    O\biggl( \sqrt{\frac{\kappa T^{\alpha+1/2} }{c_2(\alpha/2 - 1)^2\sqrt{2\alpha+1}  }}\biggr), &  {\alpha>2;} \\
    O\biggl( \sqrt{\frac{\kappa T^{5/2} }{c_2(\alpha/2 - 1)^2\sqrt{2\alpha+1}  }}\biggr), &  {-1/2<\alpha<2;} \\
    O\biggl( \sqrt{\frac{\kappa T^{2-\alpha} }{c_2(\alpha/2 - 1)^2\sqrt{-2\alpha-1}  }}\biggr), &  {\alpha<-1/2.}
  \end{array}
\right.
\end{align*}

Because $\widetilde{\W}^{(0)}$ is the result of the norm clipping, we have $\widetilde{\W}^{(0)} \in \mathcal{W}$.

Finally, taking $c_3 = \phi(\alpha)$ defined in \eqref{eq:bound_LR_main_basic_convex_phi} and $c_4 = \frac{\kappa}{c_2(\alpha/2 - 1)^2\sqrt{|2\alpha+1|} }$, under the assumption that $\W_*\in \mathcal{W}$, using Lemma \ref{th:lem_pseudo_bound_basic_convex}, we have the results for the case with no acceleration.

For the accelerated case, we use Proposition 2 of \citet{schmidt2011convergence} to have
\begin{align*}
\mathcal{E} =& \frac{2L}{m(T+1)^2}\biggl(\|\widetilde{\W}^{(0)}-\W_*\|_F \\
&+ 2\sum_{t=1}^{T}t\sqrt{\frac{2\varepsilon_t}{L}}+\sqrt{2\sum_{t=1}^{T}t^2\frac{\varepsilon_t}{L}}\biggr)^2.
\end{align*}
Then one can prove similarly combining Lemma \ref{th:lem_bound_vareps_t_LR}, Lemma \ref{th:lem_bound_sum_eps_case_2_and_3}, Lemma \ref{th:lem_bound_error_poly} and Lemma \ref{th:lem_pseudo_bound_acc_convex}.

\end{proof}

\subsection{Proof of Theorem \ref{th:th_bound_GS_main_convex}}\label{sec:proof_bound_GS_main_convex}
\begin{proof}
First, consider the case with no acceleration. We use Proposition 1 of \citet{schmidt2011convergence} and prove similarly as in Appendix \ref{sec:proof_bound_LR_main_convex}, combining Lemma \ref{th:lem_bound_vareps_t_GS}, Lemma \ref{th:lem_bound_sum_eps_case_2_and_3}, Lemma \ref{th:lem_bound_error_poly} and Lemma \ref{th:lem_pseudo_bound_basic_convex}.

For the accelerated case, we use Proposition 2 of \citet{schmidt2011convergence} and prove similarly as in Appendix \ref{sec:proof_bound_LR_main_convex},  combining Lemma \ref{th:lem_bound_vareps_t_GS}, Lemma \ref{th:lem_bound_sum_eps_case_2_and_3}, Lemma \ref{th:lem_bound_error_poly} and Lemma \ref{th:lem_pseudo_bound_acc_convex}.
\end{proof}

\subsection{Proof of Theorem \ref{th:th_bound_LR_main_strong_convex}}\label{sec:proof_bound_LR_main_strong_convex}
\begin{proof}
First, consider the case with no acceleration. We use Proposition 3 of \citet{schmidt2011convergence} to have
\begin{align*}
\mathcal{E}& =\frac{Q_0^T}{\sqrt{m}}\biggl(\|\widetilde{\W}^{(0)}-\W_*\|_F + 2\sum_{t=1}^{T}Q_0^{-t}\sqrt{\frac{2\varepsilon_t}{L}}\biggr).
\end{align*}
Then one can prove similarly as in Appendix \ref{sec:proof_bound_LR_main_convex}, combining Lemma \ref{th:lem_bound_vareps_t_LR}, Lemma \ref{th:lem_bound_sum_eps_case_2_and_3}, Lemma \ref{th:lem_bound_error_poly} and Lemma \ref{th:lem_pseudo_bound_basic_strong_convex}.

For the accelerated case, we use Proposition 4 of \citet{schmidt2011convergence} to have
\begin{align*}
\mathcal{E} =&\frac{(Q_0)^T}{m}\biggl(\sqrt{2(f(\widehat{\W}^{(0)}) - f(\W_*))} + 2\sqrt{\frac{L}{\mu}}\sum_{t=1}^{T}\sqrt{\varepsilon_t(Q_0)^{-t}} \\
&+ \sqrt{\sum_{t=1}^{T}\varepsilon_t(Q_0)^{-t}}\biggr)^2 .
\end{align*}
Then one can prove similarly as in Appendix \ref{sec:proof_bound_LR_main_convex}, using the assumption that $f(\widetilde{\W}^{(0)})-f(\W_*) = O(K^2Lm)$,  combining Lemma \ref{th:lem_bound_vareps_t_LR}, Lemma \ref{th:lem_bound_sum_eps_case_2_and_3}, Lemma \ref{th:lem_bound_error_poly} and Lemma \ref{th:lem_pseudo_bound_acc_strong_convex}.
\end{proof}

\subsection{Proof of Theorem \ref{th:th_bound_GS_main_strong_convex}}\label{sec:proof_bound_GS_main_strong_convex}
\begin{proof}
First, consider the case with no acceleration. We use Proposition 3 of \citet{schmidt2011convergence} and prove similarly as in Appendix \ref{sec:proof_bound_LR_main_strong_convex}, combining Lemma \ref{th:lem_bound_vareps_t_GS}, Lemma \ref{th:lem_bound_sum_eps_case_2_and_3}, Lemma \ref{th:lem_bound_error_poly} and Lemma \ref{th:lem_pseudo_bound_basic_strong_convex}.

For the accelerated case, we use Proposition 4 of \citet{schmidt2011convergence} and prove similarly as in Appendix \ref{sec:proof_bound_LR_main_strong_convex}, using the assumption that $f(\widetilde{\W}^{(0)})-f(\W_*) = O(K^2Lm)$,  combining Lemma \ref{th:lem_bound_vareps_t_GS}, Lemma \ref{th:lem_bound_sum_eps_case_2_and_3}, Lemma \ref{th:lem_bound_error_poly} and Lemma \ref{th:lem_pseudo_bound_acc_strong_convex}.
\end{proof}

\subsection{Proof of Theorem \ref{th:th_budget}}\label{sec:proof_th_budget}
\begin{proof}
Consider the bound in \eqref{eq:bound_LR_main_basic_convex}, whose logarithm is
\begin{align*}
&\phi(\alpha)\log\biggl(\frac{Kkd\log d\sqrt{\log(e+\epsilon/\sqrt{2}\delta)+2\epsilon} }{  \sqrt{m}\epsilon}\biggr) \\
&- \phi(\alpha)\log((\alpha/2 - 1)^2\sqrt{|2\alpha+1|}) + \log(K^2L)
\end{align*}
By Assumption \ref{th:asm_m}, the first term dominates. Then we should firstly maximize $\phi(\alpha)$, which results in that $\phi(\alpha) = 2/5$ and $-1/2<\alpha<2$. Then since $\phi(\alpha)$ is now fixed, we maximize $(\alpha/2 - 1)^2\sqrt{|2\alpha+1|}$, which results in $\alpha = 0$. Results under other settings can be proved similarly.
\end{proof}

{
\subsection{Proof of Claim \ref{claim:acc_runtime}}
\begin{proof}
First, consider the convex case. Taking Theorem \ref{th:th_bound_LR_main_convex} as an example. It is similar for Theorem \ref{th:th_bound_GS_main_convex}
Denote by $T_0$ and $T_1$ the runtimes for the no-acceleration and acceleration strategies, respectively.

By Theorem \ref{th:th_budget},  $T_0=\Theta((1/M)^{2/5}  )$ and $T_1=\Theta(((8/5)^2/M)^{2/9}  )$. Then we have $\log T_0=\Theta( -(2/5)\log M  )$ and $\log T_1=\Theta(
(4/9)\log(8/5) - (2/9)\log M)$.

Under Assumption \ref{th:asm_m}, $M\ll 1$. Then the terms with $(-\log M)$ dominate, respectively. Therefore, we have $\log T_1<\log T_0\rightarrow T_1<T_0$.

Next, consider the convex case. Taking Theorem \ref{th:th_bound_LR_main_strong_convex} as an example. It is similar for Theorem \ref{th:th_bound_GS_main_strong_convex}.

Denote by $T'_0$ and $T'_1$ the runtimes for the no-acceleration and acceleration strategies, respectively.

By Theorem \ref{th:th_budget}, we have
\begin{align*}
 T'_0&=\Theta(\log_{1/Q_0^2}((Q_0^{4/5} - 1)^2/M)  ) \\
 &= \Theta\biggl( \frac{1}{\log (1/Q_0^2)}(2\log(Q_0^{4/5} - 1) - \log M )  \biggr) \\
 T'_1&=\Theta(\log_{1/Q'_0}(((Q'_0)^{2/5} - 1)^2/M)  )\\
 &=\Theta\biggl( \frac{1}{\log (1/Q'_0)}(2\log((Q'_0)^{2/5} - 1) - \log M )  \biggr)
\end{align*}

Since $Q_0' = 1 - \sqrt{\mu/L}$ and $Q_0 = 1 - \mu/L$, assuming $\mu/L<0.3819$, we have $Q_0^2 = (1 - \mu/L)^2>1 - \sqrt{\mu/L}=Q_0'$. Therefore, $1/\log (1/Q_0^2) > 1/\log (1/Q'_0) $.

Under Assumption \ref{th:asm_m}, $M\ll 1$. Then the terms with $(-\log M)$ dominate, respectively. Therefore, we have $T'_1<T'_0$.
\end{proof}
}

\subsection{Proof of Proposition \ref{th:DP-MP-iter_example}}\label{sec:proof_DP-MP-iter_example}
\begin{proof}
First, consider the method of \citet{pathak2010multiparty}.

By Definition \ref{th:df_DP-MTL-iter}, an $(\epsilon,\delta)$-IP-MTL algorithm with $T=1$ should suffice for any set $\calS \subseteq \mathbb{R}^{d\times (m-1)}$ and all $i\in[m]$ that
\begin{align*}
\bbP( &  \hat{ \w}_{[-i]}^{(1)}  \in \calS  \mid  \W^{(0)}, \calD^m    ) \\
\leq   &e^{\epsilon} \bbP(  \hat{ \w}_{[-i]}^{(1)} \in \calS \mid  (\W')^{(0)}, (\calD')^m  )+ \delta.
\end{align*}

On the other hand, for the $\epsilon$ given in the method of~\citet{pathak2010multiparty}, using Theorem 4.1 of \citet{pathak2010multiparty}, taking $D=\calD^m$ and $D' = (\calD')^m $, we have for any set $\calS \subseteq \mathbb{R}^{d}$,
\begin{align*}
\bbP( &  {\hat{\w}}^s \in \calS  \mid \calD^m    )
\leq    e^{\epsilon} \bbP(  {\hat{\w}}^s \in \calS \mid   (\calD')^m  ),
\end{align*}
where ${\hat{\w}}^s  $ is defined in Section 3.2 of \citet{pathak2010multiparty}.

Because the method of \citet{pathak2010multiparty} uses ${\hat{\w}}^s$ for all the tasks, then we have $\hat{ \w}_{i}^{(1)} = {\hat{\w}}^s $ for all $i\in[m]$.

As such, denote $\W^{(0)}$ and $(\W')^{(0)}$ as the collections of models independently learned using $\calD^m $ and $(\calD')^m$, respectively. Then $\calD^m $ and $(\calD')^m$ contain all the information of $\W^{(0)}$ and $(\W')^{(0)}$, respectively. As such, we have for any set $\calS \subseteq \mathbb{R}^{d\times (m-1)}$, all $i\in[m]$ and $\delta = 0$ that
\begin{align*}
\bbP( &  \hat{ \w}_{[-i]}^{(1)}  \in \calS  \mid  \W^{(0)}, \calD^m    ) \\
\leq   &e^{\epsilon} \bbP(  \hat{ \w}_{[-i]}^{(1)} \in \calS \mid  (\W')^{(0)}, (\calD')^m  )+ \delta,
\end{align*}
which shows that the method of \citet{pathak2010multiparty} is an $(\epsilon,\delta)$-IP-MTL algorithm with $T=1$ and $\delta=0$.

Then we consider the method of~\citet{gupta2016differentially}. Assume a constant $\delta\geq 0$ and the number of iteration $T$ is given.

Taking $T_0 = m, t = i$ for $i\in [m]$, $\beta_t = \hat{\w}_i$, $\mathscr{D} = \calD^m$, for the $\epsilon$ given in the method of~\citet{gupta2016differentially}, using Theorem 1 of~\citet{gupta2016differentially}, for $t\in[T]$, we have \emph{in the $t$-th each iteration}, for any set $\calS \subseteq \mathbb{R}^{d\times m}$ and all $i\in[m]$,
\begin{align*}
\bbP( &  \hat{ \W}^{(t)}    \in \calS  \mid    \calD^m    )
\leq    e^{\epsilon} \bbP(  \hat{ \W}^{(t)}   \in \calS \mid   (\calD')^m  ),
\end{align*}
which suggests that for any set $\calS \subseteq \mathbb{R}^{d\times (m-1)}$ and all $i\in[m]$,
\begin{align*}
\bbP( &  \hat{ \w}_{[-i]}^{(t)}    \in \calS  \mid    \calD^m    )
\leq    e^{\epsilon} \bbP(  \hat{ \w}_{[-i]}^{(t)}   \in \calS \mid   (\calD')^m  ).
\end{align*}

Then for all $i\in[m]$ and for all $t\in[T]$, take the $t$-th output $\theta_{t,i} = \hat{\w}_{[-i]}^{(t)}$ and $\delta_t = 0$.

Therefore by the \emph{Adaptive composition} property (Property \ref{th:lem_Ad_comp}), for all $i\in[m]$ and for any set $\calS\subset\mathbb{R}^{d\times (m-1)\times T}$,
\begin{align*}
  &\mathbb{P} ((\theta_{1,i},\ldots,\theta_{T,i} )\in \mathcal{S}  \mid \bigcap_{t=1}^T(\calB_t = (\calD^m, \ttheta_{1:t-1}))  )\\
  \leq & e^{\tilde{\epsilon}}\mathbb{P}((\theta_{1,i},\cdots,\theta_{T,i})\in \mathcal{S}\mid \bigcap_{t=1}^T(\calB_t = ((\calD')^m, \ttheta_{1:t-1}) ))\\
   &+ \delta,
\end{align*}
where for all $t\in[T]$, $\calB_t$ denotes the input for the $t$-th iteration,
\begin{numcases}{\ttheta_{1:t-1} = }
\nonumber \emptyset, & $t=1$\\
\nonumber (\theta_{1,1},\dots,\theta_{1,m}) \ldots,(\theta_{t-1,1},\dots,\theta_{t-1,m}), & $t\geq2$,
\end{numcases}
and $\tilde{\epsilon}$ is defined as follows.
  \begin{align*}
 \tilde{\epsilon} = &\min\biggl\{ \sum_{t=1}^{T}\epsilon,\sum_{t=1}^{T}\frac{(e^{\epsilon}-1)\epsilon}{(e^{\epsilon}+1)}
+\sqrt{\sum_{t=1}^{T}2\epsilon^2\log\biggl(\frac{1}{\delta}\biggr)},\\
 &\sum_{t=1}^{T}\frac{(e^{\epsilon}-1)\epsilon}{(e^{\epsilon}+1)}
+\sqrt{\sum_{t=1}^{T}2\epsilon^2\log\biggl(e+\frac{\sqrt{\sum_{t=1}^{T}\epsilon^2}}{\delta}\biggr)}
\biggr\}.
 \end{align*}

As such, in each $t$-th iteration, denote $\W^{(t-1)}$ and $(\W')^{(t-1)}$ as the collections of models independently learned using $(\calD^m, \ttheta_{1:t-1})$ and $(\calD')^m,\ttheta_{1:t-1})$, respectively. Then $(\calD^m, \ttheta_{1:t-1})$ and $(\calD')^m,\ttheta_{1:t-1})$ contain all the information of $\W^{(t-1)}$ and $(\W')^{(t-1)}$, respectively.

Therefore, we have for any set $\calS\subset\mathbb{R}^{d\times (m-1)\times T}$,
\begin{align*}
  &\mathbb{P} (\w_{[-i]}^{(1:T)}\in \mathcal{S}  \mid \bigcap_{t=1}^T(\calB_t = (\W^{(t-1)},\calD^m, \ttheta_{1:t-1}))  )\\
  \leq & e^{\tilde{\epsilon}}\mathbb{P}(\w_{[-i]}^{(1:T)}\in \mathcal{S}\mid \bigcap_{t=1}^T(\calB_t = ((\W')^{(t-1)},(\calD')^m, \ttheta_{1:t-1}) ))\\
   &+ \delta,
\end{align*}
which shows that by Definition \ref{th:df_DP-MTL-iter}, the method of~\citet{gupta2016differentially} is an $(\tilde{\epsilon},\delta)$-IP-MTL algorithm.
\end{proof}

}

\subsection{Proof of Proposition \ref{th:prop_DP-MP-iter}}
\begin{proof}

 Given an $(\epsilon,\delta)$ - IP-MTL algorithm $\calA(\calB)$, by Definition \ref{th:df_DP-MTL-iter}, we have for any set $\calS\subseteq \mathbb{R}^{d\times (m-1)\times T}$ that
\begin{align*}
\bbP( &\hat{\w}_{[-i]}^{(1:T)}  \in \calS  \mid \bigcap_{t=1}^T \calB_t = (\W^{(t-1)}, \calD^m, \ttheta_{1:t-1}) ) \\
\leq   &\exp(\epsilon) \bbP(  \hat{\w}_{[-i]}^{(1:T)} \in \calS \mid \bigcap_{t=1}^T \calB_t = ((\W')^{(t-1)}, (\calD')^m, \ttheta_{1:t-1})) \\
&+ \delta.
\end{align*}

Furthermore, following the proof of the \emph{Group privacy} property (Property \ref{th:lem_group_dp}), shown by~\citet{vadhan2016complexity}, for protecting the entire dataset, $n$ data instances, of the $i$-th task, we construct a series of datasets, $\calD_{(0)}^m,\calD_{(1)}^m,\ldots,\calD_{(n)}^m$, and let $\calD_{(0)}^m = \calD^m, \calD_{(n)}^m=(\calD')^m $ such that for $j=0,\ldots,n-1$, $\calD_{(j)}^m$ and $\calD_{(j+1)}^m$ are two neighboring datasets that differ in one data instance. Let a series of model matrices, $\W_{(0)},\ldots,\W_{(n)}$, where $\W_{(0)}=\W, \W_{(n)} = \W'$,  be the input model matrices in those settings. Let a series of output objects $\ttheta_{1:t-1}^{(0)},\ldots,\ttheta_{1:t-1}^{(n)}$, where $\ttheta_{1:t-1}^{(0)}=\ttheta_{1:t-1}, \ttheta_{1:t-1}^{(n)} = \W'$, be the output objects in those settings.

Then, we have
\begin{align*}
\bbP ( \hat{\w}&_{[-i]}^{(1:T)}  \in \calS  \mid \bigcap_{t=1}^T \calB_t = (\W_{(0)}^{(t-1)}, \calD_{(0)}^m, \ttheta_{1:t-1}^{(0)})   ) \\
\leq& \exp(\epsilon) \bbP( \hat{\w}_{[-i]}^{(1:T)}  \in \calS  \mid \bigcap_{t=1}^T \calB_t = (\W_{(1)}^{(t-1)}, \calD_{(1)}^m, \ttheta_{1:t-1}^{(1)}) ) \\
&+  \delta\\
&\vdots\\
\leq& \exp(n\epsilon) \bbP( \hat{\w}_{[-i]}^{(1:T)}  \in \calS  \mid \bigcap_{t=1}^T \calB_t = (\W_{(n)}^{(t-1)}, \calD_{(n)}^m, \ttheta_{1:t-1}^{(n)})) \\
&+  (1+\exp(\epsilon)+\ldots+\exp((n-1)\epsilon)) \delta \\
\leq& \exp(n\epsilon) \bbP( \hat{\w}_{[-i]}^{(1:T)}  \in \calS  \mid \bigcap_{t=1}^T \calB_t = (\W_{(n)}^{(t-1)}, \calD_{(n)}^m, \ttheta_{1:t-1}^{(n)})) \\
&+  n\exp({n\epsilon})\delta,
\end{align*}
which renders $\calA$ as an $(n\epsilon,n\exp({n\epsilon})\delta)$ - MP-MTL algorithm.
\end{proof}

\savestatus

\end{document}